\documentclass{article}
\usepackage{ijcai25}

\usepackage{times}
\usepackage{soul}
\usepackage{url}
\usepackage[hidelinks]{hyperref}
\usepackage[utf8]{inputenc}
\usepackage[small]{caption}
\usepackage{graphicx}
\usepackage{amsmath}
\usepackage{amsthm}
\usepackage{booktabs}
\usepackage{algorithm}
\usepackage{algorithmic}
\usepackage[switch]{lineno}

\usepackage{helvet}  
\usepackage{courier}  
\usepackage{caption} 
\usepackage{adjustbox}
\usepackage{amsfonts}
\usepackage{diagbox}
\usepackage{makecell}
\usepackage{mathrsfs}
\usepackage{multirow}
\usepackage{nomencl}
\usepackage{subcaption}
\usepackage{tabularx}
\usepackage{xcolor}
\usepackage[normalem]{ulem} 

\makenomenclature
\usepackage{newfloat}
\usepackage{listings}
\usepackage{eso-pic}
\usepackage{tikz}

\newcommand{\FirstPageLeftNote}[3]{%
  \AddToShipoutPictureFG*{%
    \ifnum\value{page}=1
      \begin{tikzpicture}[remember picture,overlay]
        \node[
          anchor=south west,
          xshift=#1, yshift=#2,
          text width=0.48\textwidth, 
          inner sep=0pt
        ] at (current page.south west) {%
          \hrule\vspace{3pt}%
          {\footnotesize #3}%
        };
      \end{tikzpicture}
    \fi
  }%
}

\title{CCAD: Compressed Global Feature Conditioned Anomaly Detection}

\author{
{\Large\bfseries Xiao Jin}$^{1,*}$\and
{\Large\bfseries Liang Diao}$^{2,*}$\and
{\Large\bfseries Qixin Xiao}$^{3}$\and
{\Large\bfseries Yifan Hu}$^{3}$\\
{\Large\bfseries Ziqi Zhang}$^{4,6}$\and
{\Large\bfseries Yuchen Liu}$^{5}$\and
{\Large\bfseries Haisong Gu}$^{6,\dagger}$\\
\affiliations
$^1$Columbia University \\
$^2$Ping An Property \& Casualty Insurance Company \\
$^3$University of Michigan\\
$^4$University of California at Berkeley \\
$^5$Stevens Institute of Technology \\
$^6$VisionX LLC.\\ [3pt]
\emails
xj2285@columbia.edu, \{diaoliang91, cml.bkl789, liuyuchen1119\}@gmail.com, \\
\{qxiaocs, yfh\}@umich.edu, haisonggu@ieee.org\\
}

\begin{document}

\FirstPageLeftNote{2.0cm}{1.7cm}{%
\textsuperscript{*} indicates equal contribution.\par
\mbox{\textsuperscript{$\dagger$} corresponding author: Haisong Gu (email: haisonggu@ieee.org).}%
}

\maketitle
\begin{abstract}
Anomaly detection holds considerable industrial significance, especially in scenarios with limited anomalous data. Currently, reconstruction-based and unsupervised representation-based approaches are the primary focus. However, unsupervised representation-based methods struggle to extract robust features under domain shift, whereas reconstruction-based methods often suffer from low training efficiency and performance degradation due to insufficient constraints. To address these challenges, we propose a novel method named Compressed Global Feature Conditioned Anomaly Detection (CCAD). CCAD synergizes the strengths of both paradigms by adapting global features as a new modality condition for the reconstruction model. Furthermore, we design an adaptive compression mechanism to enhance both generalization and training efficiency. Extensive experiments demonstrate that CCAD consistently outperforms state-of-the-art methods in terms of AUC while achieving faster convergence. In addition, we contribute a reorganized and re-annotated version of the DAGM 2007 dataset with new annotations to further validate our method's effectiveness.
The code for reproducing main results is available at 
{\normalsize\ttfamily\mdseries \url{https://github.com/chloeqxq/CCAD}}.

\end{abstract}

\section{Introduction}


Anomaly Detection (AD) is a crucial task in computer vision, particularly in industrial applications such as autonomous driving, medicine, and manufacturing, where it is essential to identify anomalous images and localize regions of anomalies. Unlike traditional supervised learning, AD faces the significant challenge of inaccessible anomalies during the training phase. Consequently, many existing AD methods adopt a zero-shot learning framework, 
where models are trained solely on nominal data, learning their distribution to effectively discriminate anomalous instances based on noticeable deviations during the inference process~\cite{DN2,padim,patchcore}.

\noindent Mainstream AD methods can be mainly categorized into three approaches: unsupervised-representation-based, reconstruction-based, and knowledge distillation-based methods. For unsupervised-representation-based methods, a feature extractor is trained to model normal data distribution, and some unsupervised learning approach (\textit{e.g.,} k-nearest neighbors) is applied to detect anomalies. Recent works \cite{DN2,patchcore,ReConPatch} use models pretrained on large-scale datasets to extract the global feature space from the normal dataset, ensuring method generalization and pushing the accuracy limits of feature-based approaches. However, the domain gap between large-scale pre-training datasets and downstream data, along with the inflexibility of rule-based global feature selection, limits the performance of feature-based approaches. 
 Knowledge distillation is another widely used AD approach. Knowledge of normal samples is transferred from a large pre-trained teacher model to small-scale student models. However, student models often exhibit inferior performance on anomaly samples compared to the teacher model. This disparity serves as a key indicator for detecting anomalies.
\begin{figure}[t]
    \centering
    \begin{minipage}[c]{0.49\textwidth}
        \centering
        
        \subcaptionbox{Vanilla DM\label{1a}}[0.49\linewidth]{
            \includegraphics[width=0.98\linewidth]{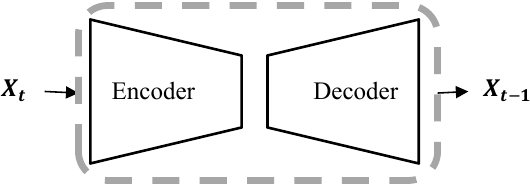}
        }
        \subcaptionbox{Local Feature CDM\label{1b}}[0.49\linewidth]{
            \includegraphics[width=0.98\linewidth]{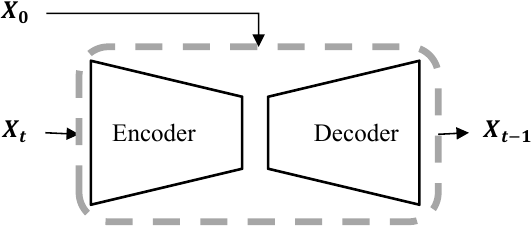}
        }

        \vspace{2mm}  

        \subcaptionbox{Global Feature CDM (Ours)\label{1c}}[1\linewidth]{
            \includegraphics[width=0.98\linewidth]{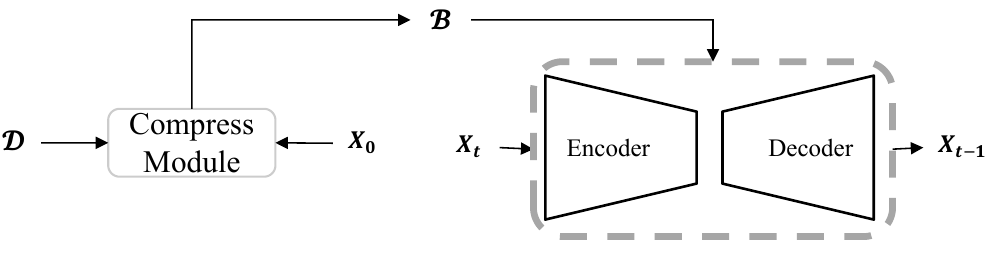}
        }
        \vspace{-2mm}
        \caption{An overview of Diffusion Modules (DM) and Conditioned Diffusion Modules (CDM): (a) The vanilla DM operates without any condition. (b) A single sample $\mathbf{x}_0$ is used as the condition (c) Compressed vectors $\mathcal{B}$ representing the distribution of a whole dataset are served as the condition.}
        \label{fig:method_overview_compare}
    \end{minipage}
\end{figure}\\
In reconstruction-based methods, image generation models trained solely on normal samples are employed to reconstruct anomalous samples into normal ones during the inference phase \cite{GANomaly,AnoDDPM,diffusionad,DDAD,DiAD}. By comparing the original sample with the reconstructed ones, anomaly locations can be identified.\\
As diffusion models progress in Generative AI, an increasing number of AD methods are now leveraging diffusion models as the core component for reconstruction-based approaches. As illustrated in Figure 1, \cite{AnoDDPM} (Figure \ref{1a}) is the pioneer in utilizing a diffusion model (DM) for anomaly detection. Subsequently, \cite{DDAD,DiAD}  (Figure \ref{1b}) advances the reconstruction quality by conditioning it on the input images (local feature). Nonetheless, the information provided by local features is inherently limited, and the absence of adequate prior conditions impairs training efficiency and constrains the potential for achieving higher accuracy. To address these challenges and improve both accuracy and resource efficiency in model training, we introduce a novel method: the Compressed Global Feature Conditioned Anomaly Detection Module (CCAD). This method integrates feature-based and reconstruction-based approaches, utilizing global features as auxiliary conditions to enhance reconstruction quality. Furthermore, CCAD incorporates a two-stage feature compression mechanism to optimize the trade-off between performance and efficiency (Figure \ref{1c}).\\
Here are the main contributions of our paper:

\begin{itemize}
    \item We propose a method called CCAD that uses global features as prior conditions for reconstruction-based anomaly detection. To the best of our knowledge, this is the first time global features have been used to enhance the reconstruction quality of diffusion models.
\end{itemize}

\begin{itemize}
    \item We explore the selection of global features and validate the feasibility and necessity of feature compression. We also propose a two-stage global feature compression mechanism that uses a coarse-to-fine approach to convert global features into prior conditions. This mechanism ensures both the performance and efficiency of CCAD.
\end{itemize}

\begin{itemize}
    \item Extensive experiments verify the effectiveness of CCAD, supported by empirical analyses that illustrate how the proposed methods enhance anomaly detection performance. In addition, the DAGM 2007 dataset is revisited, and the images are re-annotated. Compared to the original ground truth, these new annotations are more accurate, providing a more reliable evaluation of anomaly detection methods. The annotated data is made publicly available.
\end{itemize}
\vspace{-4mm}

\section{Related Works}
Mainstream anomaly detection (AD) methods can be categorized into three primary approaches: knowledge distillation-based, unsupervised-representation-based, and reconstruction-based methods.
\vspace{-2mm}
\subsection{Knowledge distillation-based approach}
Knowledge distillation-based methods were used to transfer knowledge from a large, well-trained teacher model to a smaller student model, aiming to replicate the teacher's performance. S-T AD introduced a student-teacher framework using discriminative latent embeddings \cite{stad}, and EfficientAD \cite{efficientad} optimized the distillation process resulted in better efficiency and lower computational requires. 
\subsection{Unsupervised-representation-based approach}
Earlier work \cite{yi2020patch} depended on specific unsupervised training methods to achieve effective results on downstream datasets, as early pre-trained models struggled with generalization. However, advancements in pre-trained models have enabled recent approaches to extract anomalous image features and apply unsupervised techniques, such as KNN, to detect anomalies from globally extracted features.
DN2 \cite{DN2} utilized simple ResNets \cite{resnet} with high-level feature representations from pre-trained ImageNet. SPADE \cite{SPADE} introduced the concept named as memory banks for better reuses of pre-trained features for both pixel-level and image-level anomaly detection. PaDiM \cite{padim} later proposed the patch-level feature banks to estimate the patch-level Mahalanobis distances. PatchCore \cite{patchcore} and ReConPatch \cite{ReConPatch} used similar patch-level memory banks, but they used coreset sampling methods to reduce the inference costs notably while retaining the higher performance. \\
These unsupervised learning methods heavily rely on the quality of global feature spaces from the pre-trained networks. Futhermore, previous methods are only limited on single-class AD tasks. 

\vspace{-2mm}
\begin{figure*}[t]
\centering
\includegraphics[width=0.9\textwidth]{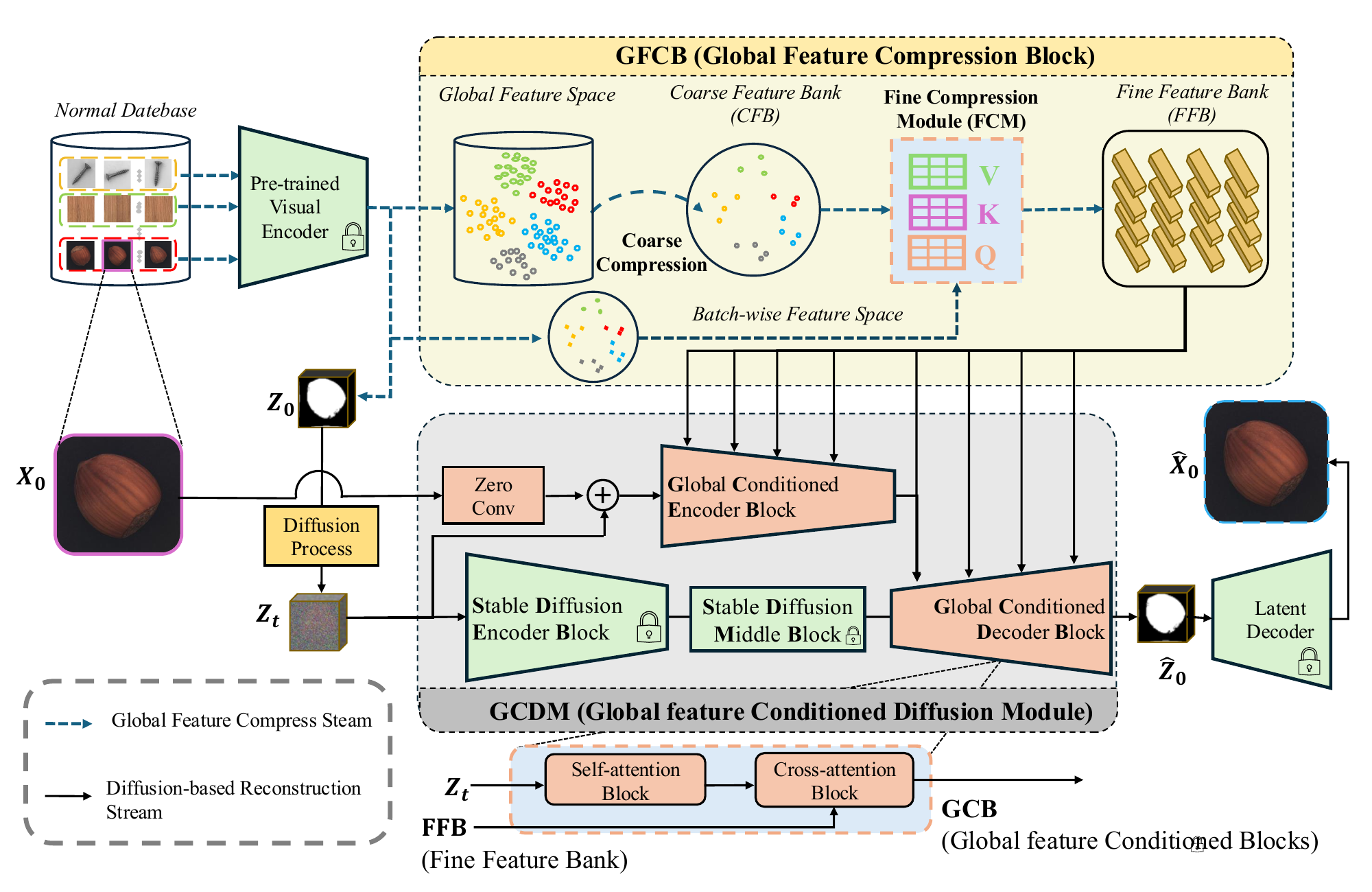} %
\vspace{-2mm}
\caption{The framework of CCAD(F). Our method consists of two main part: Global Feature Compression Block (GFCB) and Global feature Conditioned Diffusion Module (GCDM). FFB denotes Fine Feature Bank, and GCB denotes Global Feature Conditioned Block.}
\label{CCAD_F_pipline}
\end{figure*}

\subsection{Reconstruction-based approach}
This approach utilized generative models to reconstruct normal images. GANomaly \cite{GANomaly} was the first work to apply GAN in reconstruction-based anomaly detection tasks. 
As diffusion models emerged as powerful generative tools, they were also adapted for reconstruction-based anomaly detection. Diffusion models was first proposed by DDPM\cite{DDPM} in which gradually adding noise to data in a forward process and then learning to reverse this process in a way that can generate new data samples from pure noise. DDIM\cite{DDIM} improved DDPM by introducing a deterministic sampling process that reduced steps for faster, more efficient image generation. Later, Latent Diffusion Models (LDM)\cite{SD} optimized this process by working in a lower-dimensional latent space. ControlNet \cite{ControlNet} further enhanced diffusion models by adding spatial conditioning controls while preserving the original model’s parameters.\\
Partial diffusion with simplex noise in DDPMs was employed by AnoDDPM \cite{AnoDDPM} to detect large abnormalities. DiffusionAD \cite{diffusionad} introduced a diffusion model with norm-guided and one-step denoising paradigm. Following this, DDAD \cite{DDAD} utilized a conditioned denoising diffusion model which enhanced the accuracy by reconstructing normal samples to match a target image. DiAD \cite{DiAD} proposed a multi-class anomaly detection framework, combined with a pixel-space autoencoder, Semantic-Guided Network, and Spatial-aware Feature Fusion block, while POUTA \cite{pouta} leveraged encoder-decoder feature discrepancies.
\vspace{-3mm}
\section{Method}
In this section, we introduce some necessary preliminaries and implementation of our CCAD method.
As shown in Figure \ref{CCAD_F_pipline}, our method consists of two streams: the Global Feature Compression Stream, which compresses the global feature space to obtain the global feature bank, and the Diffusion-based Reconstruction Stream, which uses the global feature bank as a condition to generate normal images. To meet different application requirements, we also design three variants: CCAD(F), CCAD(C), CCAD(V).

\vspace{-2mm}

\subsection{Preliminaries}
\subsubsection{Latent diffusion Module}
The diffusion model is the foundation of our method. To facilitate understanding, we revisit the ControlNet-based latent diffusion module. A pre-trained encoder, denoted as $\mathcal{E}$, processes an input image $\mathbf{x}_0$, converting it from pixel space (with a height of $H$ and a width of $W$) into latent space as $\mathbf{z}_0$
\begin{align}
    \mathbf{z}_0 = \mathcal{E}(\mathbf{x}_0), \textbf{x}_0 \in \mathbb{R}^{H\times W\times 3}\label{encoder}.
\end{align}
The diffusion process and the training objective function
can be denoted by
\begin{align}
\mathbf{z}_t &= \sqrt{\bar{\alpha}_{t}}~\mathbf{z}_0 + \sqrt{1 - \bar{\alpha}_t}\boldsymbol{\epsilon}\\
\mathcal{L}&= \mathbb{E}_{\mathbf{z}, \boldsymbol{\epsilon} \sim \mathcal{N}(0, \boldsymbol{I}),t \sim U(1, T)}[\|\boldsymbol{\epsilon} - \boldsymbol{\epsilon}^t_{\boldsymbol{\Theta}}(\mathbf{z}_{t}; \textbf{c}_f)\|_2^2]\label{obj_DM}.
\end{align}
where $\boldsymbol{\epsilon}^t_{\boldsymbol{\Theta}}(\mathbf{z}_{t}; \textbf{c}_f)$ is the backbone with learnable parameters $\boldsymbol{\Theta}$ and $\textbf{c}_f$ is the spatial condition from ControlNet.

\subsection{Global Feature Compression Block}
\subsubsection{Global Feature Space}
\noindent Given a dataset containing $N$ nominal image samples, denoted as $\mathcal{X} = \{\textbf{x}_1, \dots, \textbf{x}_n, \dots, \textbf{x}_N\}$, we utilize a pre-trained visual encoder $\mathcal{F}$ to map the dataset into a $d$-dimensional global feature space $\mathcal{D}$ in the following way
\begin{align}
    \mathcal{F}:\mathcal{X}\mapsto\mathcal{D}, \mathcal{D} &= \{\textbf{v}_n | \textbf{v}_{n} \in \mathbb{R}^{d}, n = 1, 2\dots, M \}\label{glb_feature}\\
    M &= N\times\lfloor\frac{H}{m}\rfloor\times\lfloor\frac{W}{m}\rfloor\label{M}
\end{align}
where $\textbf{v}_n = \mathcal{F}(\textbf{x}_n)$, and $m$ indicates the downsampling ratio.

\subsubsection{Coarse Feature Bank (CFB)}
As the size of the dataset increases, the global feature space also becomes exceedingly large. Directly using the entire global feature space as a condition in a diffusion-based model is not feasible. Therefore, we perform a coarse compression on the global feature space through coreset sampling to obtain a coarse feature bank. In Patchcore \cite{patchcore}, coreset selection is applied to the global feature space with a certain ratio, such as $10\%$. Generally, to ensure the performance of anomaly detection, the coreset sampling ratio is no less than $1\%$. However, even with this small ratio, the proposed "Memory Bank" \cite{patchcore} still contains more than ten thousand samples, making it impractical to use as a condition input for a diffusion model. To address this issue, we set the number of samples in the Coarse Feature Bank as a fixed number $\xi$ and we keep $\xi$ no more than 1000 in our case. While it may suppress the performance of Patchcore, as auxiliary conditions in the model, a few hundred to a thousand samples are sufficient to be representative.
The coarse compression is denoted by
\begin{align}
    \mathcal{S}:\mathcal{D}\mapsto\mathcal{B}_c, \mathcal{B}_c = \{\textbf{v}_k | \textbf{v}_{k} \in \mathbb{R}^{d}, k = 1, 2\dots, \xi \}\label{coarse_compression}
\end{align}
where $\mathcal{S}$ is the greedy coarse compression in \cite{patchcore} and $\mathcal{B}_c$ is the Coarse Feature Bank.

\subsubsection{Fine Feature Bank (FFB)}
Through the process in \eqref{coarse_compression}, CFB $\mathcal{B}_c$ is feasible to serve as a global feature condition helping image reconstruction in the LDM backbone. However, during the image reconstruction process, unlike batch-wise training data samples, the $\mathcal{B}_c$ introduced as the external condition in each iteration remains the same. Intuitively, we hope to select out the most relevant global feature information from the $\mathcal{B}_c$ for each data sample. Inspired by the unsupervised methods such as K-NN in previous works and the encoder in LDM, we propose the concept of a Fine Feature Bank. The encoder $\mathcal{E}$ in \eqref{encoder} and the pre-trained visual encoder $\mathcal{F}$ in \eqref{glb_feature} share the same architecture and parameters to ensure consistency and efficient feature representation. As a result, besides the $\mathcal{B}_c$ generated from \eqref{coarse_compression}, we can also generate a small batch-wise feature space $\mathcal{D}_{bs}$ including $\zeta$ samples through \eqref{glb_feature} and \eqref{M}. We then build a trainable Fine Compression Module (FCM) $\boldsymbol{\tau}_{\boldsymbol{\theta}}$ mapping $\mathcal{B}_c$ and $\mathcal{D}_{bs}$ to Fine Feature Bank (FFB) via a multi-head cross-attention layer. We define $\mathbf{D}_{bs} \in \mathbb{R}^{\zeta\times d}$ as the vectorized $\mathcal{D}_{bs}$ and $\mathbf{B}_{c} \in \mathbb{R}^{\xi\times d}$ as the vectorized $\mathcal{B}_c$.
\begin{align}
    \mathbf{Q} = \mathbf{D}_{bs}\boldsymbol{\theta}_Q;
    \mathbf{K} &= \mathbf{B}_{c}\boldsymbol{\theta}_W;
    \mathbf{V} =\mathbf{B}_{c}\boldsymbol{\theta}_V\label{qkv}\\
    \mathbf{B}_f = \boldsymbol{\tau}_{\boldsymbol{\theta}}(\mathcal{D}_{bs}, \mathcal{B}_{c}) &= \text{softmax}(\frac{\mathbf{Q}\mathbf{K}^{\text{T}}}{\sqrt{d_k}})\mathbf{V}\boldsymbol{\theta}_B \in \mathbb{R}^{\zeta\times d}
\end{align}
where $\mathbf{B}_f$ is the vectorized form of Fine Feature Bank (FFB) $\mathcal{B}_f$; $d_k$ is the scaling factor and $\boldsymbol{\theta}:=\{\boldsymbol{\theta}_Q \in \mathbb{R}^{d\times d_k}, \boldsymbol{\theta}_W \in \mathbb{R}^{d\times d_k}, \boldsymbol{\theta}_V \in \mathbb{R}^{d\times d_k}, \boldsymbol{\theta}_B \in \mathbb{R}^{d_k\times d}\}$ is the learnable parameter matrices \cite{attention,perceiver}. 
To enable FFB or $\mathcal{B}_c$ to serve as conditional inputs in the reconstruction process, we modified the U-Net backbone \cite{U-Net} architectures in DDAD and DiAD by introducing Global feature Conditioned Blocks (GCB) to support the input of global features as embedding conditions.


\subsection{Global feature Conditioned Diffusion Module}
\begin{figure}[t]
\begin{minipage}[c]{0.45\textwidth}
    \centering
    \begin{subfigure}[t]{0.45\textwidth}
    \centering
    \includegraphics[width=3.0cm]{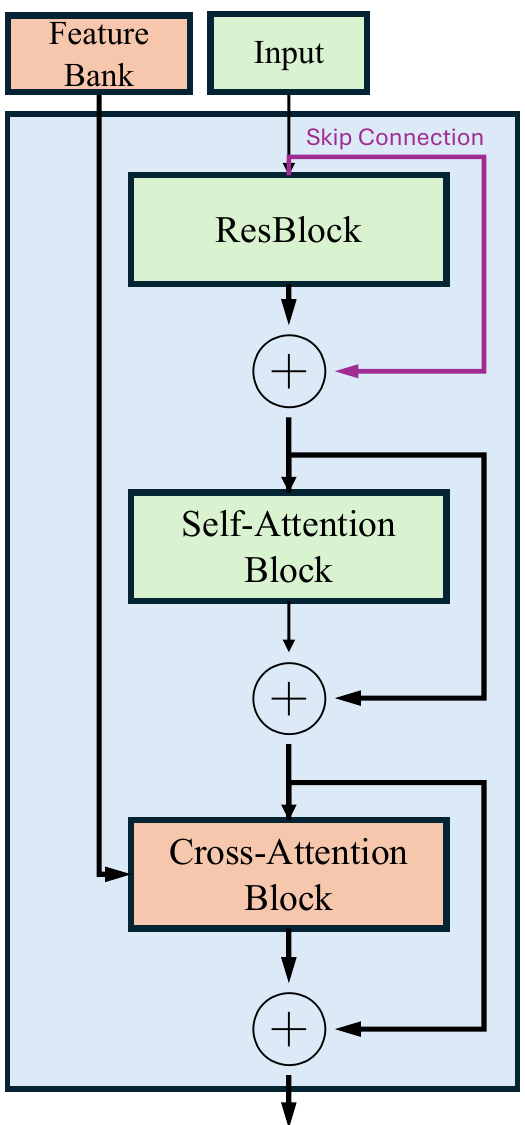}
    \subcaption{GCB in CCAD(V)}\label{GCB_1}
    \end{subfigure}
    \begin{subfigure}[t]{0.49\textwidth}
    \centering
    \includegraphics[width=3.0cm,]{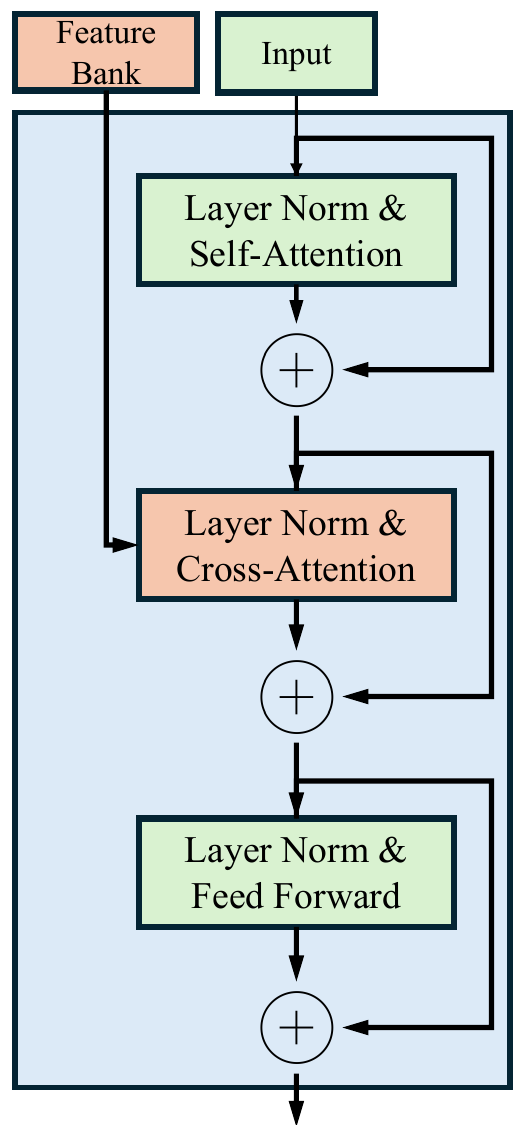}
    \subcaption{GCB in CCAD (F \& C)}\label{GCB_2}
    \end{subfigure}
    \captionsetup{justification=centering,margin=0.1cm}
    \vspace{-2mm}
     \caption{Architecture of Global feature Conditioned Blocks.}
    \label{GCB structure}
\end{minipage}
\end{figure}

The Global feature Conditioned Diffusion Module (GCDM) is the main component of the reconstruction stream. It uses the Global Feature Conditioned Block (GCB) to integrate the global feature bank and refine the reconstruction quality. Corresponding to the different UNet structures in DDAD \cite{DDAD} and DiAD \cite{DiAD}, we implement two variants of the Global Feature Conditioned Block (GCB). In DDAD, they employed a modified UNet in \cite{UNet_DDAD}. The network introduces attention blocks at 32×32, 16×16, and 8×8 resolutions. As shown in figure \ref{GCB_1}, we add an extra cross-attention block following each corresponding combination of a ResBlock and Self-Attention Block. In DiAD, the external text embedding condition is integrated through the Basic Transformer Block, and we replace that module with the GCB in figure \ref{GCB_2} instead.

\noindent Based on the backbone structures in DDAD and DiAD, we propose three different CCAD variants, named as follows:
\begin{itemize}
    \item CCAD with Fine Feature Bank, \textit{i.e.}, CCAD(F)
    \item CCAD with Coarse Feature Bank, \textit{i.e.}, CCAD(C)
    \item Vanilla CCAD, \textit{i.e.}, CCAD(V)
\end{itemize}
In CCAD(F) and CCAD(C), similar to \cite{SD}, the diffusion process and denoising process are conducted on the latent space. It uses a  pre-trained Autoencoder \cite{VQGAN} to convert variables between pixel space and latent space. Same as \cite{DiAD}, CCAD(F) and CCAD(C) support multi-class anomaly detection tasks. In CCAD(V), the diffusion process and reconstruction process are conducted on the pixel space. Since no Batch-wise Feature Space can be obtained during the training process, we only utilize CFB $\mathcal{B}_c$ as the global condition.


\begin{table*}[t]
\caption{Comparison of CCAD with state-of-the-art anomaly detection methods.}
\centering
\begin{tabular}{ccccccc}
      \toprule
      \thead{Method} & \thead{Multi-class\\ anomaly detection} & \thead{Global Feature \\Utilized} & \thead{Feature Space \\ Compressed} & \thead{Trainable\\ Model}&\thead{Minimal epochs \\ $eph$ required} & \thead{AUC on \\MVTec-AD at $eph$}\\
      \hline
      \makecell{SPADE\\} & \makecell{No} & \makecell{Yes} & No & \makecell{-} & -&0.850; 0.602\\
      \hline
      \makecell{Pathcore\\} & \makecell{No} & \makecell{Yes} & Reduced & \makecell{-} & -&0.858; 0.948\\
      \hline
      \makecell{DDAD\\} & \makecell{No} & \makecell{No} & - & \makecell{UNet}&1500&0.962; 0.966\\
      \hline
      DiAD  & \makecell{Yes} & \makecell{No} & - & \makecell{SGN \cite{DiAD}}& 200&0.950; 0.954\\
      \hline
      CCAD(V) & \makecell{No} & \makecell{Yes} & Yes & \makecell{GCDM}&1000&0.963;0.961\\
      \hline
      CCAD(C) & \makecell{Yes} & \makecell{Yes} & Yes & \makecell{GCDM}& 100&0.958; 0.958\\
      \hline
      CCAD(F)  & \makecell{Yes} & \makecell{Yes} & Yes & \makecell{GCDM+ FCM}& 110&0.951; 0.961\\
      \bottomrule
    \end{tabular}
\label{comparison_state_of_art}
\vspace{-4mm}
\end{table*}

\subsubsection{CCAD(F)}
The Global Feature Conditioned Diffusion Module of CCAD(F) is shown in figure \ref{CCAD_F_pipline} which mainly contains the following components:
\begin{itemize}
    \item Stable Diffusion Encoder Block (SDEB)
    \item Stable Diffusion Middle Block (SDMB)
    \item Global Conditioned Encoder Block (GCEB)
    \item Global Conditioned Decoder Block (GCDB)
\end{itemize}
Both CCAD(F) and CCAD(C) use ControlNet(\cite{ControlNet}) as the main network for reconstruction, with GCB to grab useful information from the global feature bank.
Given a certain image $\mathbf{x}$, Each denoising iteration can be formulated as:
\begin{align}
\mathbf{z}_{t-1} = &\sqrt{\bar{\alpha}_{t-1}}(\frac{\mathbf{z}_{t} - \sqrt{1 - \bar{\alpha}_t}\boldsymbol{\epsilon}^t_{\boldsymbol{\Theta}}(\mathbf{z}_{t}; \mathbf{x}; \boldsymbol{\tau}_{\boldsymbol{\theta}}(\mathcal{D}_{bs};\mathcal{B}_c))}{\sqrt{\bar{\alpha}_t}})\notag\\
    &+\sqrt{1 - \bar{\alpha}_{t-1} - \sigma^2_t}\boldsymbol{\epsilon}^t_{\boldsymbol{\Theta}}(\mathbf{z}_{t}; \mathbf{x}; \boldsymbol{\tau}_{\boldsymbol{\theta}}(\mathcal{D}_{bs};\mathcal{B}_c))\notag\\
    &+ \sigma_t\boldsymbol{\epsilon}_t\label{CCAD_rec}.
\end{align}
The SDEB and SDMB consist of stacked diffusion blocks that are frozen during training. The GCEB and GCDB are composed of trainable GCBs. As shown in Figure \ref{GCB_2}, each GCB includes a self-attention block to capture contextual information and a cross-attention block to extract and fuse relevant information from the global feature bank. 
objective function of the reconstruction process in CCAD can be denoted as:
\vspace{-1mm}
\begin{align}
\mathcal{L}_{\text{CCAD(F)}}&= \mathbb{E}_{\mathbf{z}, \mathbf{x}, \boldsymbol{\epsilon},t}[\|\boldsymbol{\epsilon} - \boldsymbol{\epsilon}^t_{\boldsymbol{\Theta}}(\mathbf{z}_{t}; \textbf{x};\boldsymbol{\tau}_{\boldsymbol{\theta}}(\mathcal{D}_{bs};\mathcal{B}_c)\|_2^2].\label{obj_CCAD}
\end{align}
The whole process in CCAD can be summarized in Appendix \ref{CCAD(F)}.

\subsubsection{CCAD(C)}
For tasks with a limited global feature space, our method supports directly using $\mathcal{B}_c$ as the condition to enhance efficiency, which we denote as CCAD(C). CCAD(C) shares the same GCDM architecture as CCAD(F), with the key difference using another distinct pre-trained feature extractor to generate CFB $\mathcal{B}_c$. The objective function in CCAD(C) is changed from \eqref{obj_CCAD} to
\begin{align}
    \mathcal{L}_{\text{CCAD(C)}}&= \mathbb{E}_{\mathbf{z}, \mathbf{x}, \boldsymbol{\epsilon},t}[\|\boldsymbol{\epsilon} - \boldsymbol{\epsilon}^t_{\boldsymbol{\Theta}}(\mathbf{z}_{t}; \textbf{x};\mathcal{B}_c)\|_2^2].
\end{align}

\vspace{-3mm}
\subsubsection{CCAD(V)}
We also proposed a simplified CCAD version corresponding to the backbones in DDAD \cite{DDAD} as CCAD(V). The diffusion process is conducted directly on the pixel space by \eqref{diffusion_forward}. Therefore, batch-wise feature space $\mathcal{D}_{bs}$ is not available in the scenario and we only use $\mathcal{B}_c$ as the global feature condition.
Based on the structure proposed in \cite{UNet_DDAD}, each module is trainable and composed of ResNet Blocks and Self-attention layers. We then replace all the attention layers with our designed GCB in figure \ref{GCB_1}.
 For the sampling process, derived from \eqref{sampling}, when a target image $\bar{\mathbf{x}}_0$ is given and sample $\mathbf{x}_{t-1}$ is generated from sample $\mathbf{x}_{t}$ by
\begin{align}
    \mathbf{x}_{t-1} = \sigma_t\boldsymbol{\epsilon}_t+ &\sqrt{\bar{\alpha}_{t-1}}(\frac{\mathbf{x}_{t} - \sqrt{1 - \bar{\alpha}_t}\boldsymbol{\epsilon}^t_{\boldsymbol{\Theta}}(\mathbf{x}_{t}; \bar{\mathbf{x}}_{t}; \mathcal{B}_c)}{\sqrt{\bar{\alpha}_t}})\notag\\
    +&\sqrt{1 - \bar{\alpha}_{t-1} - \sigma^2_t}\boldsymbol{\epsilon}^t_{\boldsymbol{\Theta}}(\mathbf{x}_{t}; \bar{\mathbf{x}}_{t}; \mathcal{B}_c)\label{CCAD_V_rec}\\
\boldsymbol{\epsilon}^t_{\boldsymbol{\Theta}}(\mathbf{x}_{t}; \bar{\mathbf{x}}_{t}; \mathcal{B}_c) = & \boldsymbol{\epsilon}^t_{\boldsymbol{\Theta}}(\mathbf{x}_{t}; \mathcal{B}_c)  - w\sqrt{1- \bar{\alpha}_t}(\bar{\mathbf{x}}_{t} - \mathbf{x}_{t})\label{eq_CCAD_V} \\
\bar{\mathbf{x}}_{t} = & \sqrt{\bar{\alpha}_{t}}~\bar{\mathbf{x}}_0 + \sqrt{1 - \bar{\alpha}_t}\boldsymbol{\epsilon}^t_{\boldsymbol{\Theta}}(\mathbf{x}_{t}; \mathcal{B}_c)\label{Y_forward}
\end{align}
where $\boldsymbol{\epsilon}^t_{\boldsymbol{\Theta}}(\mathbf{x}_{t}; \mathcal{B}_c)$ is the backbone with learnable parameters $\boldsymbol{\Theta}$. The objective function in CCAD(V) is denoted by
\begin{align}
    \mathcal{L}_{\text{CCAD(V)}} = \mathbb{E}_{\mathbf{x}, \boldsymbol{\epsilon},t}[\|\boldsymbol{\epsilon} - \boldsymbol{\epsilon}^t_{\boldsymbol{\Theta}}(\mathbf{x}_{t}; \mathcal{B}_c)\|_2^2]\label{obj_CCAD_C}.
\end{align}

\subsubsection{Anomaly Detection}
Similar to the method in \cite{DiAD}, in the inference phase, given a reconstructed image $
\hat{\mathbf{x}}_0$, we compare the cosine similarity of the two images on the feature domain using a pre-trained feature extractor $\boldsymbol{\psi}$ on ImageNet \cite{ImageNet} to obtain the overall anomaly score $\mathbf{M}$. The calculation process can be denoted by
\begin{align}
    \mathbf{M} &= \sum_l \sigma_l(1 - \frac{\boldsymbol{\psi}_l(\mathbf{x}_0) \cdot \boldsymbol{\psi}_l({\hat{\mathbf{x}}_0})}{\|\boldsymbol{\psi}_l(\mathbf{x}_0)\|\boldsymbol{\psi}_l({\hat{\mathbf{x}}_0})\|\|})
\end{align}
where $\sigma_l$ is the upsampling factor and $l$ is the layer index of feature extractor $\psi$.

\section{Experimental Results and Analysis}
\subsection{Datasets}
We conduct experiments on MVTec-AD \cite{mvtec,2021mvtec}, VisA \cite{VisA},  MVTec-3D \cite{MVTec-3d}, MVTec-loco \cite{MVTec-loco}, MTD \cite{MTD} and we annotated DAGM \cite{DAGM}. For MVTec-AD, VisA, MVTec-3d and MTD datasets, we use the original training and testing set as our training and testing data respectively. For MVTec-loco, we only use the RGB training and testing images in our experiment.
\subsection{New annotations on DAGM}
The original DAGM \cite{DAGM} is a synthetic dataset on textured surfaces. It contains ten categories
including $15000$ non-defective images and $2100$ defective images. However, only rough ellipses are provided as weak labels indicating the defective areas. As a result, DAGM has consistently performed poorly on pixel-level anomaly detection across various algorithms. For a long time, the DAGM dataset has been considered unsuitable for pixel-level anomaly localization tasks. To address this issue, we re-annotated $4$ categories in the DAGM dataset (class $1$ defect, class $2$ scratch, class $7$ blur, and class $9$ spots). Compared with the previous elliptical ground truth, our provided labels feature more complicated contours, resulting in better pixel-level performance using the same algorithms. We then sampled $300$ normal training images from each fine-grained annotated category as the training data. For the testing data, we sampled $75$ normal and all defective images in each category as the testing data.
\subsection{Evaluation Metrics}
Same as the prior works, we use AUROC (Area Under the Receiver Operating Characteristic Curve), F1-score-max, and Average Precision to evaluate the performance of both class-level anomaly detection and pixel-level anomaly localization.

\begin{figure}[t]
\begin{subfigure}[t]{0.49\linewidth}
\centering
\begin{minipage}[t]{0.23\textwidth}
\centering
\includegraphics[width=1.0cm]{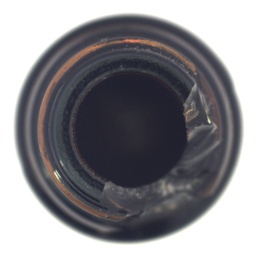}
\end{minipage}
\begin{minipage}[t]{0.23\textwidth}
\centering
\includegraphics[width=1.0cm]{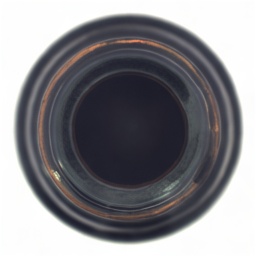}
\end{minipage}
\begin{minipage}[t]{0.23\textwidth}
\centering
\includegraphics[width=1.0cm, height=1.0cm]{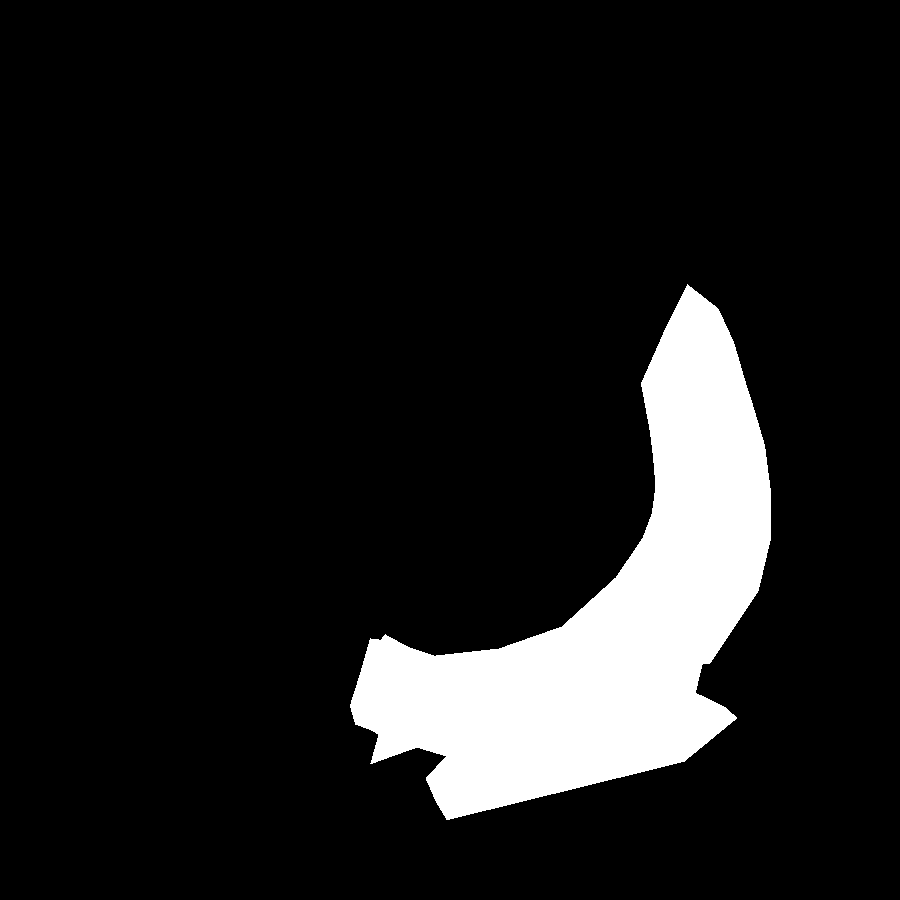}
\end{minipage}
\begin{minipage}[t]{0.23\textwidth}
\centering
\includegraphics[width=1.0cm]{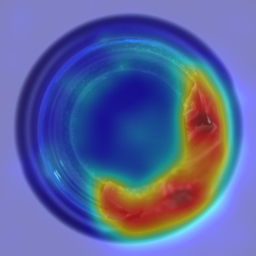}
\end{minipage}
\centering
\begin{minipage}[t]{0.23\textwidth}
\centering
\includegraphics[width=1.0cm]{exp_figure/ViSA/candle_006-input.jpg}
\end{minipage}
\begin{minipage}[t]{0.23\textwidth}
\centering
\includegraphics[width=1.0cm]{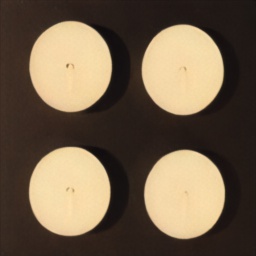}
\end{minipage}
\begin{minipage}[t]{0.23\textwidth}
\centering
\includegraphics[width=1.0cm, height=1.0cm]{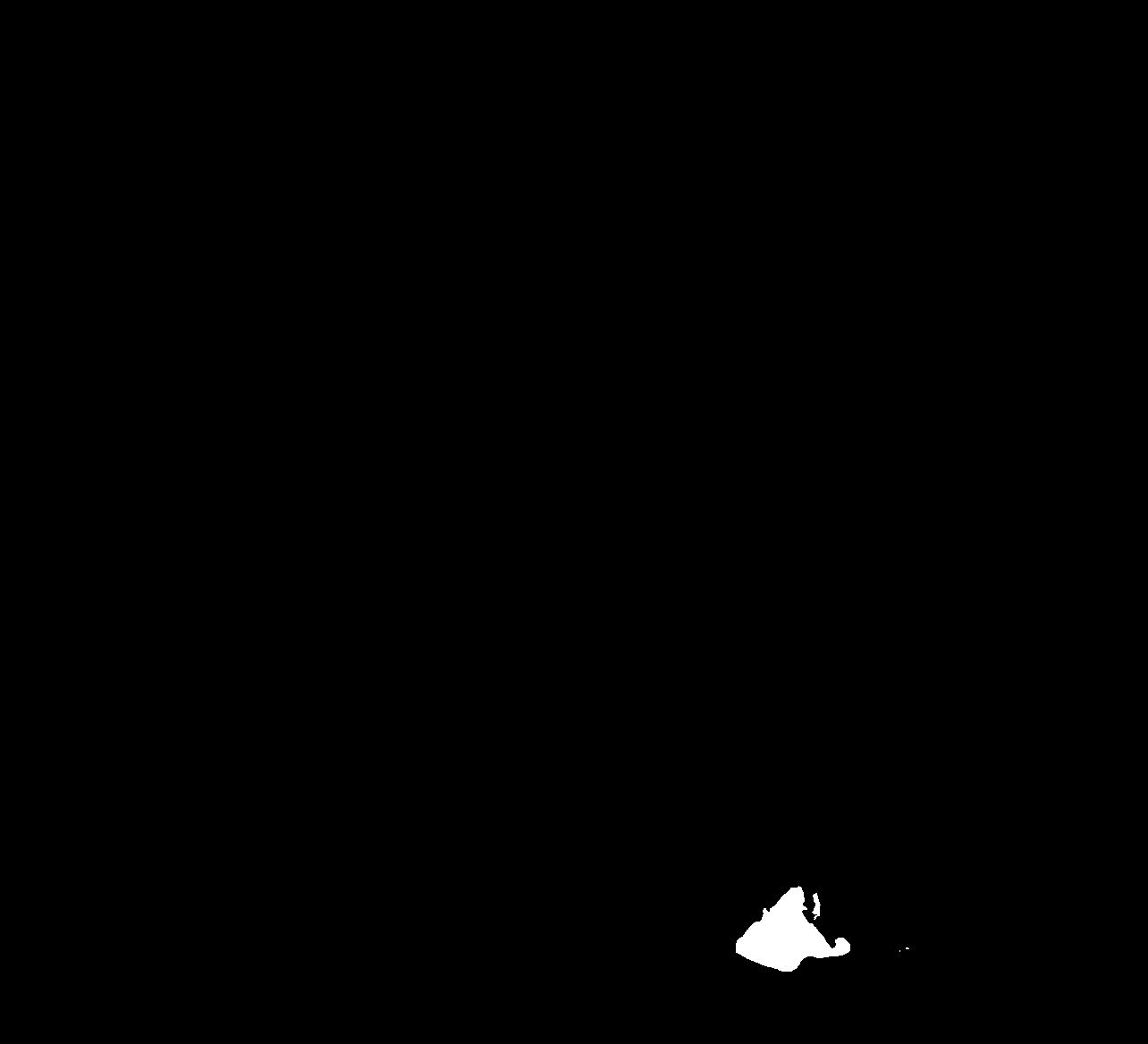}
\end{minipage}
\begin{minipage}[t]{0.23\textwidth}
\centering
\includegraphics[width=1.0cm]{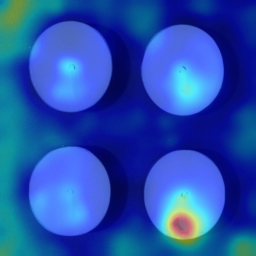}
\end{minipage}
\centering
\begin{minipage}[t]{0.23\textwidth}
\centering
\includegraphics[width=1.0cm]{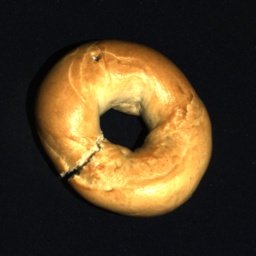}
\end{minipage}
\begin{minipage}[t]{0.23\textwidth}
\centering
\includegraphics[width=1.0cm]{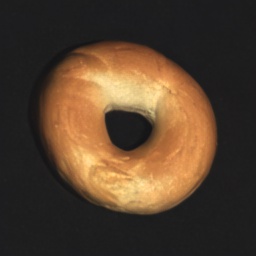}
\end{minipage}
\begin{minipage}[t]{0.23\textwidth}
\centering
\includegraphics[width=1.0cm, height=1.0cm]{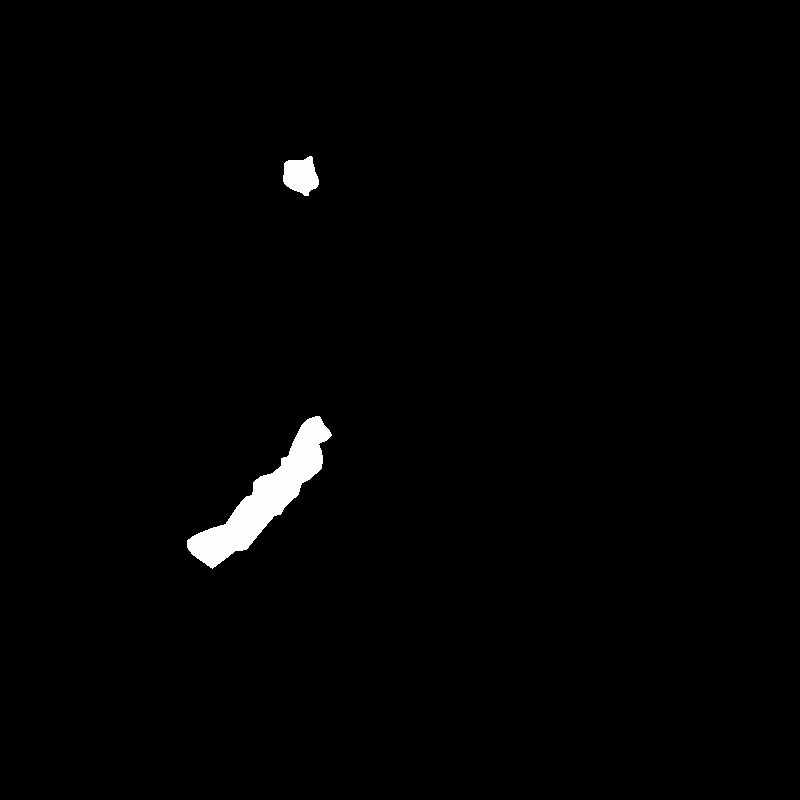}
\end{minipage}
\begin{minipage}[t]{0.23\textwidth}
\centering
\includegraphics[width=1.0cm]{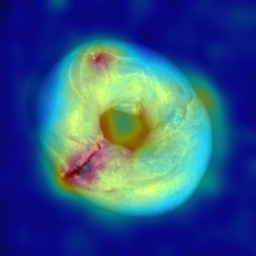}
\end{minipage}
\centering
\begin{minipage}[t]{0.23\textwidth}
\centering
\includegraphics[width=1.0cm]{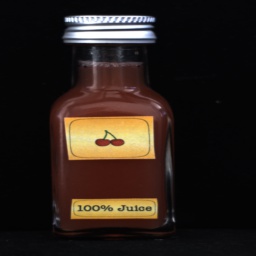}
\end{minipage}
\begin{minipage}[t]{0.23\textwidth}
\centering
\includegraphics[width=1.0cm]{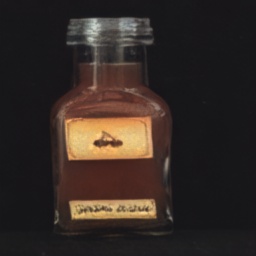}
\end{minipage}
\begin{minipage}[t]{0.23\textwidth}
\centering
\includegraphics[width=1.0cm, height=1.0cm]{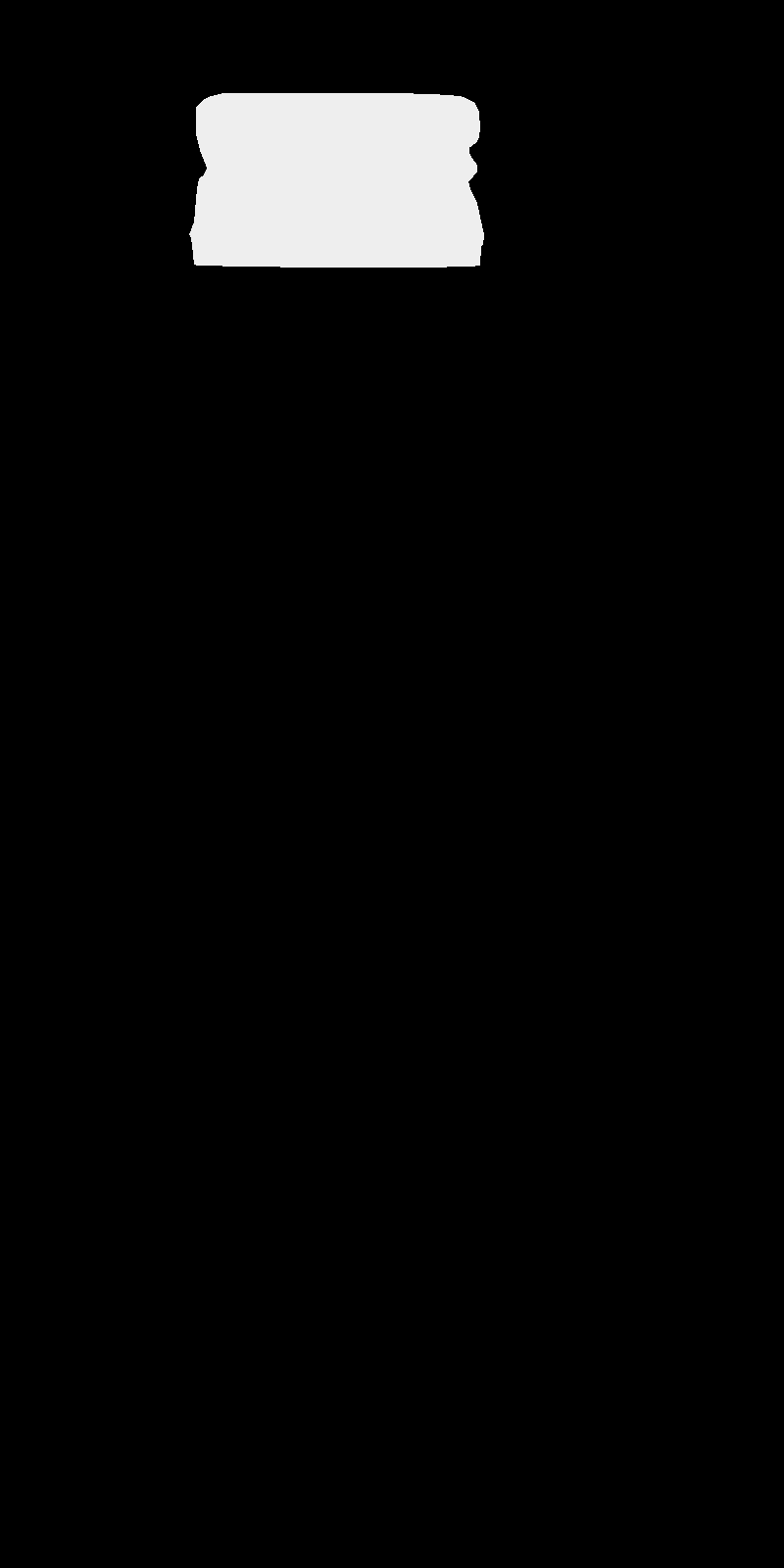}
\end{minipage}
\begin{minipage}[t]{0.23\textwidth}
\centering
\includegraphics[width=1.0cm]{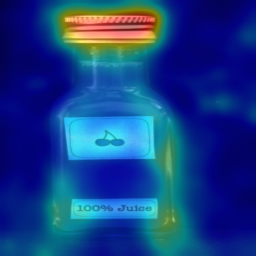}
\end{minipage}
\begin{minipage}[t]{0.23\textwidth}
Image
\end{minipage}
\begin{minipage}[t]{0.23\textwidth}
\centering
Image (rec)
\end{minipage}
\begin{minipage}[t]{0.23\textwidth}
\centering
Label
\end{minipage}
\begin{minipage}[t]{0.23\textwidth}
\centering
Heat Map
\end{minipage}
\centering
\captionsetup{justification=centering}
\subcaption[]{Visualization of CCAD on various datasets}
\label{CCAD visualization}
\end{subfigure}
\begin{subfigure}[t]{0.49\linewidth}
\centering
\begin{minipage}[t]{0.23\textwidth}
\centering
\includegraphics[width=1.0cm]{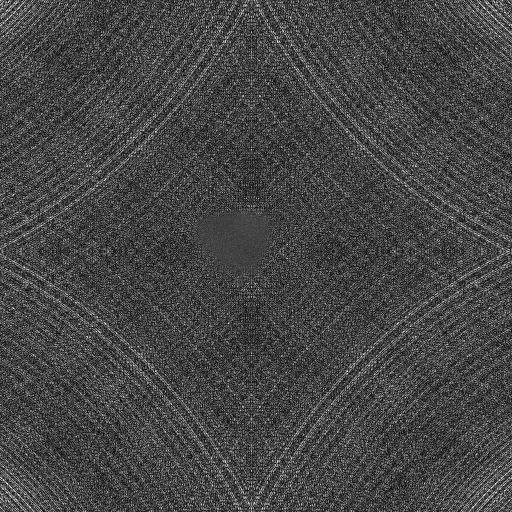}
\end{minipage}
\begin{minipage}[t]{0.23\textwidth}
\centering
\includegraphics[width=1.0cm]{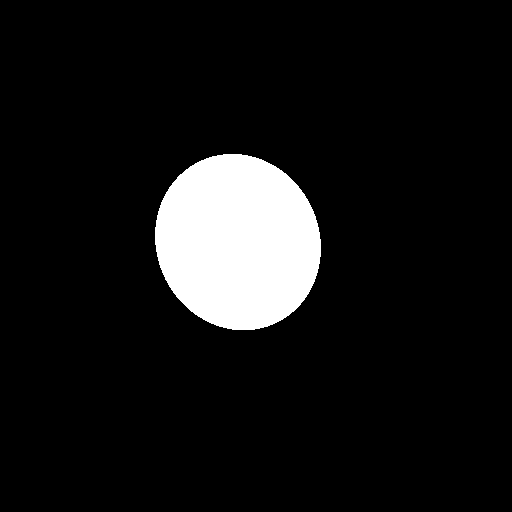}
\end{minipage}
\begin{minipage}[t]{0.23\textwidth}
\centering
\includegraphics[width=1.0cm]{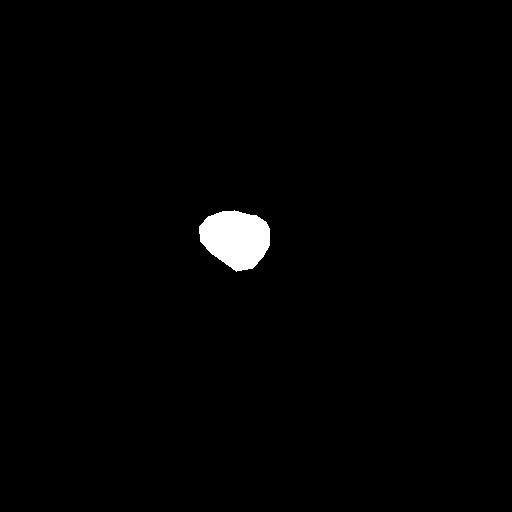}
\end{minipage}
\begin{minipage}[t]{0.23\textwidth}
\centering
\includegraphics[width=1.0cm]{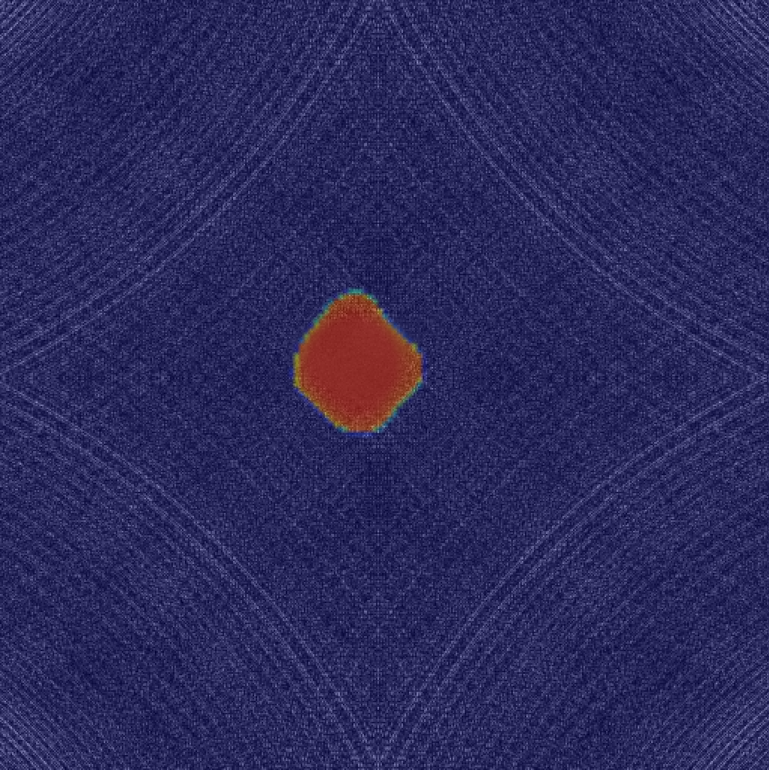}
\end{minipage}
\centering
\begin{minipage}[t]{0.23\textwidth}
\centering
\includegraphics[width=1.0cm]{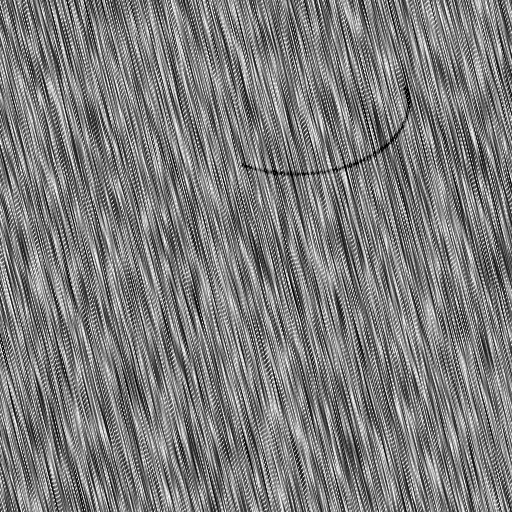}
\end{minipage}
\begin{minipage}[t]{0.23\textwidth}
\centering
\includegraphics[width=1.0cm]{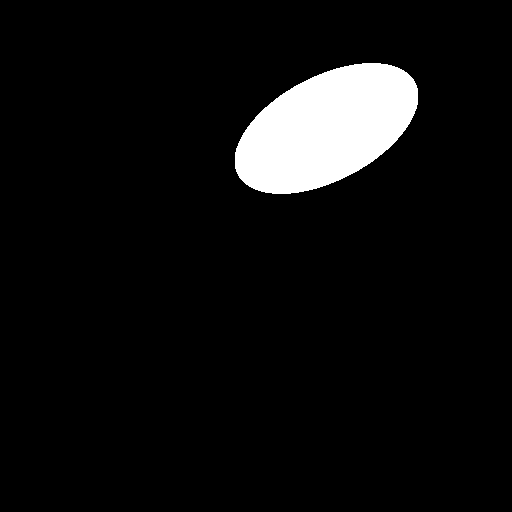}
\end{minipage}
\begin{minipage}[t]{0.23\textwidth}
\centering
\includegraphics[width=1.0cm]{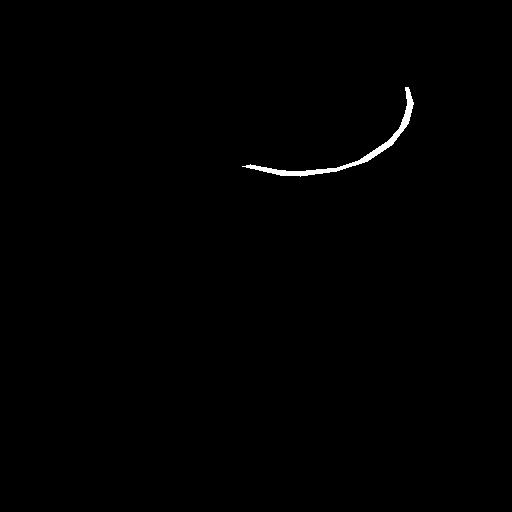}
\end{minipage}
\begin{minipage}[t]{0.23\textwidth}
\centering
\includegraphics[width=1.0cm]{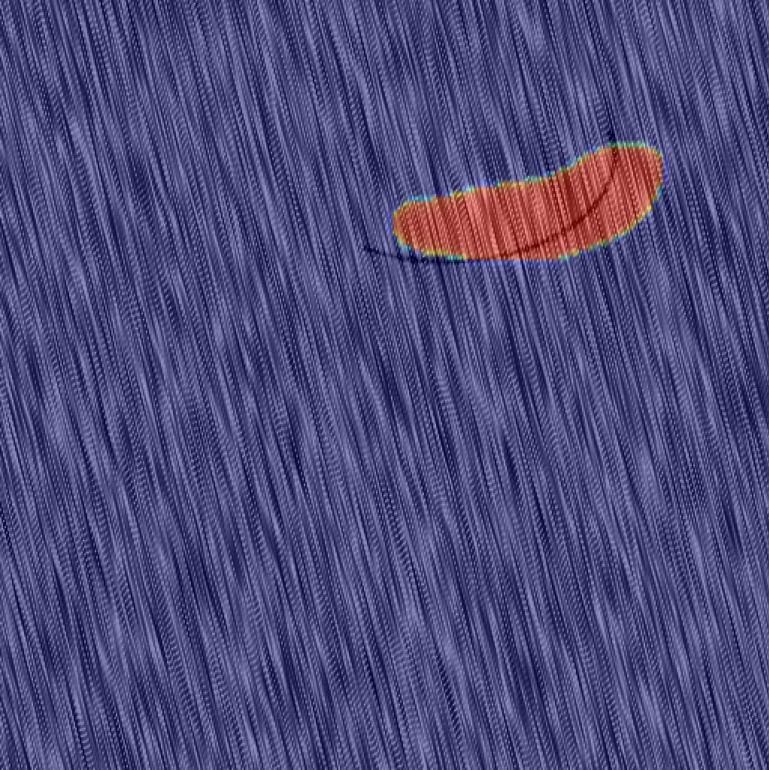}
\end{minipage}
\centering
\begin{minipage}[t]{0.23\textwidth}
\centering
\includegraphics[width=1.0cm]{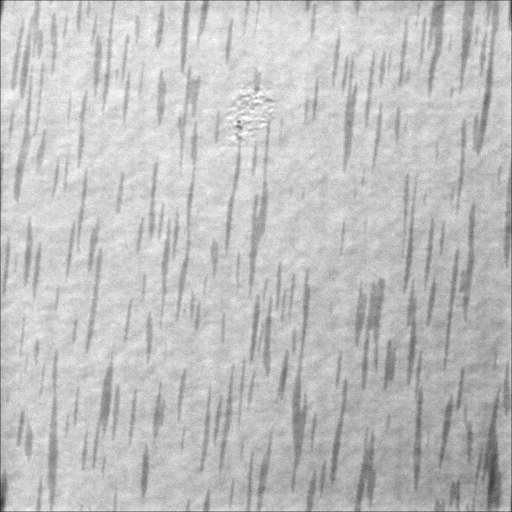}
\end{minipage}
\begin{minipage}[t]{0.23\textwidth}
\centering
\includegraphics[width=1.0cm]{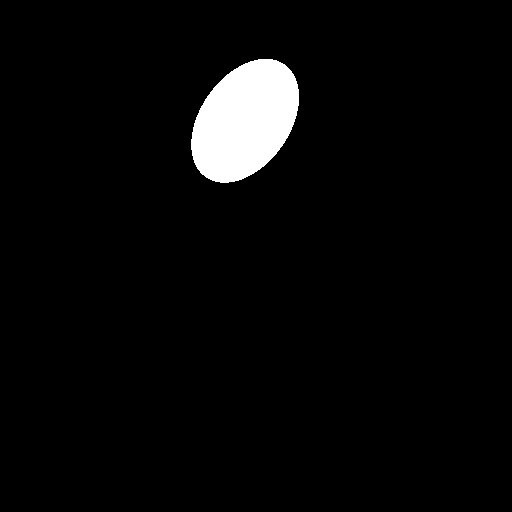}
\end{minipage}
\begin{minipage}[t]{0.23\textwidth}
\centering
\includegraphics[width=1.0cm]{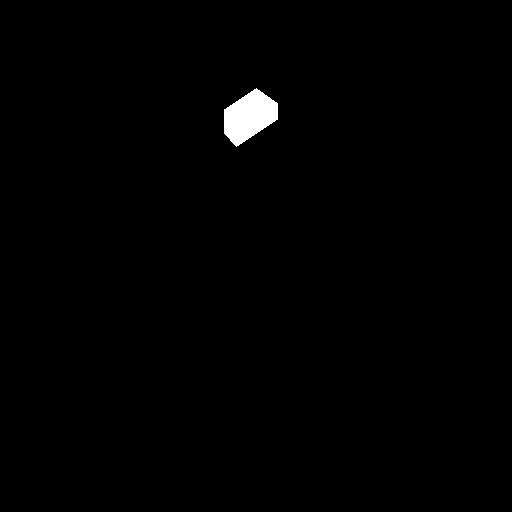}
\end{minipage}
\begin{minipage}[t]{0.23\textwidth}
\centering
\includegraphics[width=1.0cm]{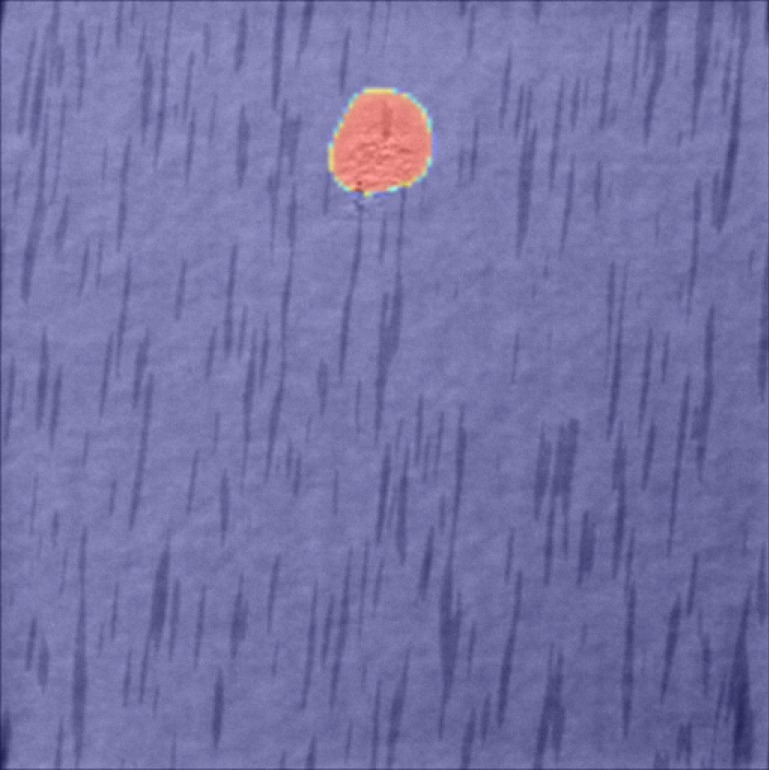}
\end{minipage}
\centering
\begin{minipage}[t]{0.23\textwidth}
\centering
\includegraphics[width=1.0cm]{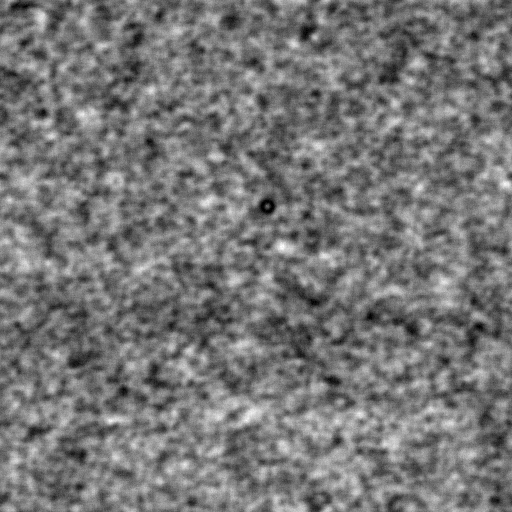}
\end{minipage}
\begin{minipage}[t]{0.23\textwidth}
\centering
\includegraphics[width=1.0cm]{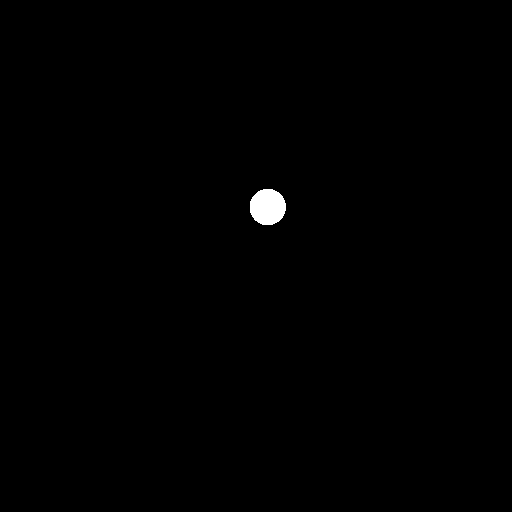}
\end{minipage}
\begin{minipage}[t]{0.23\textwidth}
\centering
\includegraphics[width=1.0cm]{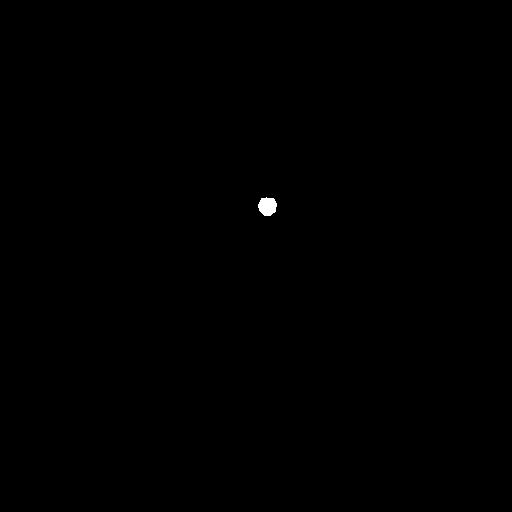}
\end{minipage}
\begin{minipage}[t]{0.23\textwidth}
\centering
\includegraphics[width=1.0cm]{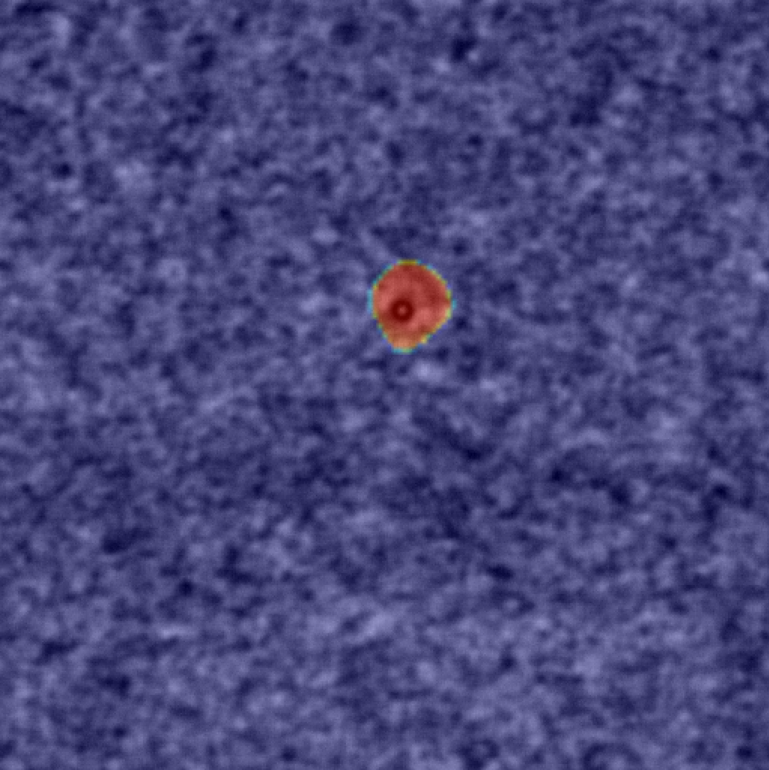}
\end{minipage}
\begin{minipage}[t]{0.23\textwidth}
Image
\end{minipage}
\begin{minipage}[t]{0.23\textwidth}
\centering
Label (old)
\end{minipage}
\begin{minipage}[t]{0.23\textwidth}
\centering
Label (new)
\end{minipage}
\begin{minipage}[t]{0.23\textwidth}
\centering
Heat Map
\end{minipage}
\centering
\captionsetup{justification=centering}
\subcaption[]{Annotation comparison on DAGM dataset}
\label{DAGM visualization}
\end{subfigure}
\vspace{-2mm}
\caption{Qualitative example visualization.}
\label{Visualization}
\end{figure}
\vspace{-2mm}

\subsection{Implement details}
We conduct each algorithm on multiple datasets with different hyperparameter settings. We also list the hyperparameter setting and source codes in Appendix \ref{hypterparameter}
\subsubsection[SPADE]{SPADE \cite{SPADE}}
We implement SPADE on MVTec dataset \cite{mvtec} using WideResNet50-2 \cite{wideresnet} as the backbone and set the $k=5$ in the k-NN algorithm.
\subsubsection[PatchCore]{PatchCore \cite{patchcore}}
We choose the encoder $ \mathcal{E}$ \cite{VQGAN} in LDM \cite{SD} as the backbone and resize all images into $256\times256$ pixels as inputs. select feature maps after the ResBlocks in the third and fourth downsampling blocks to build the global feature space. To ensure fairness, we set the number of samples in the Memory Bank in \cite{patchcore} as 1000. 
\subsubsection[CCAD(V) and DDAD]{CCAD(V) and DDAD \cite{DDAD}}
We implement DDAD and CCAD(V) on all the datasets. We completed the three steps including U-Nets training, feature extractor fine-tuning, and anomaly detection for each category respectively. To ensure a fair comparison, for a certain category, the two algorithms share the same training epochs, learning rate, and fine-tuned feature extractor. The only difference is the introduction of CFB $\mathcal{B}_c$ as the conditioning input in our CCAD(V). The samples in the $\mathcal{B}_c$ are generated by the feature maps from the second and the third stage of the residual blocks in a WideResNet50 \cite{wideresnet} pre-trained on ImageNet \cite{ImageNet}.
\subsubsection[CCAD(C \& F) and DiAD]{CCAD(C \& F) and DiAD \cite{DiAD}}
We also conduct our CCAD (C \& F) and DiAD on all the datasets we listed. For all the backbones in these three algorithms, we use the pre-trained Diffusion v1.5 model for initialization and we implement anomaly detection and localization on a ResNet50 pre-trained on ImageNet which is also shared among the three algorithms to ensure fairness. In CCAD(F), to mask sure that elements in batch-wise feature space $\mathcal{D}_{bs}$ and CFB $\mathcal{B}_c$ come from the same distribution, we choose the same feature maps as used in our PatchCore experiments to generate $\mathcal{D}_{bs}$ and $\mathcal{B}_c$. In CCAD(C). Since only $\mathcal{B}_c$ is introduced as the condition, we choose the feature maps from the second and the third stage of the residual blocks in a WideResNet50 \cite{wideresnet} pre-trained on ImageNet \cite{ImageNet}.
\begin{table*}[t]
\centering
\begin{minipage}[t]{1\textwidth}
\centering
\captionsetup{justification=centering,margin=0.1cm}
\caption[AUROC]{AUROC(Class-level; Pixel-level) comparison with SOTA methods on MVTec-AD \cite{mvtec}. We highlighted the best result(s) in bold.}\label{SOTA_MVTEC_cls}
\vspace{-0.2cm}
\begin{adjustbox}{max width=1\textwidth}
\begin{tabular}{c|cccc|ccc}
\hline 
\multirow{2}*{\diagbox[height=2\line]{Class}{AUC}{Algorithm}}&\multicolumn{4}{c}{Single class based}&\multicolumn{3}{|c}{Multi-class based}\\
\cline{2-8}
~ &SPADE&PatchCore&DDAD&CCAD(V)&DiAD&CCAD(C)&CCAD(F)\\
\hline  
\makecell{Bottle}&0.972; 0.562&\textbf{0.995}; \textbf{0.979}&0.816; 0.878&0.975; 0.939&0.996; 0.984&1.000; 0.986&0.998; \textbf{0.986}\\
\hline  
\makecell{Cable}&0.791; 0.654&0.664; 0.906&0.947; \textbf{0.965}&\textbf{0.959}; 0.960&0.850; 0.883&0.832; 0.900&\textbf{0.904}; \textbf{0.929}\\
\hline  
\makecell{Capsule}&0.897; 0.638&0.852; 0.898&0.909; 0.928&\textbf{0.943}; \textbf{0.961}&0.891; 0.963&0.916; 0.959&0.899; \textbf{0.965}\\
\hline  
\makecell{Carpet}&0.928; 0.633&\textbf{0.941}; \textbf{0.971}&0.877;  0.926&0.898; 0.941&0.990; 0.982&0.993; 0.987&\textbf{0.992}; \textbf{0.987}\\
\hline 
\makecell{Grid}&0.471; 0.566&0.769; 0.978&0.999; \textbf{0.993}&\textbf{1.000}; \textbf{0.993}&0.949; 0.947&0.945; 0.930&\textbf{0.995}; \textbf{0.990}\\
\hline
\makecell{Hazelnut}&0.881; 0.830&0.937; 0.974&0.937; 0.973&\textbf{0.941}; \textbf{0.979}&0.971; 0.973&0.972; 0.972&\textbf{0.977}; \textbf{0.981}\\
\hline
\makecell{Leather}&0.954; 0.615&\textbf{0.988}; \textbf{0.994}&0.907; 0.984&0.928; 0.987&\textbf{1.000}; \textbf{0.991}&1.000; 0.979&\textbf{1.000}; 0.990\\
\hline
\makecell{Metal Nut}&0.710; 0.509&0.727; 0.964&0.991; 0.982&\textbf{1.000}; \textbf{0.985}&\textbf{0.987}; \textbf{0.978}&0.971; 0.978&0.978; 0.971\\
\hline
\makecell{Pill}&0.803; 0.647&0.864; 0.967&0.957; 0.972&\textbf{0.965}; \textbf{0.974}&0.911; 0.962&0.942; 0.964&0.942; 0.959\\
\hline
\makecell{Screw}&0.667; 0.583&0.543; 0.967&0.964; \textbf{0.992}&\textbf{0.967}; \textbf{0.992}&\textbf{0.879}; 0.969&0.881; 0.975&0.864; \textbf{0.976}\\
\hline
\makecell{Tile}&0.965; 0.632&0.933; 0.953&\textbf{1.000}; \textbf{0.979}&\textbf{1.000}; \textbf{0.979}&0.965; 0.926&0.966; 0.923&0.983; 0.929\\
\hline
\makecell{Toothbrush}&0.889; 0.568&0.917; \textbf{0.986}&0.981; 0.985&\textbf{1.000}; 0.984&0.994; \textbf{0.990}& 0.961; 0.989&0.975; 0.962\\
\hline
\makecell{Transistor}&0.903; 0.507&0.778; 0.756&0.947; 0.880&\textbf{0.960}; \textbf{0.894}&0.945; 0.899& 0.992; 0.955&0.971; \textbf{0.921}\\
\hline
\makecell{Wood}&0.959; 0.644&0.968; 0.941&0.992; \textbf{0.944}&\textbf{0.998}; 0.931&0.982; 0.918&0.987; 0.930&\textbf{0.987}; 0.932\\
\hline
\makecell{Zipper}&0.966; 0.415&\textbf{0.995}; \textbf{0.987}&0.979; 0.955&0.987; 0.975&0.938; 0.948&0.933; 0.960&\textbf{0.954}; 0.959\\
\hline
\makecell{mean}&0.850; 0.602&0.858; 0.948&0.943; 0.956&\textbf{0.968}; \textbf{0.965}&0.950; 0.954& 0.953; 0.959&\textbf{0.961}; \textbf{0.962}\\
\hline
\end{tabular}
\end{adjustbox}
\end{minipage}
\end{table*}

\subsection{Comparison with the state-of-the-art}
Shown in table \ref{SOTA_MVTEC_cls}, \ref{SOTA_all_data} and \ref{SOTA_DAGM_cls}, we compare our CCAD with SOTA methods on all the datasets we listed. Our algorithm demonstrates a performance advantage across various datasets. From table \ref{SOTA_MVTEC_cls}, we found that although PatchCore still performs well on several categories with a memory bank including as few as $1000$ samples, its AUC decreases greatly on many categories and datasets. It indicates that for an unsupervised learning anomaly detection approach based on global representation, the performance is highly dependent on the size of the global feature bank and whether there is a domain gap in the pre-trained feature extractor. CCAD(V) outperforms DDAD in most categories, both in class-level AUC and pixel-level AUC. Comparing with DiAD, the UNet backbone in DDAD is relatively simple and there is no additional conditional mechanism designed in the UNet backbone. As a result, the GCB in CCAD(V) makes a significant contribution to the reconstruction of normal images, which enables CCAD(V) to comprehensively outperform DDAD. We found that CCAD(C) even outperforms CCAD(F) in some categories, which is likely due to the reason that the global features extracted by the pre-trained WideResNet50 are more representative compared to those extracted by the encoder in LDM.
\begin{table*}[t]
\centering
\begin{minipage}[t]{0.49\textwidth}
\centering
\captionsetup{justification=centering,margin=0.1cm}
\caption{Average AUROC(Class-level; Pixel-level) comparison with SOTA methods on other datasets.}\label{SOTA_all_data}
\vspace{-0.2cm}
\begin{adjustbox}{max width=1.\textwidth}
\renewcommand{\arraystretch}{2}
\begin{tabular}{c|ccc|ccc}
\hline 
Dataset&PatchCore&DDAD&CCAD(V)&DiAD&CCAD(C)&CCAD(F)\\
\hline  
\makecell{VisA}&0.761; 0.923&0.963; \textbf{0.958}&\textbf{0.978}; 0.957&\textbf{0.742}; 0.898&0.651; \textbf{0.906}&0.735; 0.895\\
\hline  
\makecell{MVTec-3d}&0.67; 0.697&0.755; 0.890&\textbf{0.779}; \textbf{0.935}&\textbf{0.709}; 0.957&\textbf{0.709}; \textbf{0.971}&0.683; 0.938\\
\hline
\makecell{MVTec-loco}&0.654; 0.643&0.886; 0.678&\textbf{0.897}; \textbf{0.690}&0.665; 0.714&\textbf{0.671}; \textbf{0.718}&0.670; 0.715\\
\hline
\makecell{MTD}&0.575; 0.5895&\textbf{0.934}; 0.719&0.897; \textbf{0.755}&0.966; 0.820&0.959; \textbf{0.826}&\textbf{0.968}; 0.818\\
\hline
\end{tabular}
\end{adjustbox}
\end{minipage}
\centering
\vspace{-0.1cm}
\begin{minipage}[t]{0.49\textwidth}
\captionsetup{justification=centering,margin=0.1cm}
\vspace{0.1cm}
\caption{AUROC(Class-level; Pixel-level) comparison with SOTA methods on our new annotated DAGM2007 dataset.}\label{SOTA_DAGM_cls}
\vspace{-0.1cm}
\begin{adjustbox}{max width=1.\textwidth}
\renewcommand{\arraystretch}{1.2}
\begin{tabular}{c|cc|ccc}
\hline 
Class&DDAD&CCAD(V)&DiAD&CCAD(C)&CCAD(F)\\
\hline  
\makecell{class 1}&0.646; 0.882&\textbf{0.669}; \textbf{0.886}&0.588; 0.789&0.584; 0.806&\textbf{0.613}; \textbf{0.830}\\
\hline  
\makecell{class 2}&\textbf{1.000}; 0.992&\textbf{1.000}; \textbf{0.993}&0.937; 0.970&0.988; 0.991&\textbf{0.991}; \textbf{0.992}\\
\hline  
\makecell{class 7}&0.973; 0.991&\textbf{0.985};\textbf{0.992}&0.586; 0.787&0.627; 0.799&\textbf{0.747}; \textbf{0.899}\\
\hline  
\makecell{class 9}&0.907; 0.973&\textbf{0.925}; \textbf{0.982}&0.941; 0.993&0.910; 0.995&\textbf{0.947}; \textbf{0.996}\\
\hline 
\makecell{mean}&0.882; 0.960&\textbf{0.895}; \textbf{0.963}&0.763; 0.885&0.777; 0.898&\textbf{0.824}; \textbf{0.929}\\
\hline
\end{tabular}
\end{adjustbox}
\end{minipage}
\end{table*}

\begin{table}[t]
\centering
\begin{minipage}[t]{0.49\textwidth}
\captionsetup{justification=centering,margin=0.1cm}
\caption{AUROC(Class-level; Pixel-level) on MVTec with different $\xi$ in CCAD(V).}\label{MVTEC_CCAD_DDAD}
\vspace{-0.2cm}
\begin{adjustbox}{max width=1\textwidth}
\begin{tabular}{c|ccc}
\hline 
\multirow{2}*{\diagbox[height=2\line]{Class}{AUC}{Algorithm}}&DDAD&\multicolumn{2}{c}{CCAD(V)}\\
\cline{2-4}
~ &$\xi = 0$&$\xi = 10$&$\xi = 200$\\
\hline  
\makecell{bottle}&0.816; 0.878&\textbf{0.976}; \textbf{0.939}&0.975; \textbf{0.939}\\
\hline  
\makecell{cable}&0.947; 0.965&0.960; \textbf{0.966}&\textbf{0.959}; 0.960\\
\hline  
\makecell{capsule}&0.909; 0.928&\textbf{0.943}; \textbf{0.961}&\textbf{0.943}; \textbf{0.961}\\
\hline  
\makecell{carpet}&0.877; 0.926&0.891; 0.928&\textbf{0.898}; \textbf{0.941}\\
\hline 
\makecell{grid}&0.999; \textbf{0.993}&\textbf{1.000}; \textbf{0.993}&\textbf{1.000}; \textbf{0.993}\\
\hline
\makecell{hazelnut}&0.937; 0.973&\textbf{0.982}; 0.971&0.941; \textbf{0.979}\\
\hline
\makecell{leather}&0.907; 0.984&\textbf{0.963}; \textbf{0.989}&0.928; 0.987\\
\hline
\makecell{metal}&0.991; 0.982&0.999; 0.982&\textbf{1.000}; \textbf{0.985}\\
\hline
\makecell{pill}&0.957; 0.972&\textbf{0.967}; \textbf{0.987}&0.965; 0.974\\
\hline
\makecell{screw}&0.964; 0.992&\textbf{0.975}; \textbf{0.993}&0.967; 0.992\\
\hline
\makecell{tile}&\textbf{1.000}; \textbf{0.979}&\textbf{1.000}; \textbf{0.979}&\textbf{1.000}; \textbf{0.979}\\
\hline
\makecell{toothbrush}&0.981; \textbf{0.985}&0.994; 0.984&\textbf{1.000}; 0.984\\
\hline
\makecell{transistor}&0.947; 0.880&0.952; 0.893&\textbf{0.960}; \textbf{0.894}\\
\hline
\makecell{wood}&0.992; \textbf{0.944}&0.997; 0.930&\textbf{0.998}; 0.931\\
\hline
\makecell{zipper}&0.979; 0.955&0.979; 0.973&\textbf{0.987}; \textbf{0.975}\\
\hline
\makecell{mean}&0.943; 0.956&\textbf{0.971}; 0.963&0.968; \textbf{0.965}\\
\hline
\end{tabular}
\end{adjustbox}
\end{minipage}
\end{table}

\begin{figure}[t]
\begin{minipage}[c]{0.49\textwidth}
    \centering
    \begin{subfigure}[t]{0.49\textwidth}
    \centering
    \includegraphics[width=4.0cm]{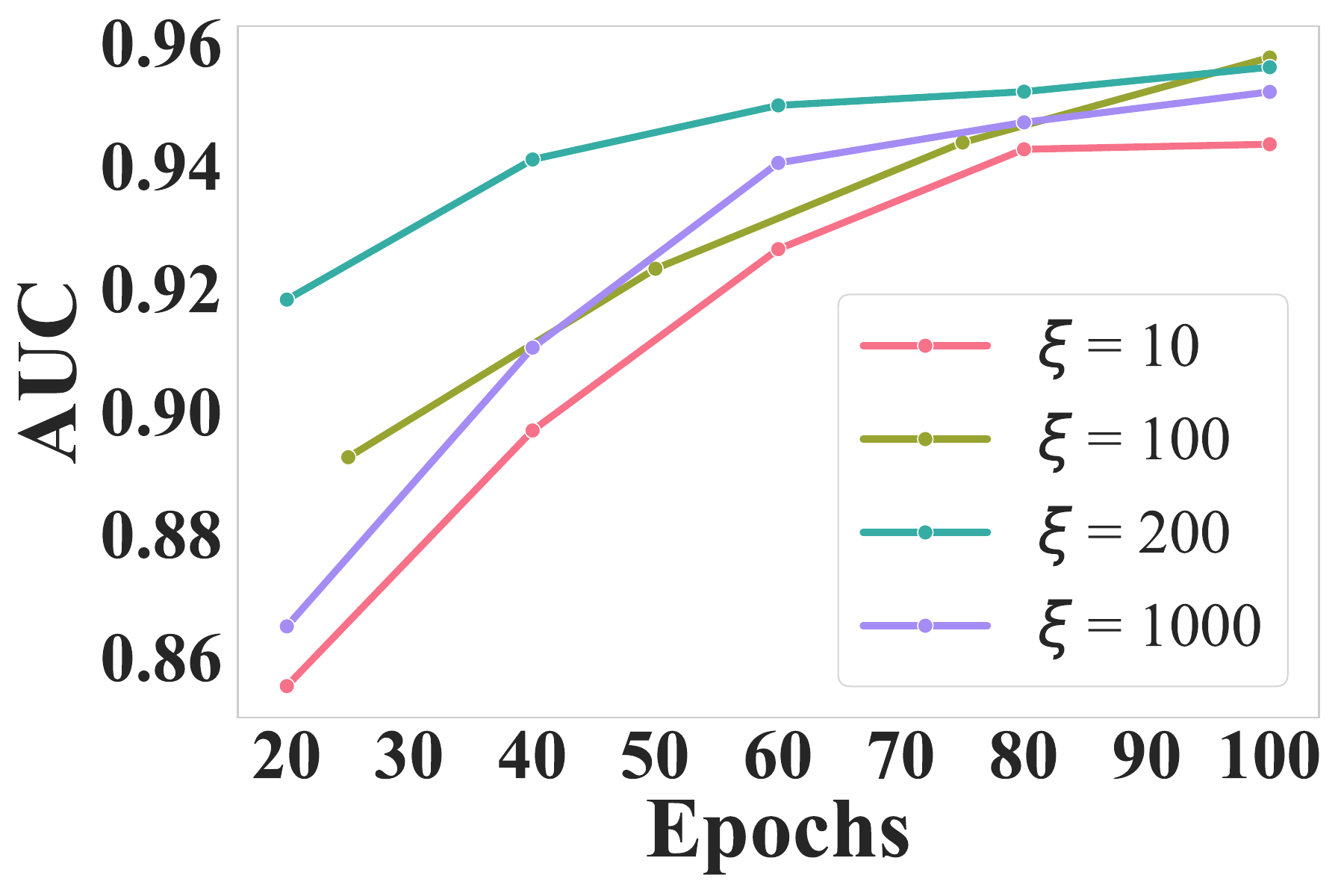}
    \centering
    \captionsetup{justification=centering}
    \subcaption{Class-wise AUC on MVTec with different $\xi$}\label{ablation_cls_CCAD_C}
    \end{subfigure}
    \begin{subfigure}[t]{0.49\textwidth}
    \centering
    \includegraphics[width=4.0cm]{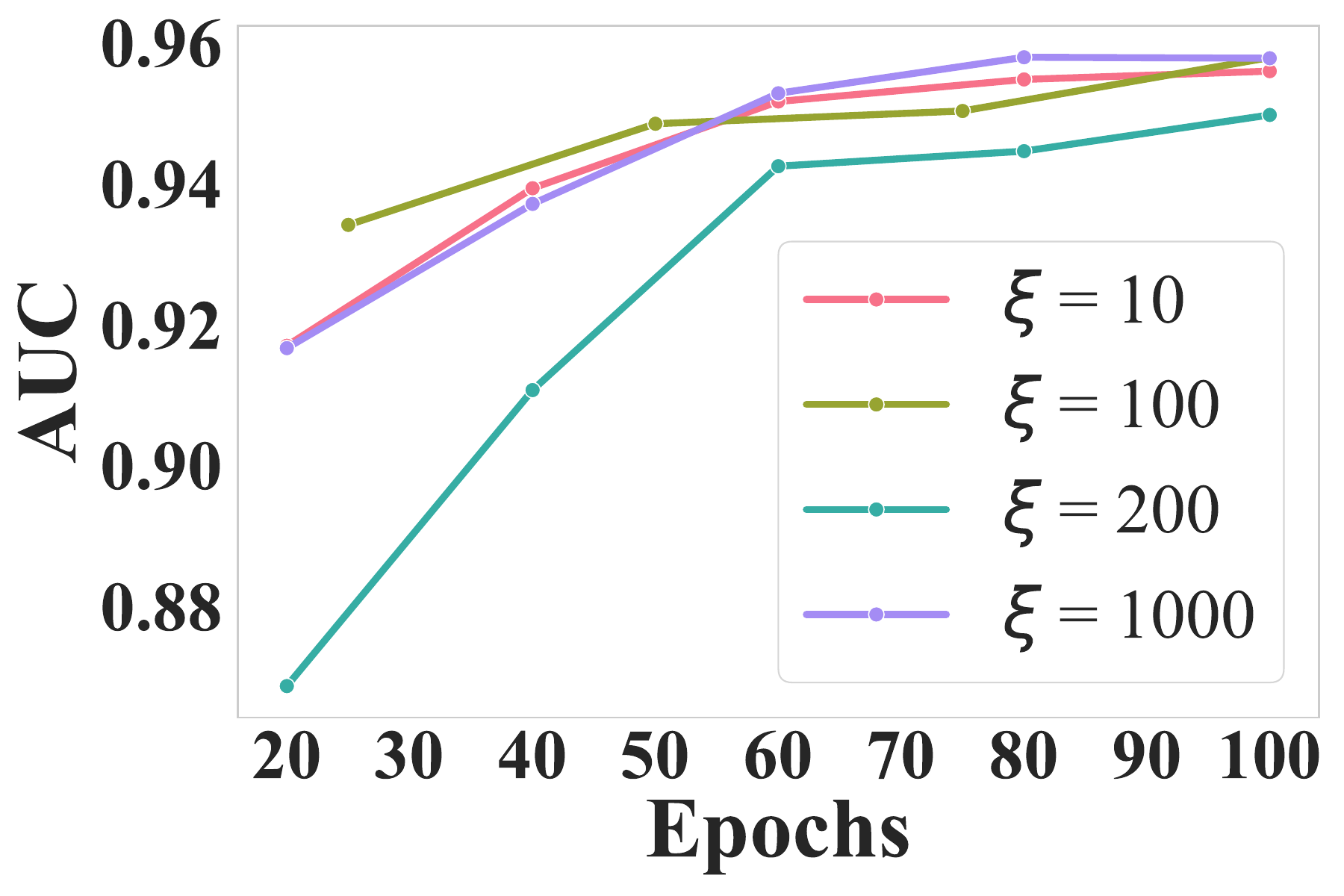}
    \captionsetup{justification=centering}
    \subcaption{Pixel-wise AUC on MVTec with different $\xi$}\label{ablation_pix_CCAD_C}
    \end{subfigure}
    \centering
    \captionsetup{justification=centering,margin=0.1cm}
    \vspace{-2mm}
     \caption{AUC on MVTec with different $\xi$.}
    \label{diff_dim}
\end{minipage}
\end{figure}

\vspace{-2mm}
\subsection{Ablation Studies}
Shown in table \ref{MVTEC_CCAD_DDAD}, we compare CCAD(V) with different samples $\xi$ in CFB $\mathcal{B}_c$. Surprisingly, even with a $\xi$ of only 10, CCAD already outperforms DDAD, and as the dimension increases, there is no clear upward trend in AUC. This is likely because the GCBs, as learnable blocks embedding in the U-Net at each layer, can selectively utilize representative global feature information through the cross-attention layer during training. Even $\xi$ is small, the available information learn by GVBs is already sufficient for nominal image reconstruction. Additionally, since our algorithm performs well on MVTec, the potential for AUC improvement as dimensionality increases is inherently limited. We also compare CCAD(C) with different $\xi$ on MVTec. Shown in figure \ref{ablation_cls_CCAD_C} and \ref{ablation_pix_CCAD_C}, when $\xi$ is set to 10, the AUC is at its lowest, while the best AUC is achieved with $\xi$ of 100 and 200. However, no significant AUC improvement is observed when $\xi$ is set to 1000. This indicates that with $\xi = 100$, the $\mathcal{B}_c$ is already sufficient to aid in image reconstruction.
\subsection{Faster Convergence in CCAD}
We also compare the convergence speed of DDAD, DiAD, and CCAD in table \ref{SOTA_MVTEC_cls}. CCAD significantly outperforms DDAD and DiAD in speed while achieving a relatively high AUC performance. A significant reason for this phenomenon is that, in the early stages of model training, the global information from the condition provides a great contribution to model learning, and enables the reconstructed images to retain features from the training data. We demonstrate more experimental results and analysis in the appendix

\subsection{New annotated labels}
Shown in figure \ref{DAGM visualization}, our new annotated ground truth is qualitatively more aligned with the anomaly maps in terms of both outlines and locations indicating that our annotated labels are more proper and accurate. We demonstrate the results and analysis in the appendix.
\vspace{-2mm}
\section{Conclusion}
In this paper, we proposed CCAD: Anomaly Detection Conditioned on Compressed Global Feature Space that uses a two-stage coarse-to-fine approach converting global features as conditions for image reconstruction on anomaly detection tasks. Extensive experimental results show that CCAD exhibits a notable advantage over SOTA algorithm across multiple datasets. We also re-annotated the DAGM2007 dataset \cite{DAGM}, providing more reliable and accurate labels. 

\bibliographystyle{named}
\bibliography{ijcai25}
\clearpage
\appendix
\begin{center}
	\Large \textbf{Supplementary Material} \\
\end{center}
\nomenclature{\textbf{CFB}, $\mathcal{B}_c$}{ Coarse Feature Bank}
\nomenclature{\textbf{FFB}, $\mathcal{B}_f$}{ Fine Feature Bank}
\nomenclature{$\textbf{M}$}{Anomaly Score}
\nomenclature{$t$}{time step in diffusion process}
\nomenclature{$\textbf{v}$}{Samples in Feature Banks}
\nomenclature{$\textbf{x}/\mathcal{X}$}{Training / Testing data (images)}
\nomenclature{$\bar{\textbf{x}}$}{Target image in DDAD and CCAD(V)}
\nomenclature{$\textbf{z}$}{Training / Testing input in latent space}
\nomenclature{$\mathcal{D} / \mathcal{D}_{bs}$}{Global Feature Space / Batch-wise Feature Space}
\nomenclature{$\mathcal{E}$}{Pre-trained latent encoder in LDM}
\nomenclature{$\mathcal{F}$}{Pre-trained visual encoder}
\nomenclature{$\mathcal{L}(.)$}{Objective function}
\nomenclature{$\boldsymbol{\epsilon}$}{Noise in diffusion process}
\nomenclature{$\boldsymbol{\epsilon}_{\boldsymbol{\Theta}}$}{Diffusion model trainable backbones}
\nomenclature{$\sigma$}{upsamling factor}
\nomenclature{$\xi$}{Number of samples in $\mathcal{B}_c$}
\nomenclature{$\psi$}{Pre-trained feature extractor from anomaly detection}
\nomenclature{\textbf{FCM}, $\boldsymbol{\tau}_{\boldsymbol{\theta}}$}{Fine Compression Module}
\nomenclature{\textbf{GCB}}{Global feature Conditioned Block}
\nomenclature{\textbf{GCDM}}{Global feature Conditioned Diffusion Module}
\nomenclature{\textbf{GCEB}}{Global feature Conditioned Encoder Blocks}
\nomenclature{\textbf{GCDB}}{Global feature Conditioned Decoder Blocks}
\nomenclature{\textbf{SDEB}}{Stable Diffusion Encoder Blocks}
\nomenclature{\textbf{SDMB}}{Stable Diffusion Midddle Blocks}
\printnomenclature[0.8in]

\section{Preliminary of Diffusion Model}
In this section, we introduce some preliminary knowledge related to diffusion models, which serves as the foundation for the equations we derived. For an input image $\textbf{x}_0$, the diffusion process of DDPM \cite{DDPM} can be denoted by
\begin{align}
    \mathbf{x}_t = \sqrt{\bar{\alpha}_{t}}~\mathbf{x}_0 + \sqrt{1 - \bar{\alpha}_t}\boldsymbol{\epsilon},~\bar{\alpha}_t = \prod_{i = 1}^t\alpha_i,~\boldsymbol{\epsilon} \sim \mathcal{N}(0, \boldsymbol{I})\label{diffusion_forward}
\end{align}
where $\alpha_i = 1 - \beta_i$ and $\{\beta_i | i=1, \dots, t\}$ is a pre-defined variance schedule.
In DDIM \cite{DDIM}, for the sampling process, a sample $\mathbf{x}_{t-1}$ is generated from sample $\mathbf{x}_{t}$ by
\begin{align}
    \mathbf{x}_{t-1} = &\sqrt{\bar{\alpha}_{t-1}}(\frac{\mathbf{x}_{t} - \sqrt{1 - \bar{\alpha}_t}\boldsymbol{\epsilon}^t_{\boldsymbol{\Theta}}(\mathbf{x}_{t})}{\sqrt{\bar{\alpha}_t}})\notag\\
    &+\sqrt{1 - \bar{\alpha}_{t-1} - \sigma^2_t}\boldsymbol{\epsilon}^t_{\boldsymbol{\Theta}}(\mathbf{x}_{t}) + \sigma_t\boldsymbol{\epsilon}_t\label{sampling}
\end{align}
where $\sigma_t$ is the value determining the 
randomness in the sampling process.

\section{Frameworks of CCAD(C \& V)}
\begin{figure}[H]
\begin{minipage}[c]{0.49\textwidth}
    \centering
    \begin{subfigure}[t]{1\textwidth}
    \centering
    \includegraphics[width=1\textwidth]{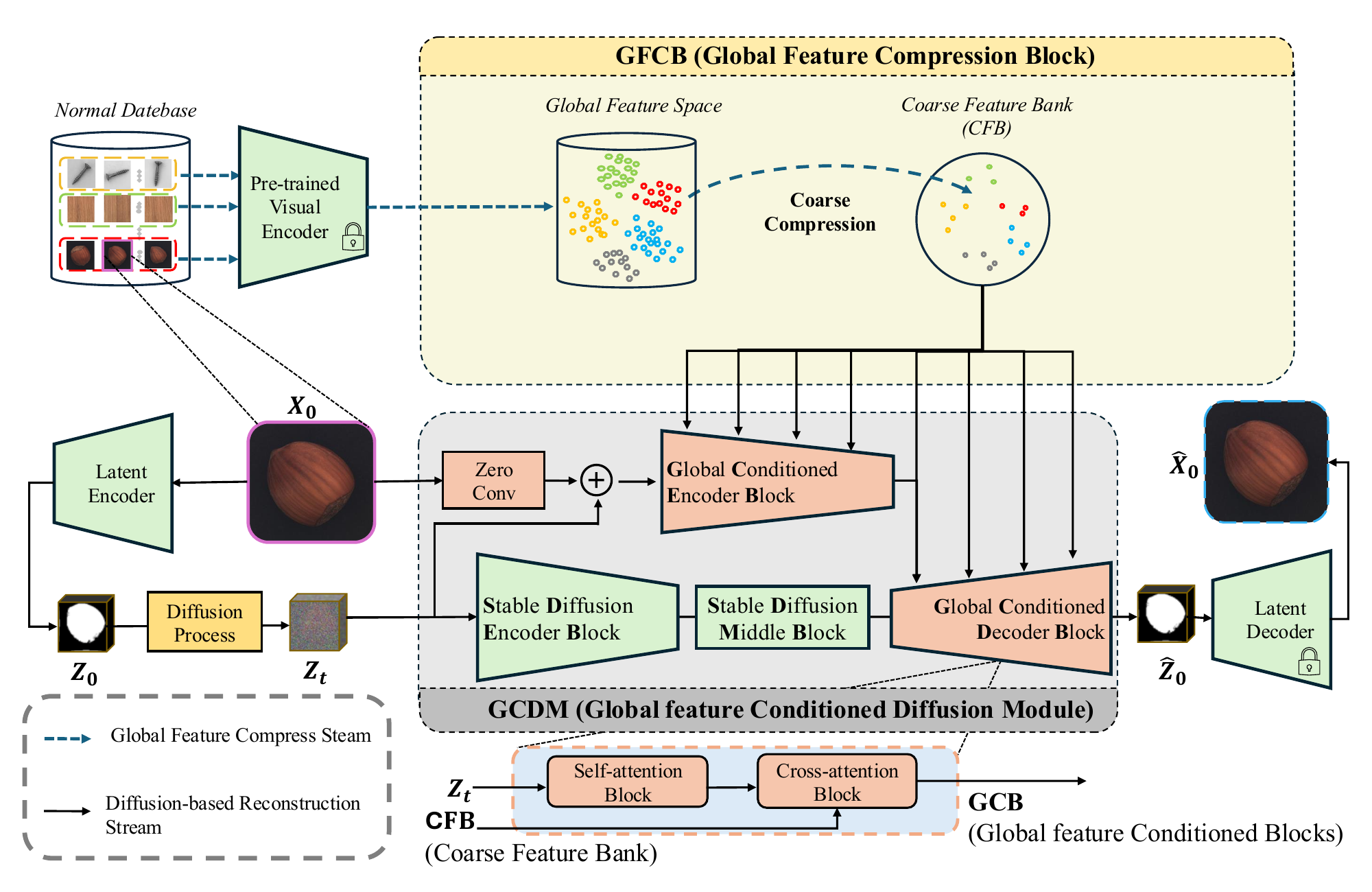}
    \subcaption{CCAD(C)}\label{CCAD_C}
    \end{subfigure}
    \begin{subfigure}[t]{1\textwidth}
    \centering
    \includegraphics[width=1\textwidth]{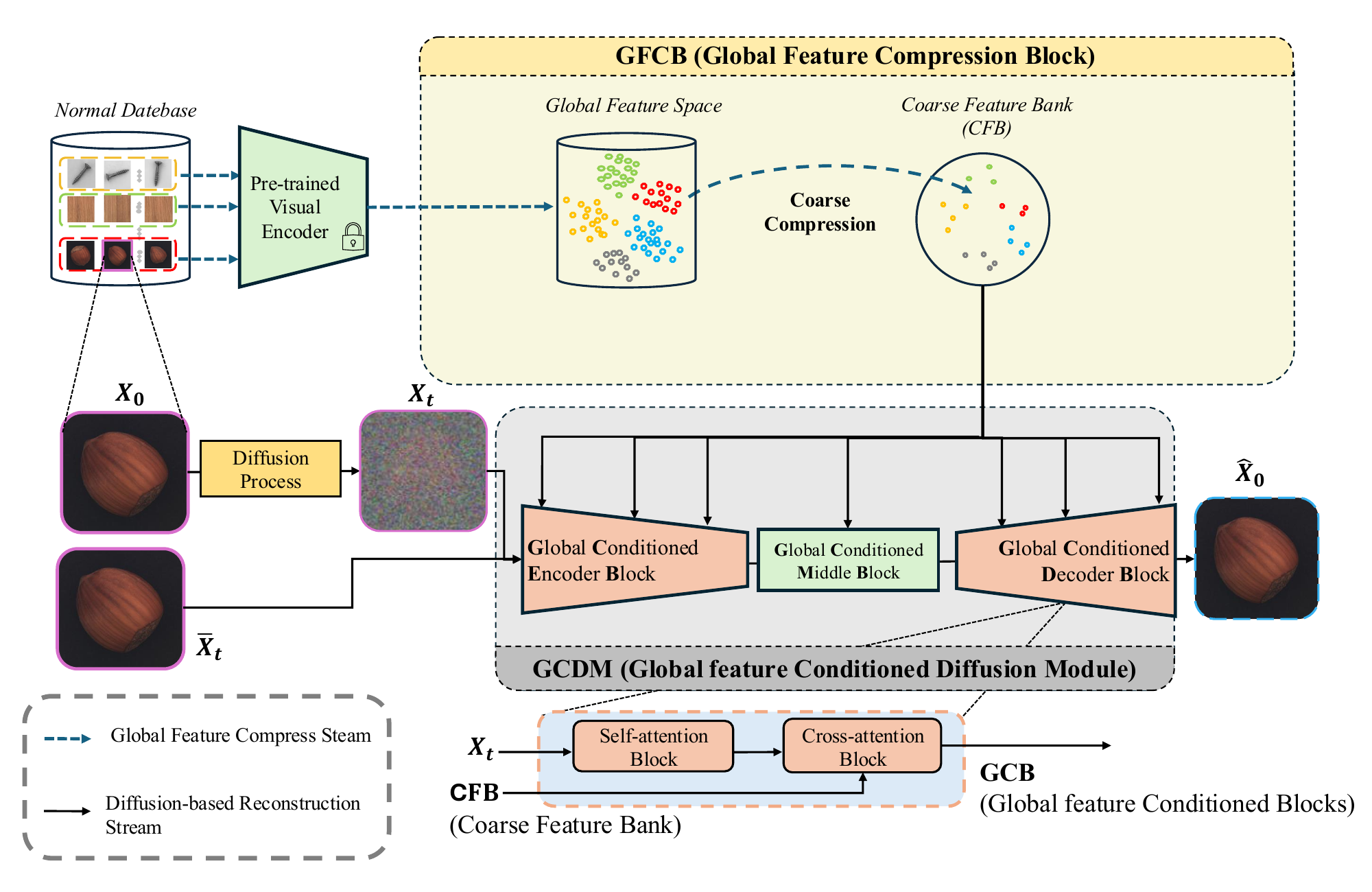}
    \subcaption{CCAD(V)}\label{CCAD_V}
    \end{subfigure}
    \captionsetup{justification=centering,margin=0.1cm}
    \vspace{-2mm}
     \caption{The frameworks of CCAD(C \& V)}
    \label{Overview of CCAD}
\end{minipage}
\end{figure}
The framework of CCAD(C) is shown in figure \ref{CCAD_C}. Unlike the framework in figure \ref{CCAD_F_pipline}, we replace the FFB with CFB. The framework of CCAD(V) is shown in figure \ref{CCAD_V}. The diffusion process is conducted on the pixel space and we only implement a backbone consisting of GCEB and GCDB to reconstruction images

\section{The complete algorithm of CCAD}
In this section, we list the pseudo-code of our CCAD algorithm
\begin{algorithm}[H]
\caption{CCAD(F) (\colorbox{rgb:red!2,65;green!30,60;blue!20,125}{Training} and \colorbox{rgb:red!30,155;green!20,20;blue!20,30}{Reconstruction})}
\label{CCAD(F)}
\textbf{Input}: $\mathcal{B}_c, \mathcal{X}$\\
\textbf{Pre-trained Autoencoder}: $\mathcal{E}, \mathscr{D}$\\
\textbf{Trainable Model}: $\boldsymbol{\tau}_{\boldsymbol{\theta}}$; $\boldsymbol{\epsilon}_{\boldsymbol{\Theta}}$
\begin{algorithmic}[1] 
\colorbox{rgb:red!2,65;green!30,60;blue!20,125}{
\parbox{0.3\textwidth}{\vbox{
\REPEAT
\STATE $\textbf{x} \sim \mathcal{X}$, $\textbf{z} = \mathcal{E}(\textbf{x})$, $t \sim U(1, T)$, $\boldsymbol{\epsilon} \sim \mathcal{N}(0, \boldsymbol{I})$
\STATE $\textbf{z}_t = \sqrt{\bar{\alpha}_{t}}~\mathbf{z}_0 + \sqrt{1 - \bar{\alpha}_t}\boldsymbol{\epsilon}$
\STATE Generate $\mathcal{D}_{bs}$
\STATE $\min\limits_{\boldsymbol{\Theta};\boldsymbol{\theta}}\|\boldsymbol{\epsilon} - \boldsymbol{\epsilon}^t_{\boldsymbol{\Theta}}(\mathbf{z}_{t}; \mathbf{x}; \boldsymbol{\tau}_{\boldsymbol{\theta}}(\mathcal{D}_{bs};\mathcal{B}_c)\|_2^2$
\UNTIL converged}}}
\colorbox{rgb:red!30,155;green!20,20;blue!20,30}{\parbox{0.305\textwidth}{\vbox{
\FOR{$\textbf{x}_i$ in $\mathcal{X}$}
\STATE $\textbf{z}_T \sim \mathcal{N}(0, \boldsymbol{I})$
\FOR{$t = T, \dots, 1$}
\STATE sample $\mathbf{z}_{t-1}$ through \eqref{CCAD_rec}
\ENDFOR
\STATE $\hat{\mathbf{x}}_i = \mathscr{D}(\mathbf{z_0})$
\ENDFOR
\STATE \textbf{return} $\hat{\mathcal{X}} = \{\hat{\mathbf{x}}_i\}$}}}
\end{algorithmic}
\end{algorithm}

\noindent Similar as equation \eqref{CCAD_rec}, we define the sampling equation in CCAD(C) as 
\begin{align}
\mathbf{z}_{t-1} &= \sqrt{\bar{\alpha}_{t-1}}(\frac{\mathbf{z}_{t} - \sqrt{1 - \bar{\alpha}_t}\boldsymbol{\epsilon}^t_{\boldsymbol{\Theta}}(\mathbf{z}_{t}; \mathbf{x}; \mathcal{B}_c)}{\sqrt{\bar{\alpha}_t}})\notag\\ &+\sqrt{1 - \bar{\alpha}_{t-1} - \sigma^2_t}\boldsymbol{\epsilon}^t_{\boldsymbol{\Theta}}(\mathbf{z}_{t}; \mathbf{x}; \mathcal{B}_c)\notag\\
    &+ \sigma_t\boldsymbol{\epsilon}_t.\label{CCAD_C_rec}
\end{align}

\noindent The pseudo-code of CCAD(C) can be denoted by
\begin{algorithm}[H]
\caption{CCAD(C) (\colorbox{rgb:red!2,65;green!30,60;blue!20,125}{Training} and \colorbox{rgb:red!30,155;green!20,20;blue!20,30}{Reconstruction})}
\label{CCAD(C)}
\textbf{Input}: $\mathcal{B}_c, \mathcal{X}$\\
\textbf{Pre-trained Autoencoder}: $\mathcal{E}, \mathscr{D}$\\
\textbf{Trainable Model}: $\boldsymbol{\epsilon}_{\boldsymbol{\Theta}}$
\begin{algorithmic}[1] 
\colorbox{rgb:red!2,65;green!30,60;blue!20,125}{\parbox{0.305\textwidth}{\vbox{
\REPEAT
\STATE $\textbf{x} \sim \mathcal{X}$, $\textbf{z} = \mathcal{E}(\textbf{x})$, $t \sim U(1, T)$, $\boldsymbol{\epsilon} \sim \mathcal{N}(0, \boldsymbol{I})$
\STATE $\textbf{z}_t = \sqrt{\bar{\alpha}_{t}}~\mathbf{z}_0 + \sqrt{1 - \bar{\alpha}_t}\boldsymbol{\epsilon}$
\STATE $\min\limits_{\boldsymbol{\Theta}}\|\boldsymbol{\epsilon} - \boldsymbol{\epsilon}^t_{\boldsymbol{\Theta}}(\mathbf{z}_{t}; \mathbf{x}; \mathcal{B}_c)\|_2^2$
\UNTIL converged}}}
\colorbox{rgb:red!30,155;green!20,20;blue!20,30}{\parbox{0.305\textwidth}{\vbox{
\FOR{$\textbf{x}_i$ in $\mathcal{X}$}
\STATE $\textbf{z}_T \sim \mathcal{N}(0, \boldsymbol{I})$
\FOR{$t = T, \dots, 1$}
\STATE sample $\mathbf{z}_{t-1}$ through \eqref{CCAD_C_rec}
\ENDFOR
\STATE $\hat{\mathbf{x}}_i = \mathscr{D}(\mathbf{z_0})$
\ENDFOR
\STATE \textbf{return} $\hat{\mathcal{X}} = \{\hat{\mathbf{x}}_i\}$}}}
\end{algorithmic}
\end{algorithm}

\noindent The pseudo-code of CCAD(V) can be denoted by
\begin{algorithm}[H]
\caption{CCAD(V) (\colorbox{rgb:red!2,65;green!30,60;blue!20,125}{Training} and \colorbox{rgb:red!30,155;green!20,20;blue!20,30}{Reconstruction})}
\label{CCAD(V)}
\textbf{Input}: $\mathcal{B}_c, \mathcal{X}$\\
\textbf{Trainable Model}: $\boldsymbol{\epsilon}_{\boldsymbol{\Theta}}$
\begin{algorithmic}[1] 
\colorbox{rgb:red!2,65;green!30,60;blue!20,125}{\parbox{0.305\textwidth}{\vbox{
\REPEAT
\STATE $\mathbf{x}_0 \sim \mathcal{X}$, $t \sim U(1, T)$, $\boldsymbol{\epsilon} \sim \mathcal{N}(0, \boldsymbol{I})$
\STATE $\mathbf{x}_t = \sqrt{\bar{\alpha}_{t}}~\mathbf{x}_0 + \sqrt{1 - \bar{\alpha}_t}\boldsymbol{\epsilon}$
\STATE $\min\limits_{\boldsymbol{\Theta}}\|\boldsymbol{\epsilon} - \boldsymbol{\epsilon}^t_{\boldsymbol{\Theta}}(\mathbf{x}_{t}; \mathcal{B}_c)\|_2^2$
\UNTIL converged}}}
\colorbox{rgb:red!30,155;green!20,20;blue!20,30}{\parbox{0.305\textwidth}{\vbox{
\FOR{$\bar{\mathbf{x}}_i$ in $\mathcal{X}$}
\STATE $\mathbf{x}_T \sim \mathcal{N}(0, \boldsymbol{I})$
\FOR{$t = T, \dots, 1$}
\STATE sample $\mathbf{x}_{t-1}$ by \eqref{CCAD_V_rec}
\ENDFOR
\STATE $\hat{\mathbf{x}}_i = \mathbf{x_0}$
\ENDFOR
\STATE \textbf{return} $\hat{\mathcal{X}} = \{\hat{\mathbf{x}}_i\}$}}}
\end{algorithmic}
\end{algorithm}
\vspace{-0.6cm}
\section{Additional Details on Experiment}
We list more experimental details in this section. The best result(s) are highlighted in bold.
\vspace{-0.2cm}
\subsection{MVTec Dataset}
\begin{table}[H]
\centering
\vspace{-0.3cm}
\begin{minipage}[t]{0.49\textwidth}
\captionsetup{justification=centering,margin=0.1cm}
\caption{F1 score (Class-level; Pixel-level) on MVTec-AD with DiAD, CCAD(C \& F).}\label{MVTEC_f1}
\vspace{-0.2cm}
\begin{adjustbox}{max width=1\textwidth}
\renewcommand{\arraystretch}{1.0}
\setlength{\tabcolsep}{5mm}
\begin{tabular}{c|ccc}
\hline
{Class}&DiAD&CCAD(C)&CCAD(F)\\
\hline  
\makecell{bottle}&\textbf{1.000}; 0.750&\textbf{1.000}; \textbf{0.760}&0.863; 0.119\\
\hline  
\makecell{cable}&0.848; 0.389&\textbf{0.854}; \textbf{0.398}&0.760; 0.105\\
\hline
\makecell{capsule}&\textbf{0.944}; 0.433&0.938; \textbf{0.437}&0.905; 0.124\\
\hline  
\makecell{carpet}&\textbf{0.983}; 0.566&0.967; \textbf{0.616}&0.943; 0.532\\
\hline 
\makecell{grid}&\textbf{0.982}; 0.147&\textbf{0.982}; \textbf{0.211}&0.874; 0.019\\
\hline
\makecell{hazelnut}&0.929; \textbf{0.498}&\textbf{0.958}; 0.484&0.904; 0.437\\
\hline
\makecell{leather}&\textbf{1.000}; 0.435&\textbf{1.000}; \textbf{0.465}&0.961; 0.377\\
\hline
\makecell{metal}&0.961; 0.834&\textbf{0.962}; \textbf{0.839}&0.894; 0.294\\
\hline
\makecell{pill}&0.953; 0.634&\textbf{0.954}; \textbf{0.646}&0.916; 0.097\\
\hline
\makecell{screw}&\textbf{0.907}; \textbf{0.280}&0.895; 0.216&0.869; 0.044\\
\hline
\makecell{tile}&\textbf{0.959}; 0.575&0.952; \textbf{0.578}&0.921; 0.500\\
\hline
\makecell{toothbrush}&0.938; 0.544&\textbf{0.952}; \textbf{0.644}&0.833; 0.092\\
\hline
\makecell{transistor}&\textbf{0.975}; 0.604&\textbf{0.975}; \textbf{0.614}&0.571; 0.096\\
\hline
\makecell{wood}&0.968; 0.466&0.967; \textbf{0.467}&\textbf{0.975}; 0.432\\
\hline
\makecell{zipper}&\textbf{0.954}; \textbf{0.516}&\textbf{0.954}; 0.480&0.881; 0.074\\
\hline
\makecell{mean}&0.953; 0.511&\textbf{0.954}; \textbf{0.524}&0.871; 0.223\\
\hline
\end{tabular}
\end{adjustbox}
\end{minipage}
\end{table}
\vspace{-0.6cm}
\begin{table}[H]
\centering
\begin{minipage}[t]{0.49\textwidth}
\captionsetup{justification=centering,margin=0.1cm}
\caption{Average precision (Class-level; Pixel-level) on MVTec-AD with DiAD, CCAD(C \& F).}\label{MVTEC_ap}
\vspace{-0.2cm}
\begin{adjustbox}{max width=1\textwidth}
\renewcommand{\arraystretch}{1.0}
\setlength{\tabcolsep}{5mm}
\begin{tabular}{c|ccc}
\hline  
Class&DiAD&CCAD(C)&CCAD(F)\\
\hline
\makecell{bottle} &\textbf{1.000}; 0.786&\textbf{1.000}; \textbf{0.790}&0.703; 0.048\\
\hline  
\makecell{cable} &\textbf{0.920}; 0.331&0.915; \textbf{0.347}&0.623; 0.057\\
\hline  
\makecell{capsule} &0.960; \textbf{0.405}&\textbf{0.981}; \textbf{0.405}&0.773; 0.062\\
\hline  
\makecell{carpet} &\textbf{0.997}; 0.612&0.994; \textbf{0.630}&0.987; 0.495\\
\hline 
\makecell{grid} &\textbf{0.965}; 0.051&0.933; \textbf{0.082}&0.914; 0.007\\
\hline
\makecell{hazelnut} &\textbf{0.980}; \textbf{0.492}&0.976; 0.473&0.968; 0.403\\
\hline
\makecell{leather} &\textbf{1.000}; 0.423&\textbf{1.000}; \textbf{0.479}&0.987; 0.341\\
\hline
\makecell{metal} &\textbf{0.994}; 0.848&0.993; \textbf{0.862}&0.744; 0.168\\
\hline
\makecell{pill} &\textbf{0.993}; 0.657&0.992; \textbf{0.661}&0.903; 0.056\\
\hline
\makecell{screw} &\textbf{0.936}; \textbf{0.171}&0.896; 0.107&0.802; 0.022\\
\hline
\makecell{tile} &\textbf{0.993}; \textbf{0.500}&0.991; 0.497&0.979; 0.435\\
\hline
\makecell{toothbrush} &0.941; 0.540&\textbf{0.967}; \textbf{0.640}&0.802; 0.048\\
\hline
\makecell{transistor} &\textbf{0.992}; \textbf{0.535}&0.990; 0.502&0.467; 0.051\\
\hline
\makecell{wood} &\textbf{0.997}; 0.431&0.995; \textbf{0.440}&0.995; 0.395\\
\hline
\makecell{zipper} &0.984; \textbf{0.439}&\textbf{0.988}; 0.394&0.760; 0.030\\
\hline
\makecell{mean} &\textbf{0.977}; 0.481&0.974; \textbf{0.487}&0.823; 0.174\\
\hline
\end{tabular}
\end{adjustbox}
\end{minipage}
\end{table}

\subsection{VisA Dataset}
\begin{table}[ht]
\centering
\vspace{-0.4cm}
\begin{minipage}[t]{0.98\textwidth}
\centering
\captionsetup{justification=centering,margin=0.1cm}
\caption[AUROC]{AUROC(Class-level; Pixel-level) comparison with SOTA methods on VisA \cite{VisA}}\label{SOTA_VisA_cls}
\begin{adjustbox}{max width=1\textwidth}
\setlength{\tabcolsep}{5mm}
\renewcommand{\arraystretch}{1.0}
\begin{tabular}{c|ccc|ccc}
\hline 
\multirow{2}*{\diagbox[height=2\line]{Class}{AUC}{Algorithm}}&\multicolumn{3}{c|}{Single class based}&\multicolumn{3}{c}{Multi-class based.}\\
\cline{2-7}
~ &PatchCore&DDAD&CCAD(V)&DiAD&CCAD(C)&CCAD(F)\\
\hline  
\makecell{candle}&0.717; 0.881&0.973; 0.982&\textbf{0.979}; \textbf{0.987}&0.826; 0.951&\textbf{0.846}; \textbf{0.957}&0.684; 0.938\\
\hline  
\makecell{cashew}&0.851; 0.777&\textbf{0.990}; \textbf{0.847}&0.988; 0.795&\textbf{0.842}; 0.806&0.812; 0.830&0.594; \textbf{0.947}\\
\hline
\makecell{capsules}&0.688; 0.974&0.981; \textbf{0.996}&\textbf{0.986}; \textbf{0.996}&\textbf{0.583}; \textbf{0.951}&0.527; 0.836&0.509; 0.793\\
\hline
\makecell{chewinggum}&0.925; \textbf{0.921}&0.975; 0.886&\textbf{0.999}; \textbf{0.921}&\textbf{0.909}; \textbf{0.920}&0.878; 0.916&0.787; 0.912\\
\hline 
\makecell{fryum}&0.769; 0.885&0.993; \textbf{0.947}&\textbf{0.999}; 0.930&0.743; 0.782&\textbf{0.773}; 0.852&0.675; \textbf{0.923}\\
\hline
\makecell{macaroni1}&0.785; 0.944&0.879; 0.969&\textbf{0.945}; \textbf{0.980}&\textbf{0.760}; \textbf{0.928}&0.733; 0.916&0.536; 0.884\\
\hline
\makecell{macaroni2}&0.472; 0.936&\textbf{0.930}; \textbf{0.994}&0. 916; \textbf{0.994}&\textbf{0.550}; \textbf{0.849}&0.524; 0.837&0.503; 0.841\\
\hline
\makecell{pipe fryum}&0.901; 0.920&0.903; \textbf{0.943}&\textbf{0.995}; 0.938&0.943; 0.984&\textbf{0.945}; \textbf{0.986}&0.636; 0.979\\
\hline
\makecell{pcb1}&0.730; 0.988&0.960; 0.992&\textbf{0.971}; \textbf{0.993}&0.686; \textbf{0.920}&\textbf{0.723}; 0.915&0.631; 0.889\\
\hline
\makecell{pcb2}&0.896; 0.971&\textbf{0.972}; 0.967&0.959; \textbf{0.973}&\textbf{0.740}; 0.915&0.735; \textbf{0.923}&0.670; 0.897\\
\hline
\makecell{pcb3}&0.701; 0.950&\textbf{1.000}; \textbf{0.982}&\textbf{1.000}; 0.980&\textbf{0.617}; 0.933&0.599; \textbf{0.950}&0.593; 0.948\\
\hline
\makecell{pcb4}&0.697; 0.934&\textbf{0.999}; 0.987&\textbf{0.999}; \textbf{0.991}&0.737; 0.856&\textbf{0.795}; 0.857&0.792; \textbf{0.865}\\
\hline
\makecell{mean}&0.761; 0.923&0.963; \textbf{0.958}&\textbf{0.978}; 0.957&\textbf{0.745}; 0.900&0.741; 0.898&0.634; \textbf{0.901}\\
\hline
\end{tabular}
\end{adjustbox}
\end{minipage}
\end{table}
\begin{minipage}{1\textwidth}
\begin{minipage}[t]{0.47\textwidth}
\makeatletter\def\@captype{table}
\captionsetup{justification=centering,margin=0.1cm, hypcap=false}
\caption{F1 score (Class-level; Pixel-level) on VisA with DiAD, CCAD(C \& F).}
\setlength{\tabcolsep}{2mm}
\begin{tabular}{c|ccc}
\hline 
F1&DiAD&CCAD(C)&CCAD(F)\\
\hline  
\makecell{candle}&\makecell{\small 0.741; \textbf{0.220} }&\makecell{\small \textbf{0.788}; \textbf{0.220} }&\makecell{\small 0.757; 0.180 }\\
\hline  
\makecell{cashew}&\makecell{\small \textbf{0.868}; 0.260}&\makecell{\small 0.855; \textbf{0.454}}&\makecell{\small 0.861; 0.282}\\
\hline  
\makecell{capsules}&\makecell{\small \textbf{0.769}; \textbf{0.291}}&\makecell{\small \textbf{0.769}; 0.049}&\makecell{\small \textbf{0.769}; 0.185}\\
\hline  
\makecell{chewinggum}&\makecell{\small 0.876; 0.404}&\makecell{\small 0.841; \textbf{0.481}}&\makecell{\small \textbf{0.884}; 0.380}\\
\hline 
\makecell{fryum}&\makecell{\small 0.800; 0.232}&\makecell{\small 0.803; \textbf{0.515}}&\makecell{\small \textbf{0.806}; 0.233}\\
\hline
\makecell{macaroni1}&\makecell{\small 0.688; 0.119}&\makecell{\small 0.708; \textbf{0.129}}&\makecell{\small \textbf{0.727}; 0.103}\\
\hline
\makecell{macaroni2}&\makecell{\small \textbf{0.669}; \textbf{0.021}}&\makecell{\small 0.667; 0.012}&\makecell{\small 0.667; \textbf{0.021}}\\
\hline
\makecell{pipe fryum}&\makecell{\small \textbf{0.942}; 0.478}&\makecell{\small 0.898; \textbf{0.633}}&\makecell{\small 0.933; 0.486}\\
\hline
\makecell{pcb1}&\makecell{\small \textbf{0.691}; 0.105}&\makecell{\small 0.677; \textbf{0.271}}&\makecell{\small 0.678; 0.127}\\
\hline
\makecell{pcb2}&\makecell{\small 0.703; 0.084}&\makecell{\small \textbf{0.747}; \textbf{0.105}}&\makecell{\small 0.708; 0.085}\\
\hline
\makecell{pcb3}&\makecell{\small \textbf{0.664}; 0.129}&\makecell{\small \textbf{0.664}; \textbf{0.210}}&\makecell{\small \textbf{0.664}; 0.146}\\
\hline
\makecell{pcb4}&\makecell{\small 0.664; 0.134}& \makecell{\small \textbf{0.746}; \textbf{0.162}}&\makecell{\small 0.693; 0.161}\\
\hline
\makecell{mean}&\makecell{\small 0.756; 0.206}&\makecell{\small \textbf{0.763}; \textbf{0.270}}& \makecell{\small 0.762; 0.199}\\
\hline
\end{tabular}
\label{VisA_f1}
\end{minipage}
\begin{minipage}[t]{0.5\textwidth}
\makeatletter\def\@captype{table}
\centering
\captionsetup{justification=centering,margin=0.1cm,hypcap=false}
\caption{Average precision (Class-level; Pixel-level) on VisA with DiAD, CCAD(C \& F).}
\setlength{\tabcolsep}{2mm}
\begin{tabular}{c|ccc}
\hline 
AP&DiAD&CCAD(C)&CCAD(F)\\
\hline  
\makecell{candle}&\makecell{\small 0.803; \textbf{0.116}}&\makecell{\small \textbf{0.846}; \textbf{0.116}}&\makecell{\small 0.798; 0.075}\\
\hline  
\makecell{cashew}&\makecell{\small 0.935; 0.170}&\makecell{\small 0.936; \textbf{0.361}}&\makecell{\small \textbf{0.948}; 0.176}\\
\hline  
\makecell{capsules}&\makecell{\small \textbf{0.673}; \textbf{0.157}}&\makecell{\small 0.593; 0.017}&\makecell{\small 0.625; 0.094}\\
\hline  
\makecell{chewinggum}&\makecell{\small \textbf{0.953}; 0.335}&\makecell{\small 0.933; \textbf{0.391}}&\makecell{\small 0.943; 0.293}\\
\hline 
\makecell{fryum}&\makecell{\small 0.866; 0.165}&\makecell{\small \textbf{0.902}; \textbf{0.410}}&\makecell{\small 0.857; 0.163}\\
\hline
\makecell{macaroni1}&\makecell{\small 0.719; 0.050}&\makecell{\small 0.741;\textbf{0.080}}&\makecell{\small \textbf{0.753}; 0.049}\\
\hline
\makecell{macaroni2}&\makecell{\small \textbf{0.483}; \textbf{0.007}}&\makecell{\small 0.463; 0.003}&\makecell{\small 0.434; 0.004}\\
\hline
\makecell{pipe fryum}&\makecell{\small \textbf{0.982}; 0.413}&\makecell{\small 0.961; \textbf{0.612}}&\makecell{\small 0.975; 0.425}\\
\hline
\makecell{pcb1}&\makecell{\small 0.731; 0.044}&\makecell{\small \textbf{0.769}; \textbf{0.122}}&\makecell{\small 0.717; 0.055}\\
\hline
\makecell{pcb2}&\makecell{\small 0.716; 0.027}&\makecell{\small \textbf{0.787}; \textbf{0.039}}&\makecell{\small 0.725; 0.028}\\
\hline
\makecell{pcb3}&\makecell{\small 0.636; 0.043}&\makecell{\small \textbf{0.692}; \textbf{0.073}}&\makecell{\small 0.652; 0.052}\\
\hline
\makecell{pcb4}&\makecell{\small 0.717; \textbf{0.068}}&\makecell{\small \textbf{0.832}; \textbf{0.068}}&\makecell{\small 0.777; 0.067}\\
\hline
\makecell{mean}&\makecell{\small 0.768; 0.133}&\makecell{\small \textbf{0.788}; \textbf{0.191}}&\makecell{\small 0.767; 0.123}\\
\hline
\end{tabular}
\label{VisA_ap}
\end{minipage}
\end{minipage}

\subsection{MVTec-loco Dataset}
\begin{table}[ht]
\centering
\vspace{-0.3cm}
\begin{minipage}[t]{1\textwidth}
\centering
\captionsetup{justification=centering,margin=0.1cm}
\caption[AUROC]{AUROC (Class-level; Pixel-level) comparison with SOTA methods on MVTec-loco \cite{MVTec-loco}.}\label{SOTA_loco_cls}
\begin{adjustbox}{max width=1\textwidth}
\renewcommand{\arraystretch}{1.4}
\setlength{\tabcolsep}{5mm}
\begin{tabular}{c|ccc|ccc}
\hline 
\multirow{2}*{\diagbox[height=2.8\line]{Class}{AUC}{Algorithm}}&\multicolumn{3}{c|}{Single class based}&\multicolumn{3}{c}{Multi-class based}\\
\cline{2-7}
~ &PatchCore&DDAD&CCAD(V)&DiAD&CCAD(C)&CCAD(F)\\
\hline  
\makecell{breakfast box}&0.609; 0.743&0.914; 0.870&\textbf{0.966}; \textbf{0.872}&0.533; 0.772&0.560; \textbf{0.794}&\textbf{0.561}; 0.751\\
\hline
\makecell{juice bottle}&0.783; 0.583&0.989; 0.833&\textbf{0.997}; \textbf{0.849}&0.883; 0.893&\textbf{0.919}; \textbf{0.898}&0.882; 0.893\\
\hline
\makecell{pushpins}&0.648; \textbf{0.653}&0.819; 0.612&\textbf{0.854}; 0.635&\textbf{0.703}; 0.576&0.662; 0.593&0.655; \textbf{0.597}\\
\hline
\makecell{screw bag}&0.596; \textbf{0.658}&0.741; 0.479&\textbf{0.749}; 0.503&0.559; \textbf{0.723}&0.542; 0.720&\textbf{0.561}; 0.717\\
\hline
\makecell{splicing connectors}&0.633; 0.578&\textbf{0.969}; \textbf{0.598}&0.920; 0.591&0.650; 0.604&0.669; 0.588&\textbf{0.692}; \textbf{0.617}\\
\hline
\makecell{mean}&0.654; 0.643&0.886; 0.678&\textbf{0.897}; \textbf{0.690}&0.665; 0.714&\textbf{0.671}; \textbf{0.718}&0.670; 0.715\\
\hline
\end{tabular}
\end{adjustbox}
\end{minipage}
\end{table}

\begin{minipage}{1\textwidth}
\begin{minipage}[t]{0.49\textwidth}
\makeatletter\def\@captype{table}
\captionsetup{justification=centering,margin=0.1cm,hypcap=false}
\caption{F1 score (Class-level; Pixel-level) on MVTec-loco with DiAD, CCAD(C \& F).}
\setlength{\tabcolsep}{1.5mm}
\centering
\begin{tabular}{c|ccc}
\hline 
{F1}&DiAD&CCAD(C)&CCAD(F)\\
\hline  
\makecell{breakfast box}&\makecell{\small \textbf{0.772}; \textbf{0.331}}&\makecell{\small \textbf{0.772}; 0.131}&\makecell{\small \textbf{0.772}; 0.314}\\
\hline
\makecell{juice bottle}&\makecell{\small \textbf{0.888}; 0.471}&\makecell{\small 0.834; 0.222}&\makecell{\small 0.881; \textbf{0.483}}\\
\hline
\makecell{pushpins}&\makecell{\small \textbf{0.720}; 0.068}&\makecell{\small 0.714; 0.019}&\makecell{\small 0.714; \textbf{0.073}}\\
\hline
\makecell{screw bag}&\makecell{\small 0.783; 0.140}&\makecell{\small 0.782; 0.102}&\makecell{\small \textbf{0.785}; \textbf{0.141}}\\
\hline
\makecell{splicing connectors}&\makecell{\small 0.764; \textbf{0.186}}&\makecell{\small 0.764; 0.145}&\makecell{\small \textbf{0.768}; 0.168}\\
\hline
\makecell{mean}&\makecell{\small \textbf{0.785}; \textbf{0.239}}&\makecell{\small 0.773; 0.124}&\makecell{\small 0.784; 0.236}\\
\hline
\end{tabular}
\label{loco_f1}
\end{minipage}
\begin{minipage}[t]{0.49\textwidth}
\makeatletter\def\@captype{table}
\centering
\captionsetup{justification=centering,margin=0.1cm,hypcap=false}
\caption{Average precision (Class-level; Pixel-level) on MVTec-loco with DiAD, CCAD(C \& F).}
\setlength{\tabcolsep}{1.5mm}
\begin{tabular}{c|ccc}
\hline 
AP&DiAD&CCAD(C)&CCAD(F)\\
\hline  
\makecell{breakfast box}&\makecell{\small 0.728; \textbf{0.237}}&\makecell{\small 0.622; 0.051}&\makecell{\small \textbf{0.732}; 0.230}\\
\hline
\makecell{juice bottle}&\makecell{\small \textbf{0.963}; 0.451}&\makecell{\small 0.705; 0.129}&\makecell{\small \textbf{0.963}; \textbf{0.476}}\\
\hline
\makecell{pushpins}&\makecell{\small \textbf{0.707}; 0.018}&\makecell{\small 0.583; 0.008}&\makecell{\small 0.689; \textbf{0.019}}\\
\hline
\makecell{screw bag}&\makecell{\small 0.650; 0.081}&\makecell{\small 0.660; 0.054}&\makecell{\small \textbf{0.690}; \textbf{0.083}}\\
\hline
\makecell{splicing connectors}&\makecell{\small \textbf{0.786}; \textbf{0.137}}&\makecell{\small 0.629; 0.092}&\makecell{\small 0.753; 0.105}\\
\hline
\makecell{mean}&\makecell{\small \textbf{0.767}; \textbf{0.185}}&\makecell{\small 0.640; 0.067}&\makecell{\small 0.765; 0.183}\\
\hline
\end{tabular}
\label{loco_ap}
\end{minipage}
\end{minipage}

\clearpage
\subsection{MVTec-3d Dataset}
\begin{table}[ht]
\centering
\vspace{-0.4cm}
\begin{minipage}[t]{1\textwidth}
\centering
\captionsetup{justification=centering,margin=0.1cm}
\caption[AUROC]{AUROC(Class-level; Pixel-level) comparison with SOTA methods on MVTec-3d \cite{MVTec-3d}.}\label{SOTA_3d_cls}
\begin{adjustbox}{max width=1\textwidth}
\setlength{\tabcolsep}{5mm}
\vspace{-1.0cm}
\renewcommand{\arraystretch}{1.2}
\begin{tabular}{c|ccc|ccc}
\hline 
\multirow{2}*{\diagbox[height=2.4\line]{Class}{AUC}{Algorithm}}&\multicolumn{3}{c|}{Single class based}&\multicolumn{3}{c}{Multi-class based}\\
\cline{2-7}
~ &PatchCore&DDAD&CCAD(V)&DiAD&CCAD(C)&CCAD(F)\\
\hline  
\makecell{bagel}&0.671; 0.439&0.878; 0.974&\textbf{0.892}; \textbf{0.977}&0.756; \textbf{0.990}&\textbf{0.955}; 0.987&0.920; 0.983\\
\hline  
\makecell{cable gland}&0.682; 0.981&0.957; 0.991&\textbf{0.959}; \textbf{0.994}&\textbf{0.842}; 0.969&0.607; \textbf{0.977}&0.553; 0.944\\
\hline  
\makecell{carrot}&0.669; 0.621&0.669; 0.649&\textbf{0.722}; \textbf{0.824}&0.804; \textbf{0.982}&0.674; 0.980&\textbf{0.891}; 0.936\\
\hline  
\makecell{cookie}&0.662; 0.675&\textbf{0.677}; 0.955&0.671; \textbf{0.960}&\textbf{0.714}; 0.937&0.631; \textbf{0.946}&0.585; 0.932\\
\hline  
\makecell{dowel}&0.711; \textbf{0.995}&0.996; 0.994&\textbf{0.999}; 0.994&0.837;\textbf{0.987}&\textbf{0.891}; 0.985&0.769; 0.939\\
\hline 
\makecell{foam}&0.671; 0.719&\textbf{0.918}; 0.867&0.914; \textbf{0.891}&0.533; 0.833&\textbf{0.774}; \textbf{0.944}&0.563; 0.811\\
\hline 
\makecell{peach}&\textbf{0.637}; 0.498&0.559; 0.953&0.563; \textbf{0.956}&0.564; 0.982&0.543; \textbf{0.987}&\textbf{0.594}; 0.969\\
\hline 
\makecell{potato}&\textbf{0.645}; 0.535&0.458; \textbf{0.817}&0.456; 0.793&\textbf{0.777}; \textbf{0.985}&0.647; \textbf{0.985}&0.689;0.936\\
\hline 
\makecell{rope}&0.683; 0.591&0.875; 0.986&\textbf{0.887}; \textbf{0.987}&0.741; 0.982&\textbf{0.750}; 0.974&0.697; \textbf{0.983}\\
\hline 
\makecell{tire}&0.666; 0.918&0.664; 0.867&\textbf{0.803}; \textbf{0.980}&0.524; 0.921&\textbf{0.619}; 0.942&0.569; \textbf{0.944}\\
\hline 
\makecell{mean}&0.67; 0.697&0.765; 0.905&\textbf{0.787}; \textbf{0.936}&\textbf{0.709}; 0.957&\textbf{0.709}; \textbf{0.971}&0.683; 0.938\\
\hline 
\end{tabular}
\end{adjustbox}
\end{minipage}
\end{table}

\begin{minipage}{1\textwidth}
\begin{minipage}[t]{0.5\textwidth}
\makeatletter\def\@captype{table}
\captionsetup{justification=centering,margin=0.1cm,hypcap=false}
\caption{F1 score (Class-level; Pixel-level) on MVTec-3d with DiAD, CCAD(C \& F)}
\setlength{\tabcolsep}{2mm}
\centering
\renewcommand{\arraystretch}{1.5}
\begin{tabular}{c|ccc}
\hline 
F1&DiAD&CCAD(C)&CCAD(F)\\
\hline  
\makecell{bagel}&\makecell{\small \textbf{0.960}; 0.537}&\makecell{\small 0.947; \textbf{0.545}}&\makecell{\small 0.927; 0.422}\\
\hline  
\makecell{cable gland}&\makecell{\small \textbf{0.952}; 0.149}&\makecell{\small 0.900; \textbf{0.313}}&\makecell{\small \textbf{0.952}; 0.042}\\
\hline  
\makecell{carrot}&\makecell{\small 0.927; \textbf{0.308}}&\makecell{\small 0.894; 0.286}&\makecell{\small \textbf{0.941}; 0.079}\\
\hline  
\makecell{cookie}&\makecell{\small \textbf{0.902}; \textbf{0.262}}&\makecell{\small 0.875; 0.165}&\makecell{\small \textbf{0.902}; 0.242}\\
\hline  
\makecell{dowel}&\makecell{\small 0.898; 0.324}&\makecell{\small 0.957; \textbf{0.396}}&\makecell{\small \textbf{0.958}; 0.065}\\
\hline 
\makecell{foam}&\makecell{\small 0.857; \textbf{0.349}}&\makecell{\small 0.919; 0.231}&\makecell{\small \textbf{0.947}; 0.114}\\
\hline 
\makecell{peach}&\makecell{\small \textbf{0.926}; \textbf{0.380}}&\makecell{\small 0.894; 0.265}&\makecell{\small \textbf{0.926}; 0.131}\\
\hline 
\makecell{potato}&\makecell{\small 0.955; \textbf{0.296}}&\makecell{\small 0.926; 0.232}&\makecell{\small \textbf{0.978}; 0.051}\\
\hline 
\makecell{rope}&\makecell{\small 0.762; 0.451}&\makecell{\small 0.769; \textbf{0.476}}&\makecell{\small \textbf{0.833}; 0.369}\\
\hline 
\makecell{tire}&\makecell{\small \textbf{0.963}; 0.104}&\makecell{\small 0.818; \textbf{0.117}}&\makecell{\small 0.902; 0.076}\\
\hline 
\makecell{mean}&\makecell{\small 0.910; \textbf{0.316}}&\makecell{\small 0.890; 0.303}&\makecell{\small \textbf{0.927}; 0.159}\\
\hline 
\end{tabular}
\label{3d_f1}
\end{minipage}
\begin{minipage}[t]{0.49\textwidth}
\makeatletter\def\@captype{table}
\centering
\captionsetup{justification=centering,margin=0.1cm,hypcap=false}
\caption{Average precision (Class-level; Pixel-level) on MVTec-3d with DiAD, CCAD(C \& F)}
\setlength{\tabcolsep}{2mm}
\renewcommand{\arraystretch}{1.5}
\begin{tabular}{c|ccc}
\hline  
AP&DiAD&CCAD(C)&CCAD(F)\\
\hline  
\makecell{bagel}&\makecell{\small 0.976; 0.500}&\makecell{\small \textbf{0.987}; \textbf{0.517}}&\makecell{\small 0.964; 0.342}\\
\hline  
\makecell{cable gland}&\makecell{\small 0.883; 0.077}&\makecell{\small 0.905; \textbf{0.196}}&\makecell{\small \textbf{0.941}; 0.018}\\
\hline  
\makecell{carrot}&\makecell{\small \textbf{0.980}; 0.238}&\makecell{\small 0.862; \textbf{0.240}}&\makecell{\small 0.957; 0.037}\\
\hline  
\makecell{cookie}&\makecell{\small 0.843; \textbf{0.240}}&\makecell{\small 0.788; 0.098}&\makecell{\small \textbf{0.904}; 0.218}\\
\hline  
\makecell{dowel}&\makecell{\small 0.903; 0.216}&\makecell{\small \textbf{0.985}; \textbf{0.331}}&\makecell{\small 0.903; 0.032}\\
\hline 
\makecell{foam}&\makecell{\small 0.943; \textbf{0.240}}&\makecell{\small 0.963; 0.122}&\makecell{\small \textbf{0.971}; 0.045}\\
\hline 
\makecell{peach}&\makecell{\small 0.918; \textbf{0.308}}&\makecell{\small 0.840; 0.224}&\makecell{\small \textbf{0.920}; 0.060}\\
\hline 
\makecell{potato}&\makecell{\small 0.892; \textbf{0.232}}&\makecell{\small \textbf{0.939}; 0.194}&\makecell{\small 0.914; 0.023}\\
\hline 
\makecell{rope}&\makecell{\small 0.868; \textbf{0.455}}&\makecell{\small 0.820; 0.449}&\makecell{\small \textbf{0.903}; 0.346}\\
\hline 
\makecell{tire}&\makecell{\small \textbf{0.923}; \textbf{0.046}}&\makecell{\small 0.733; \textbf{0.046}}&\makecell{\small 0.840; 0.027}\\
\hline 
\makecell{mean}&\makecell{\small 0.913; \textbf{0.255}}&\makecell{\small 0.882; 0.242}&\makecell{\small \textbf{0.922}; 0.115}\\
\hline 
\end{tabular}
\label{3d_ap}
\end{minipage}
\end{minipage}

\begin{minipage}{1\textwidth}
\begin{minipage}[t]{0.45\textwidth}
\makeatletter\def\@captype{table}
\vspace{8mm}
\small
\captionsetup{justification=centering,margin=0.1cm,hypcap=false}
\caption{Minimal epochs required to reach the same AUC level on each dataset.}
\setlength{\tabcolsep}{2mm}
\centering
\renewcommand{\arraystretch}{1.8}
\begin{tabular}{c|ccc|c}
\hline 
$eph$&DiAD&\makecell{CCAD\\(C)}&\makecell{CCAD\\(F)}&\makecell{AUC\\Lv}\\
\hline  
\makecell{MVTec-AD}&200&\textbf{100}&110&0.958; 0.958\\
\hline  
\makecell{VisA}&200&\textbf{60}&180&0.749; 0.901\\
\hline  
\makecell{MVTec-3d}&200&75&\textbf{10}&0.784; 0.969\\
\hline
\makecell{MVTec-loco}&200&\textbf{120}&\textbf{120}&0.668; 0.719\\
\hline
\makecell{MTD}&190&\textbf{110}&180&0.956; 0.821\\
\hline
\end{tabular}
\label{faster_convergence_speed}
\end{minipage}
\begin{minipage}[t]{0.55\textwidth}
\makeatletter\def\@captype{table}
\vspace{8mm}
\centering
\captionsetup{justification=centering,margin=0.1cm,hypcap=false}
\caption{AUROC(Class-level; Pixel-level) on datasets with different $\xi$ in CCAD(C \& F).}
\renewcommand{\arraystretch}{1.8}
\setlength{\tabcolsep}{0.8mm}
\centering
\begin{tabular}{c|cc|cc}
\hline 
{AUROC}&\multicolumn{2}{c}{CCAD(C)}&\multicolumn{2}{c}{CCAD(F)}\\
\hline
$\xi$&$10$&$100$&$10$&$100$\\
\hline  
\makecell{MVTec-AD}&\makecell{\small \textbf{0.960}; \textbf{0.963}}&\makecell{\small0.957; 0.961}&\makecell{\small \textbf{0.955}; \textbf{0.964}}&\makecell{\small0.953; 0.961}\\
\hline  
\makecell{DAGM}&\makecell{\small \textbf{0.825}; \textbf{0.932}}&\makecell{\small0.823; \textbf{0.932}}&\makecell{\small0.812; 0.924}&\makecell{\small \textbf{0.830}; \textbf{0.933}}\\
\hline  
\makecell{MVTec-3d}&\makecell{\small0.780; 0.970}&\makecell{\small \textbf{0.781}; \textbf{0.972}}&\makecell{\small \textbf{0.649}; \textbf{0.942}}& \makecell{\small0.636; 0.939}\\
\hline
\end{tabular}
\label{diff_dim2}
\end{minipage}
\end{minipage}
\clearpage

\noindent Shown in table \ref{MVTEC_f1} - \ref{3d_ap}, we list AUROC, F1 score and AP on MVTec-AD \cite{mvtec}, VisA \cite{VisA}, MVTec-loco \cite{MVTec-loco} and MVTec-3d \cite{mvtec}. Extensive experimental data has consistently demonstrated that both CCAD(C) and CCAD(F) exhibit a performance advantage over DiAD across various metrics. Furthermore, CCAD(V) demonstrates a similar performance advantage when compared to both DDAD and Patchcore.
This superiority is observed in multiple aspects of the evaluation, indicating that CCAD generally achieves more robust and reliable results in comparison to DiAD, DDAD, and Patchcore.

\subsection{Faster Convergence in CCAD}
In table \ref{faster_convergence_speed}, CCAD(C) and CCAD(F) achieve the same AUC level faster than DiAD, which we attribute to the introduction of feature banks. These conditions contribute significantly by letting the model learn relevant and representative features of the overall dataset more efficiently, thereby accelerating the convergence process.

\subsection[Ablation Studies on different]{Ablation Studies on different $\xi$}
We conducted a comparative analysis of the AUC performance of DiAD, CCAD(C), and CCAD(F) across different $\xi$ settings on multiple datasets. Notably, even when the $\xi$ value is as low as 10, the AUC remains significantly high. This indicates that only a few samples are sufficient to serve as conditions, enabling the model to efficiently reconstruct normal images. This efficiency highlights the strength of the CCAD models in leveraging minimal conditions to achieve robust image reconstruction, which is crucial for effective anomaly detection.

\subsection{DAGM re-annotation comparison}
Shown in figure \ref{DAGM_hist}, we compare the AUC performance of the original DAGM2007 dataset \cite{DAGM} with our re-annotated dataset on SOTAs. Under the same methods, we observed that the class-wise AUC remains similar and the pixel-wise AUC significantly improved. This indicates that our labeled data is better aligned with the anomaly detection task compared to the original data.

\begin{figure*}[t]
\begin{minipage}[c]{1\textwidth}
    \centering
    \begin{subfigure}[t]{0.49\textwidth}
    \centering
    \includegraphics[width=8.0cm]{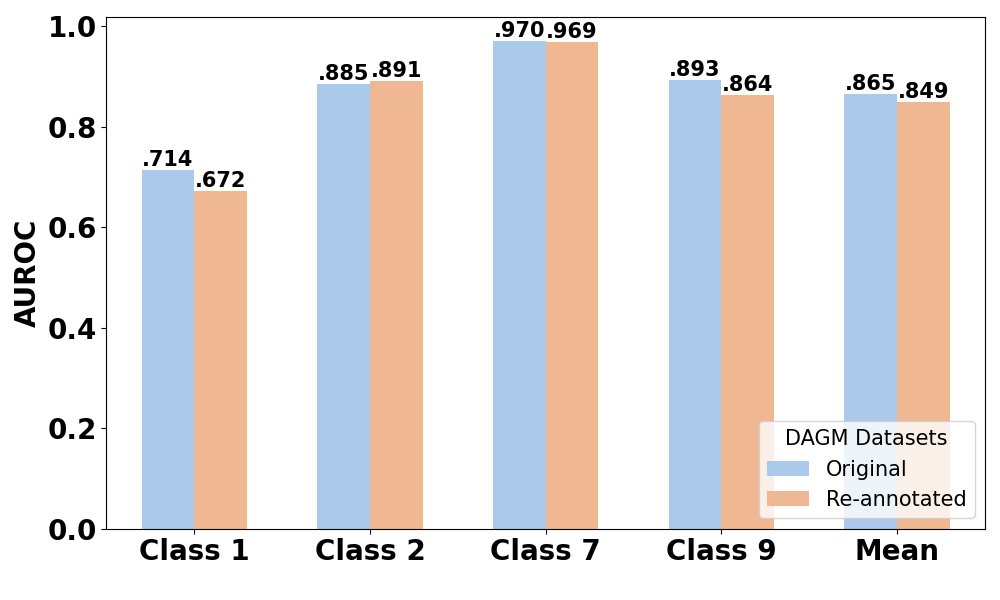}
    \centering
    \captionsetup{justification=centering}
    \subcaption{Class-wise AUC on PatchCore}
    \end{subfigure}
    \begin{subfigure}[t]{0.49\textwidth}
    \centering
    \includegraphics[width=8.0cm]{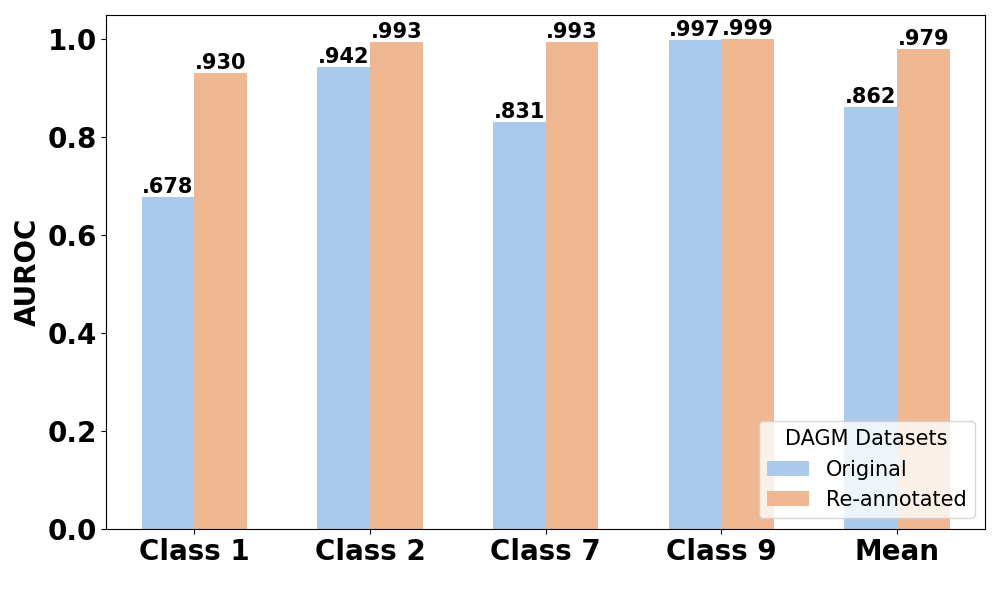}
    \captionsetup{justification=centering}
    \subcaption{Pixel-wise AUC on PatchCore}
    \end{subfigure}
    \begin{subfigure}[t]{0.49\textwidth}
    \centering
    \includegraphics[width=8.0cm]{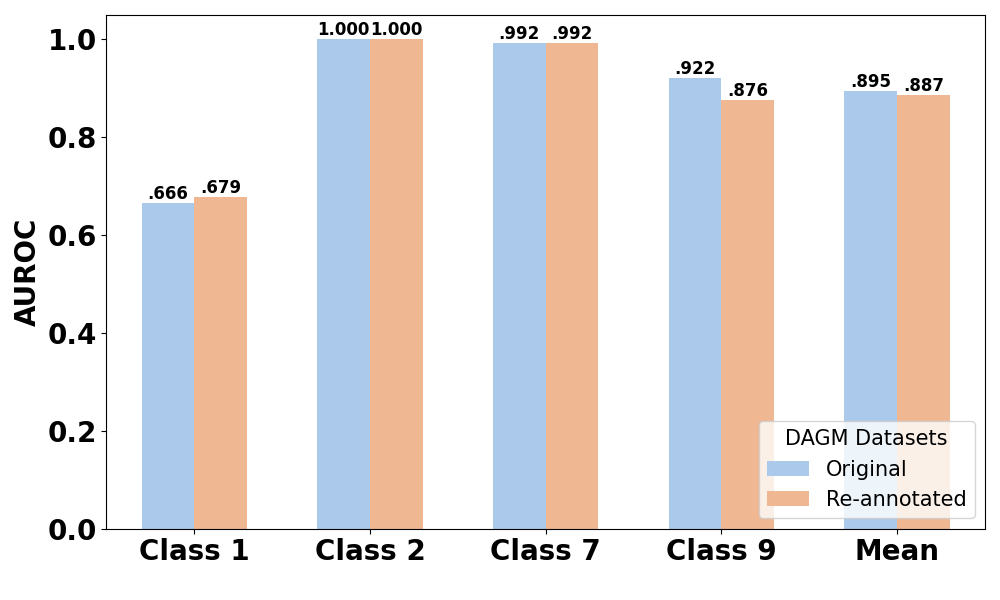}
    \centering
    \captionsetup{justification=centering}
    \subcaption{Class-wise AUC on DDAD}
    \end{subfigure}
    \begin{subfigure}[t]{0.49\textwidth}
    \centering
    \includegraphics[width=8.0cm]{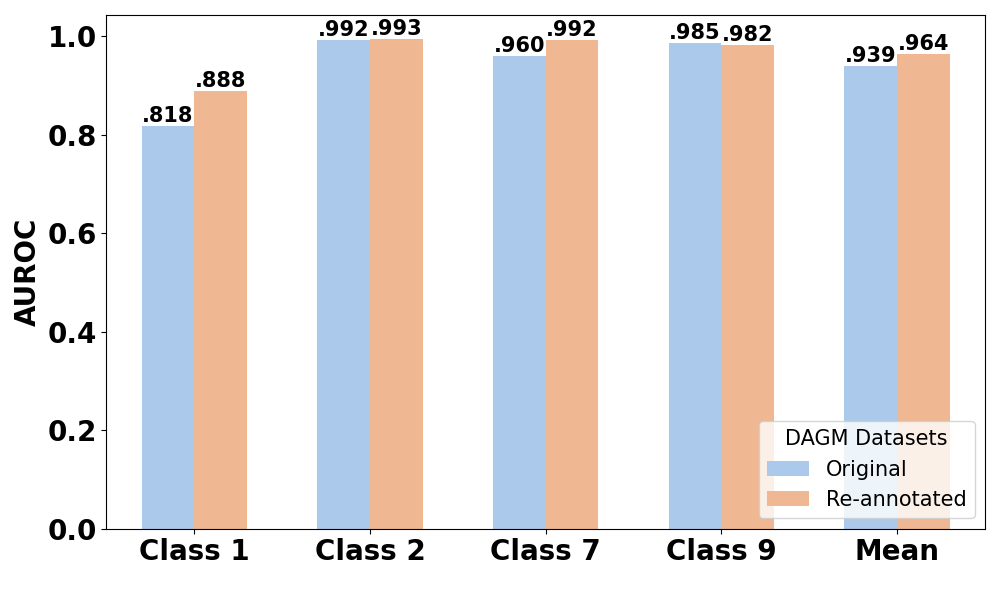}
    \captionsetup{justification=centering}
    \subcaption{Pixel-wise AUC on DDAD}
    \end{subfigure}
\end{minipage}
\end{figure*}
\begin{figure*}[t]\ContinuedFloat
\begin{minipage}[c]{1\textwidth}
    \begin{subfigure}[t]{0.49\textwidth}
    \centering
    \includegraphics[width=8.0cm]{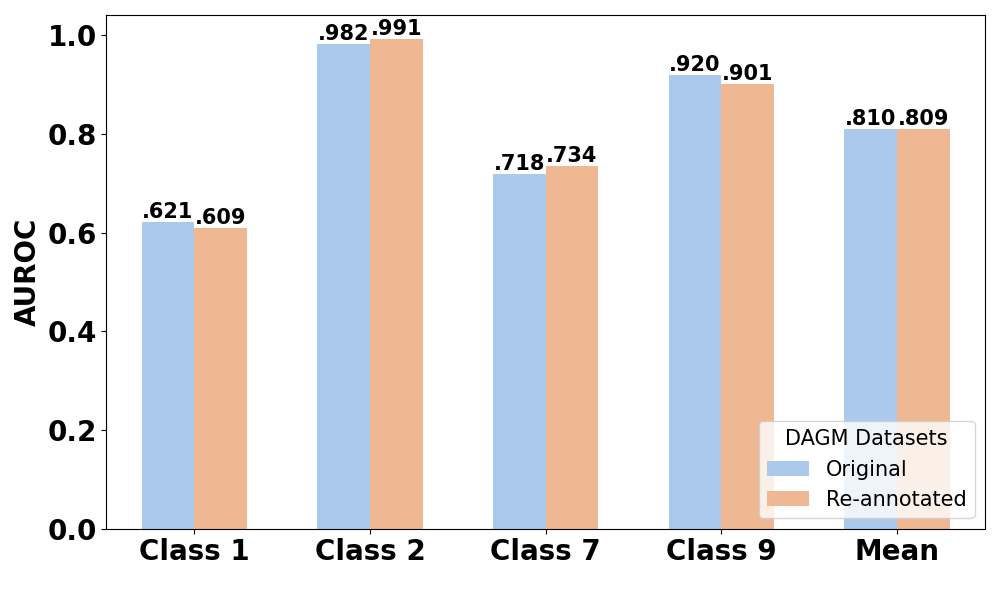}
    \centering
    \captionsetup{justification=centering}
    \subcaption{Class-wise AUC on DiAD}
    \end{subfigure}
    \begin{subfigure}[t]{0.49\textwidth}
    \centering
    \includegraphics[width=8.0cm]{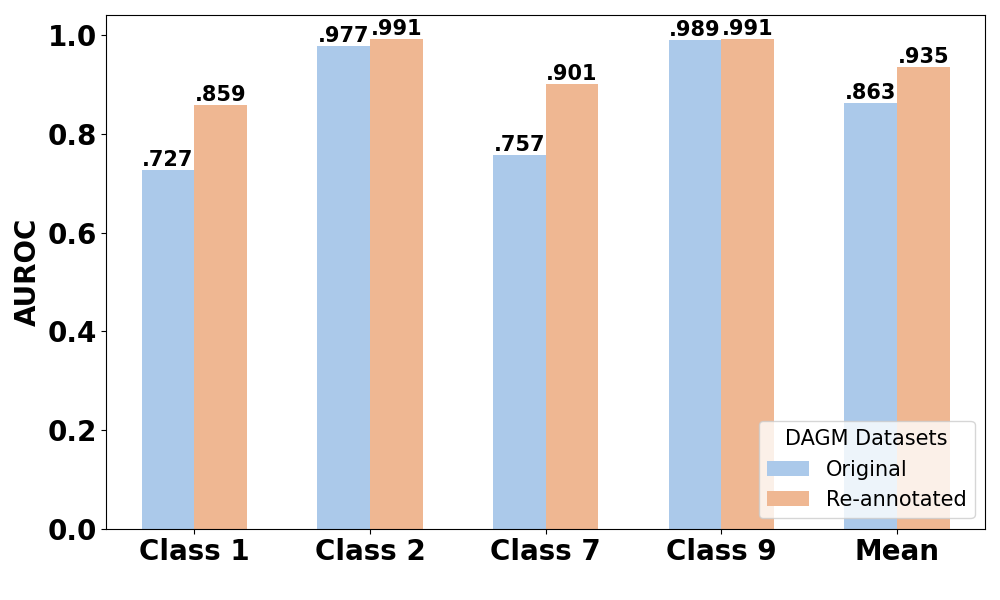}
    \captionsetup{justification=centering}
    \subcaption{Pixel-wise AUC on DiAD}
    \end{subfigure}
    \begin{subfigure}[t]{0.49\textwidth}
    \centering
    \includegraphics[width=8.0cm]{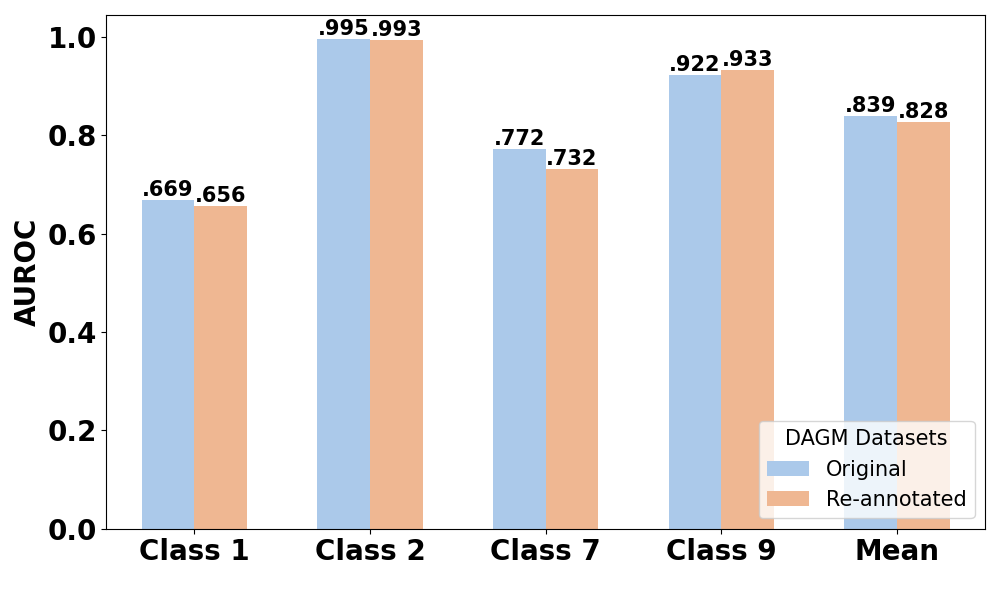}
    \centering
    \captionsetup{justification=centering}
    \subcaption{Class-wise AUC on CCAD (C)}
    \end{subfigure}
    \begin{subfigure}[t]{0.49\textwidth}
    \centering
    \includegraphics[width=8.0cm]{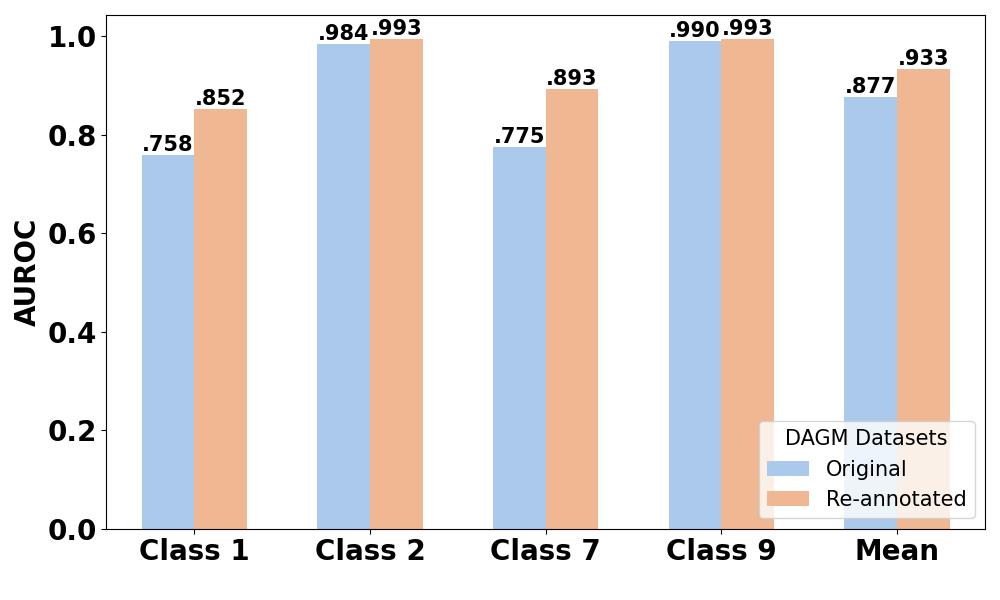}
    \captionsetup{justification=centering}
    \subcaption{Pixel-wise AUC on CCAD(C)}
    \end{subfigure}
    \centering
    \captionsetup{justification=centering,margin=0.1cm}
    \vspace{-2mm}
    \caption{Class-wise AUC of DAGM on STOAs}
    \label{DAGM_hist}
\end{minipage}
\end{figure*}

\section{Hyperparameters Setting}\label{hypterparameter}
As shown in table \ref{Hyper-parameter}, we provide a summary of hyperparameter for our proposed algorithms. To further validate our approach, we have included additional detailed hyperparameters below to address the results. We also listed the source code of each SOTA in table \ref{source_code} and hyperparameters of each experiment in table \ref{abbrev} and \ref{hyper} to demonstrate the validity.

\begin{table}[H]
\centering
\begin{minipage}[t]{0.49\textwidth}
\captionsetup{justification=centering,margin=0.1cm}
\caption{Hyper-parameter Setting.}\label{Hyper-parameter}
\vspace{-0.2cm}
\begin{adjustbox}{max width=1\textwidth}
\renewcommand{\arraystretch}{1.8}
\begin{tabular}{c|ccc}
\hline
        Hyper-parameter & CCAD(V) & CCAD(C) & CCAD(F) \\
        \hline
        \makecell{\# of Epoch} & 500-3000 & 50-200 & 50-200 \\
        \hline
        \makecell{\# of Trainable \\ Parameters} & 45.3M & 1.4B & 1.5B \\
        \hline
        Batchsize & 32 & 12 & 12 \\
        \hline
        \makecell{Pretrain \\ Feature Extractor} & WideResNet101 & ResNet50 & ResNet50 \\
        \hline
        Learning Rate& $3\times 10^{-4}$ & $1\times10^{-6}$ - $1\times10^{-4}$ & $1\times10^{-6}$ - $1\times10^{-4}$ \\
        \hline
        Model Input & $256\times256\times3$ & $256\times256\times3$ & $256\times256\times3$ \\
        \hline
        Optimizer & Adam & AdamW & AdamW \\
        \hline
        Weight decay& 0.05 & 0.05 & 0.05 \\
        \hline
\end{tabular}
\end{adjustbox}
\end{minipage}
\begin{minipage}[t]{0.49\textwidth}
\captionsetup{justification=centering,margin=0.1cm}
\vspace{0.2cm}
\caption{Source code link.}\label{source_code}
\vspace{-0.2cm}
\begin{adjustbox}{max width=1\textwidth}
\renewcommand{\arraystretch}{1.8}
\begin{tabular}{c|c}
\hline
        SOTA & link\\
        \hline
        PatchCore &\makecell{ https://github.com/LuigiFederico/PatchCore-for-\\Industrial-Anomaly-Detection/tree/main}\\
        \hline
        SPADE & https://github.com/byungjae89/SPADE-pytorch\\
        \hline
        DDAD & https://github.com/arimousa/DDAD/tree/main\\
        \hline
        DiAD & https://github.com/lewandofskee/DiAD\\
        \hline
\end{tabular}
\end{adjustbox}
\end{minipage}
\begin{minipage}[t]{0.49\textwidth}
\centering
\captionsetup{justification=centering,margin=0.1cm}
\vspace{0.2cm}
\caption{Abbreviations and Their Corresponding Definitions in the Experimental Setup.}\label{abbrev}
\vspace{-0.2cm}
\begin{adjustbox}{max width=1\textwidth}
\renewcommand{\arraystretch}{1.8}
\begin{tabular}{c|c}
\hline
        abbreviation & meaning\\
        \hline
        PFE & Pretrained feature extractor\\
        \hline
        $\xi$ & $\#$ of samples\\
        \hline
        $\ell$ & learning rate\\
        \hline
        $eph$ & $\#$ of epochs\\
        \hline
        $Bs$ & Batch size\\
        \hline
\end{tabular}
\end{adjustbox}
\end{minipage}
\end{table}

\begin{table*}[ht]
\centering
\vspace{-0.4cm}
\begin{minipage}[t]{1\textwidth}
\centering
\captionsetup{justification=centering,margin=0.5cm}
\caption[AUROC]{Hyper-parameter setting of SOTA on different datasets\cite{VQGAN,resnet}.}\label{hyper}
\begin{adjustbox}{max width=1\textwidth}
\setlength{\tabcolsep}{1mm}
\vspace{-1.0cm}
\renewcommand{\arraystretch}{1.8}
\begin{tabular}{c|ccc|ccc}
\hline 
\multirow{2}*{\diagbox[width=16em, height=3.8\line]{Dataset}{Hyperparameter}{Algorithm}}&\multicolumn{3}{c|}{Single class based}&\multicolumn{3}{c}{Multi-class based}\\
\cline{2-7}
~ &\makecell{PatchCore\\\cite{patchcore}}&\makecell{DDAD\\ \cite{DDAD}}&CCAD(V)&\makecell{DiAD\\\cite{DiAD}}&CCAD(C)&CCAD(F)\\
\hline  
Hyperparameters&PFE; $\xi$&$eph$*; $\ell$&$eph$*; $\ell$; $\xi$&PFE; $\xi$;$\ell$;$eph$;$Bs$&PFE; $\xi$;$\ell$;$eph$;$Bs$&PFE; $\xi$;$\ell$;$eph$;$Bs$\\
\hline  
MVTec-AD\cite{mvtec}&\makecell{AutoEncoderKL;\\ $1000$}&\makecell{$500-2000$; \\$3 \times 10^{-4}$ }&\makecell{$500-2000$; \\$3 \times 10^{-4}$; $200$}&\makecell{ResNet50; $1000$;\\$1 \times 10^{-4}$;$200$;$12$}&\makecell{ResNet50; $1000$;\\$1 \times 10^{-4}$;$200$;$12$}&\makecell{ResNet50; $1000$;\\$5 \times 10^{-5}$;$200$;$12$}\\
\hline  
VisA\cite{VisA}&\makecell{AutoEncoderKL; \\$1000$}&\makecell{$500-1000$; \\$3 \times 10^{-4}$ }&\makecell{$1000$; \\$3 \times 10^{-4}$; $200$}&\makecell{ResNet50; $1000$;\\$1 \times 10^{-5}$;$150$;$12$}&\makecell{ResNet50; $1000$;\\$4.5 \times 10^{-6}$;$150$;$12$}&\makecell{ResNet50; $1000$;\\$5 \times 10^{-6}$;$150$;$12$}\\
\hline  
MVTec-3d\cite{MVTec-3d}&\makecell{AutoEncoderKL; \\$1000$}&\makecell{$500$\\ $3 \times 10^{-4}$}&\makecell{$500-1000$\\ $3 \times 10^{-4}$; $200$}&\makecell{ResNet50; $1000$;\\$1 \times 10^{-4}$;$100$;$12$}&\makecell{ResNet50; $1000$;\\$5 \times 10^{-5}$;$100$;$12$}&\makecell{ResNet50; $1000$;\\$2 \times 10^{-5}$;$100$;$12$}\\
\hline  
MVTec-loco\cite{MVTec-loco}&\makecell{AutoEncoderKL;\\ $1000$}&\makecell{$500-1000$; $3 \times 10^{-4}$ }&\makecell{$500-1000$;\\ $3 \times 10^{-4}$; $200$}&\makecell{ResNet50; $1000$;\\$1 \times 10^{-5}$;$200$;$12$}&\makecell{ResNet50; $1000$;\\$1 \times 10^{-5}$;$200$;$12$}&\makecell{ResNet50; $1000$;\\$1 \times 10^{-5}$;$200$;$12$}\\
\hline 
MTD\cite{MTD}&\makecell{AutoEncoderKL; \\$1000$}&\makecell{$1000$; \\$3 \times 10^{-4}$ }&\makecell{$1000$; \\$3 \times 10^{-4}$; $200$}&\makecell{ResNet50; $1000$;\\$1 \times 10^{-4}$;$200$;$12$}&\makecell{ResNet50; $1000$;\\$1 \times 10^{-4}$;$200$;$12$}&\makecell{ResNet50; $1000$;\\$5 \times 10^{-5}$;$200$;$12$}\\
\hline 
DAGM 2007\cite{DAGM}&\makecell{AutoEncoderKL; \\$1000$}&\makecell{$1000-2500$; \\$3 \times 10^{-4}$ }&\makecell{$1000-2500$; \\$3 \times 10^{-4}$; $200$}&\makecell{ResNet50; $1000$;\\$1 \times 10^{-5}$;$40$;$12$}&\makecell{ResNet50; $1000$;\\$1 \times 10^{-5}$;$40$;$12$}&\makecell{ResNet50; $1000$;\\$1 \times 10^{-5}$;$40$;$12$}\\
\hline 
different $\xi$ on MVTec-AD&\makecell{$-$}&\makecell{$500-2000$;\\ $3 \times 10^{-4}$}&\makecell{$500-2000$; \\$3 \times 10^{-4}$; $10 ~\& ~200$}&\makecell{$-$}&\makecell{$-$}&\makecell{$-$}\\
\hline 
\end{tabular}
\end{adjustbox}
\end{minipage}
\end{table*}
\section{Data Visualization}
More Qualitative example visualization can be seen in figure \ref{Visualization DAGM 1} and \ref{Visualization Datasets}.
\begin{figure*}[p]
\begin{subfigure}[t]{0.49\linewidth}
\centering
\begin{minipage}[t]{0.23\textwidth}
\centering
\includegraphics[width=2.0cm]{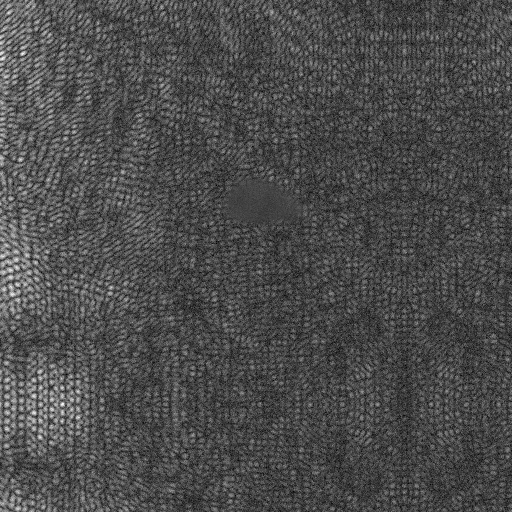}
\end{minipage}
\begin{minipage}[t]{0.23\textwidth}
\centering
\includegraphics[width=2.0cm]{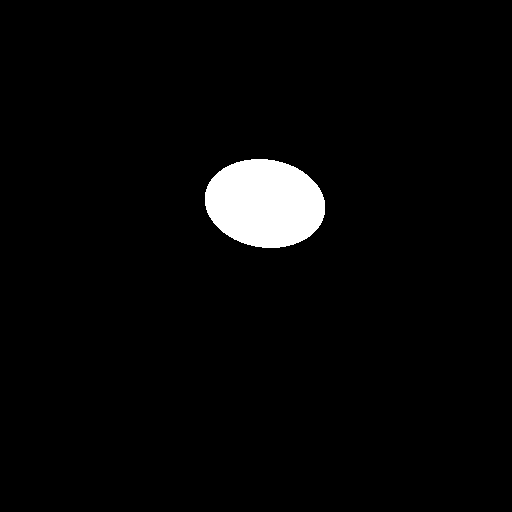}
\end{minipage}
\begin{minipage}[t]{0.23\textwidth}
\centering
\includegraphics[width=2.0cm, height=2.0cm]{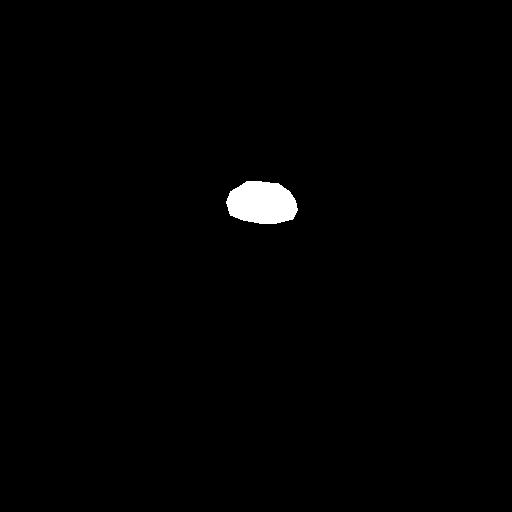}
\end{minipage}
\begin{minipage}[t]{0.23\textwidth}
\centering
\includegraphics[width=2.0cm]{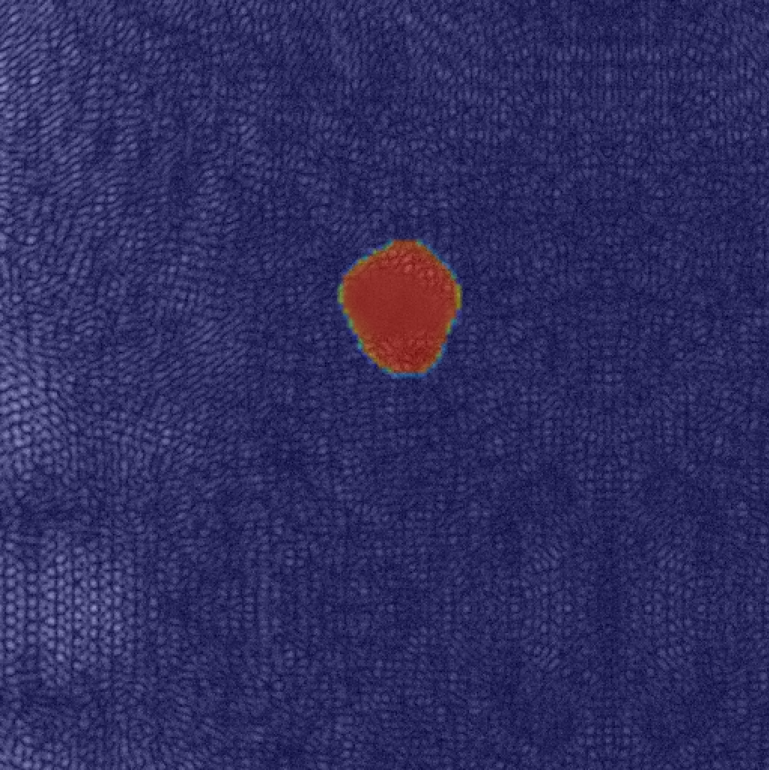}
\end{minipage}
\centering
\begin{minipage}[t]{0.23\textwidth}
\centering
\includegraphics[width=2.0cm]{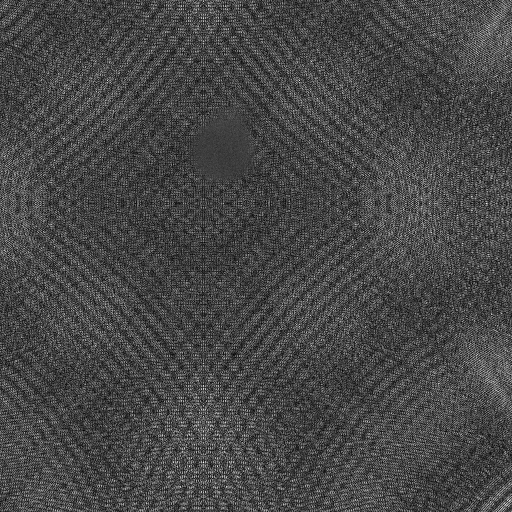}
\end{minipage}
\begin{minipage}[t]{0.23\textwidth}
\centering
\includegraphics[width=2.0cm]{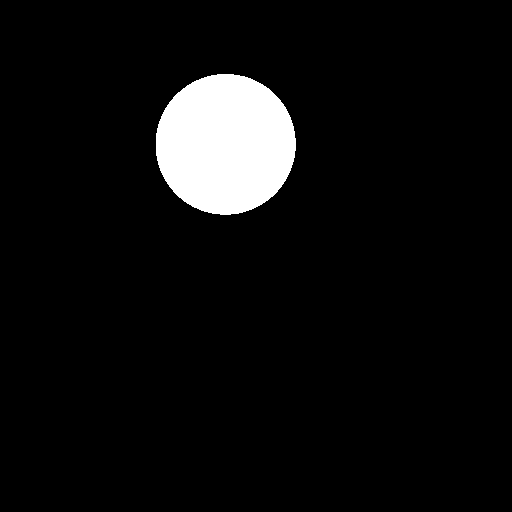}
\end{minipage}
\begin{minipage}[t]{0.23\textwidth}
\centering
\includegraphics[width=2.0cm, height=2.0cm]{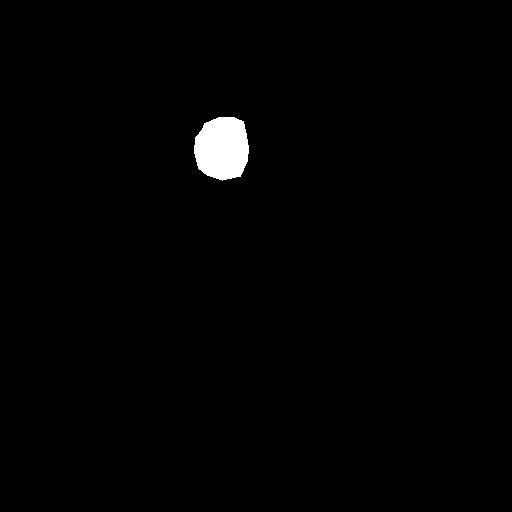}
\end{minipage}
\begin{minipage}[t]{0.23\textwidth}
\centering
\includegraphics[width=2.0cm]{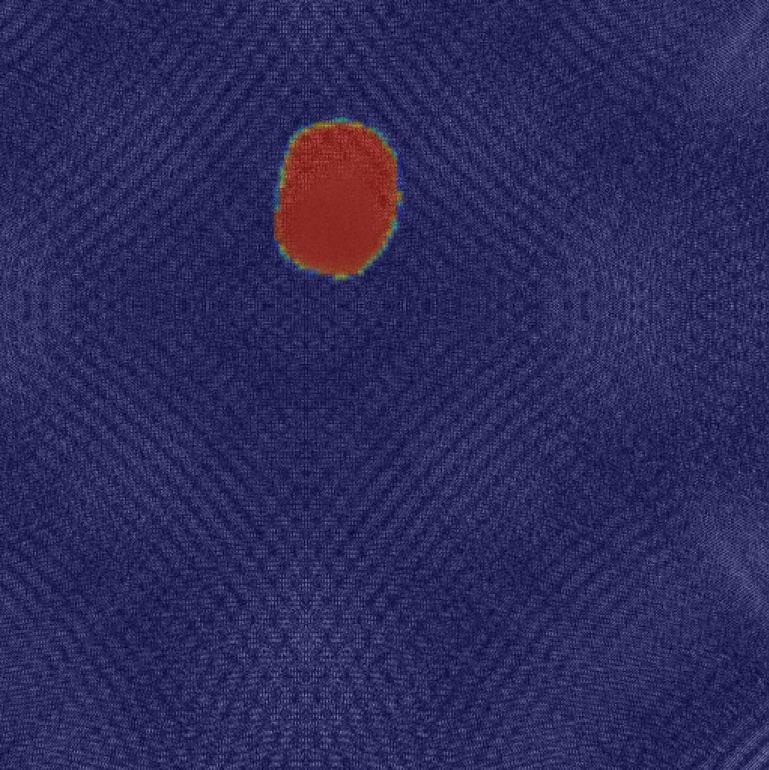}
\end{minipage}
\centering
\begin{minipage}[t]{0.23\textwidth}
\centering
\includegraphics[width=2.0cm]{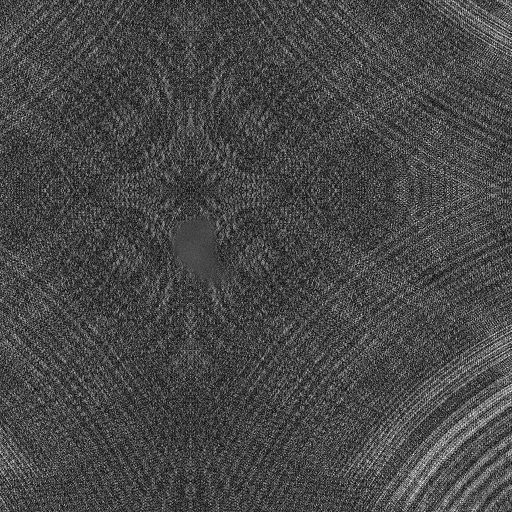}
\end{minipage}
\begin{minipage}[t]{0.23\textwidth}
\centering
\includegraphics[width=2.0cm]{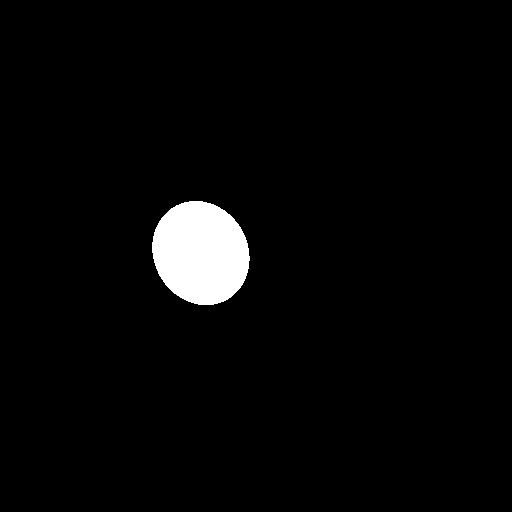}
\end{minipage}
\begin{minipage}[t]{0.23\textwidth}
\centering
\includegraphics[width=2.0cm, height=2.0cm]{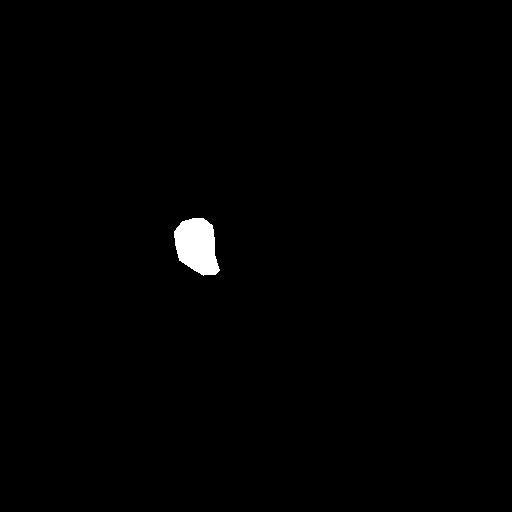}
\end{minipage}
\begin{minipage}[t]{0.23\textwidth}
\centering
\includegraphics[width=2.0cm]{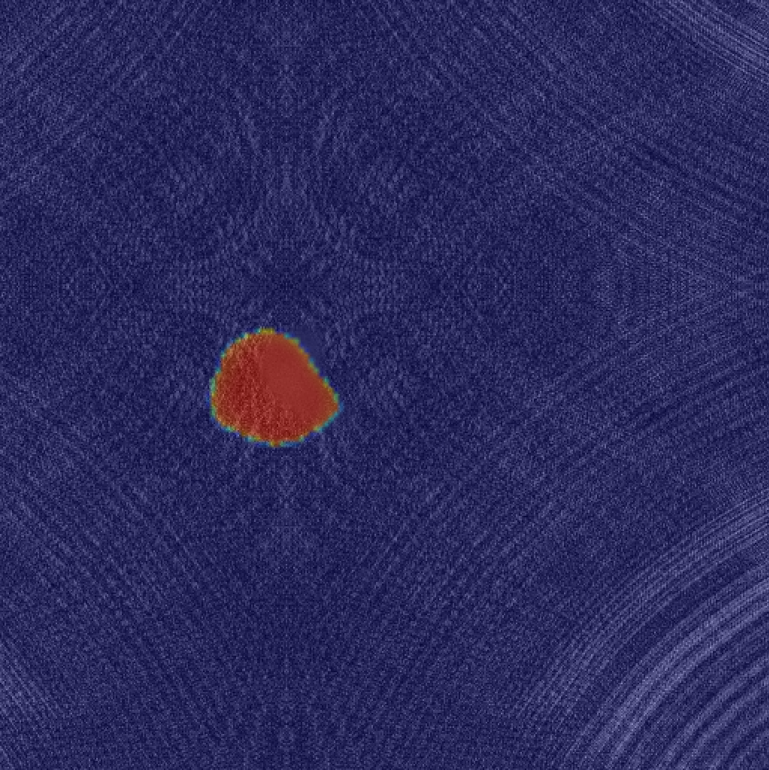}
\end{minipage}
\centering
\begin{minipage}[t]{0.23\textwidth}
\centering
\includegraphics[width=2.0cm]{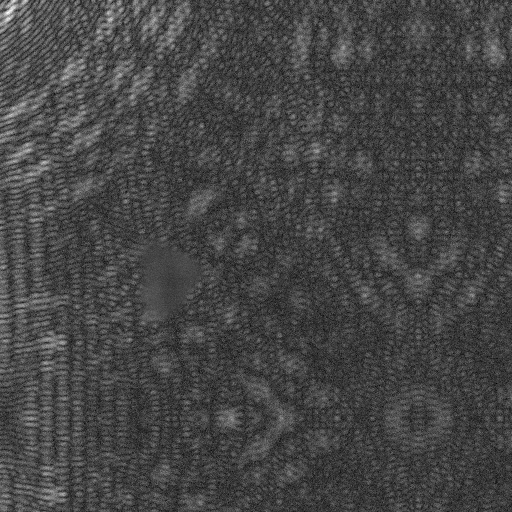}
\end{minipage}
\begin{minipage}[t]{0.23\textwidth}
\centering
\includegraphics[width=2.0cm]{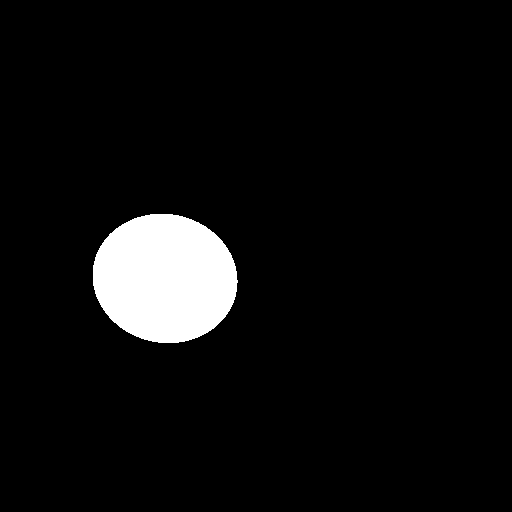}
\end{minipage}
\begin{minipage}[t]{0.23\textwidth}
\centering
\includegraphics[width=2.0cm, height=2.0cm]{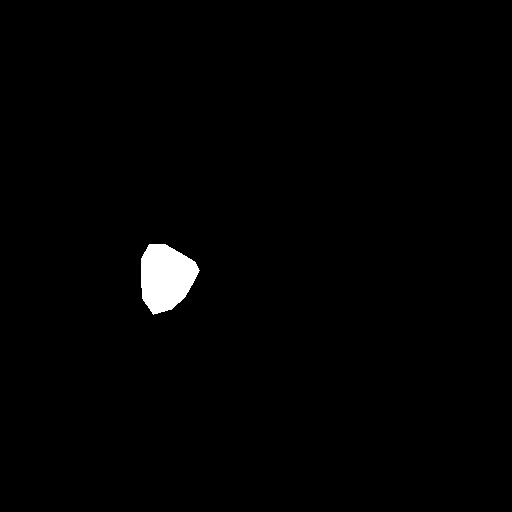}
\end{minipage}
\begin{minipage}[t]{0.23\textwidth}
\centering
\includegraphics[width=2.0cm]{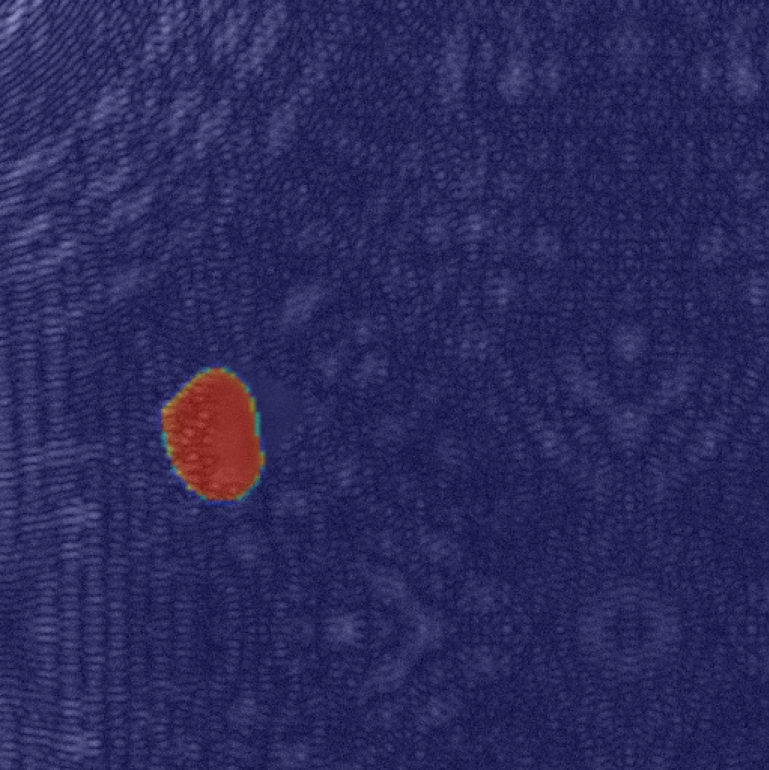}
\end{minipage}
\begin{minipage}[t]{0.23\textwidth}
\centering
Image
\end{minipage}
\begin{minipage}[t]{0.23\textwidth}
\centering
Label (old)
\end{minipage}
\begin{minipage}[t]{0.23\textwidth}
\centering
Label (new)
\end{minipage}
\begin{minipage}[t]{0.23\textwidth}
\centering
Heat Map
\end{minipage}
\centering
\captionsetup{justification=centering}
\subcaption[]{Visualization of DAGM Class 1}

\end{subfigure}
\begin{subfigure}[t]{0.49\linewidth}
\centering
\begin{minipage}[t]{0.23\textwidth}
\centering
\includegraphics[width=2.0cm]{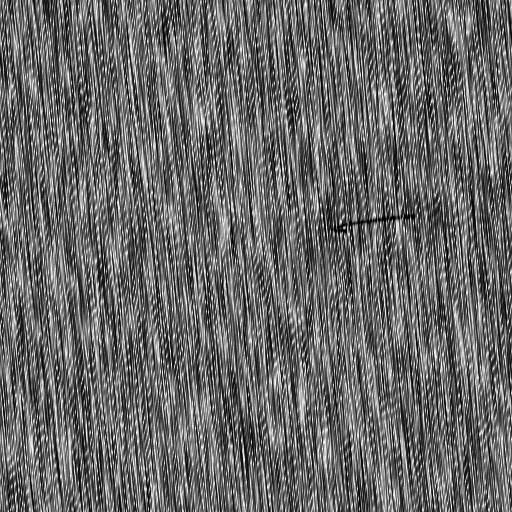}
\end{minipage}
\begin{minipage}[t]{0.23\textwidth}
\centering
\includegraphics[width=2.0cm]{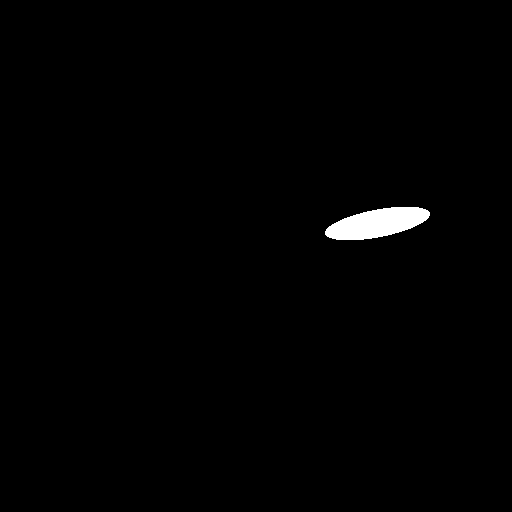}
\end{minipage}
\begin{minipage}[t]{0.23\textwidth}
\centering
\includegraphics[width=2.0cm]{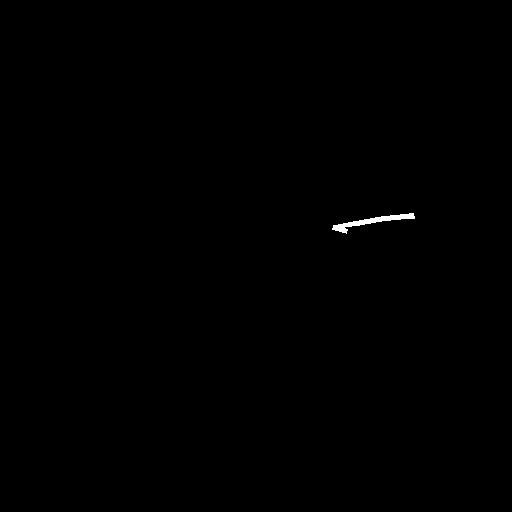}
\end{minipage}
\begin{minipage}[t]{0.23\textwidth}
\centering
\includegraphics[width=2.0cm]{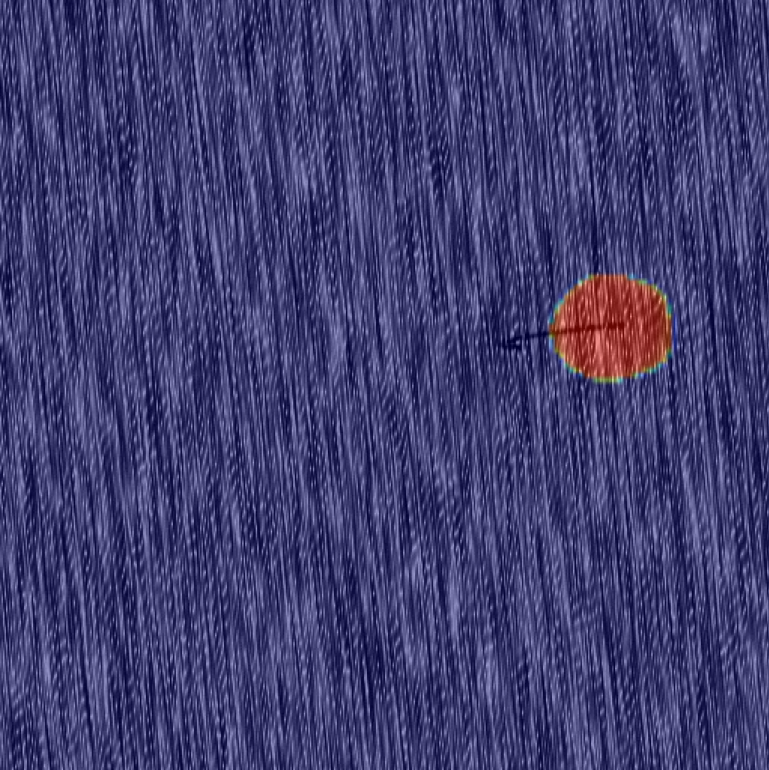}
\end{minipage}
\centering
\begin{minipage}[t]{0.23\textwidth}
\centering
\includegraphics[width=2.0cm]{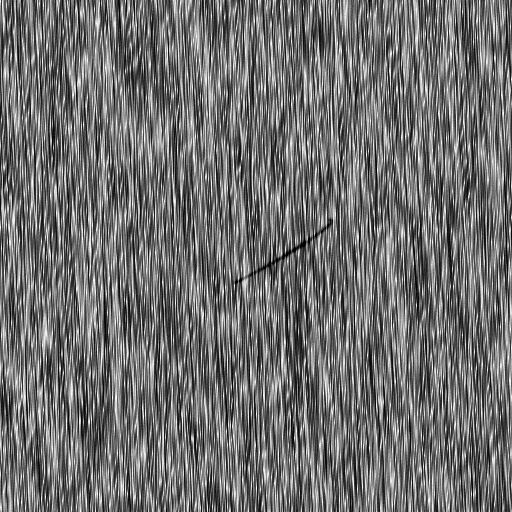}
\end{minipage}
\begin{minipage}[t]{0.23\textwidth}
\centering
\includegraphics[width=2.0cm]{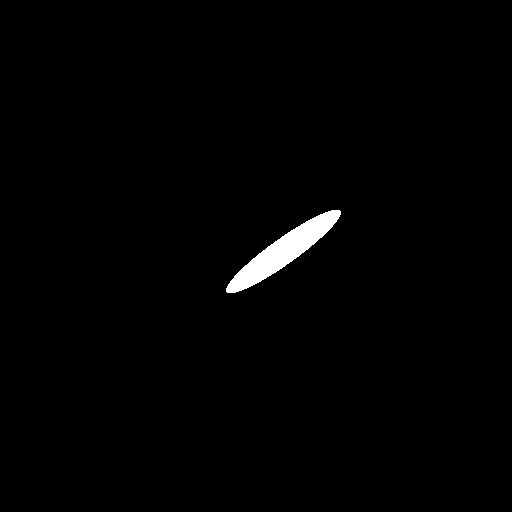}
\end{minipage}
\begin{minipage}[t]{0.23\textwidth}
\centering
\includegraphics[width=2.0cm]{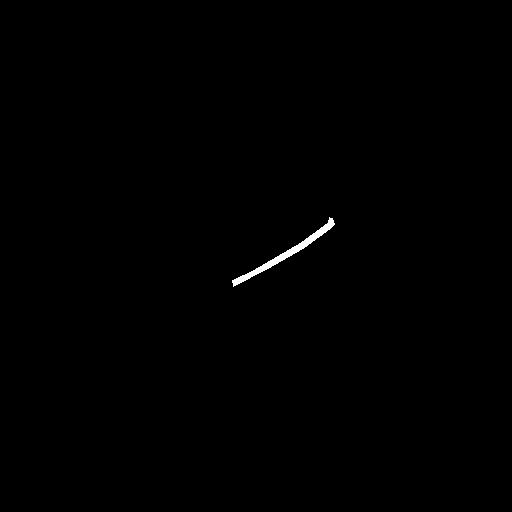}
\end{minipage}
\begin{minipage}[t]{0.23\textwidth}
\centering
\includegraphics[width=2.0cm]{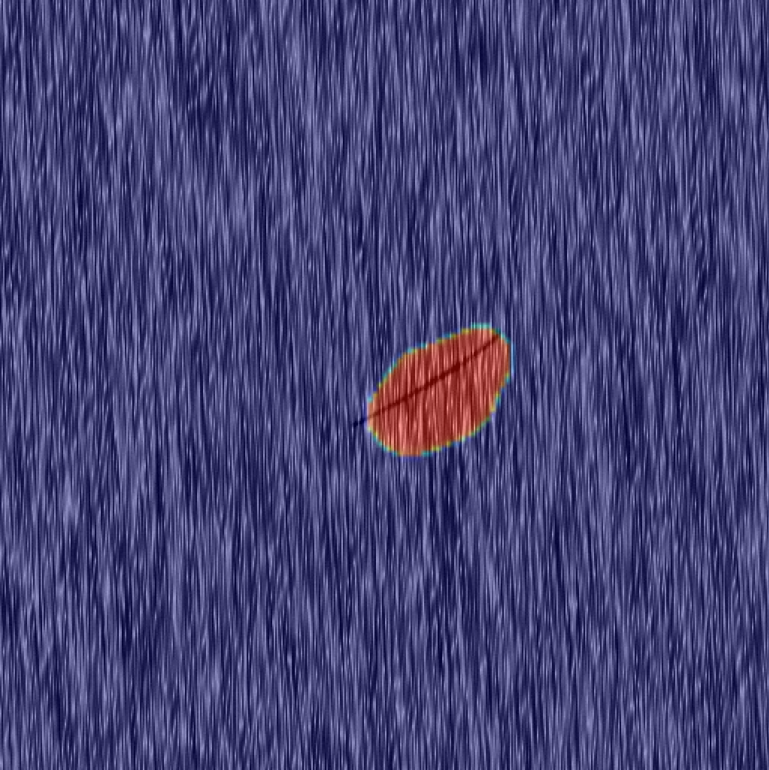}
\end{minipage}
\centering
\begin{minipage}[t]{0.23\textwidth}
\centering
\includegraphics[width=2.0cm]{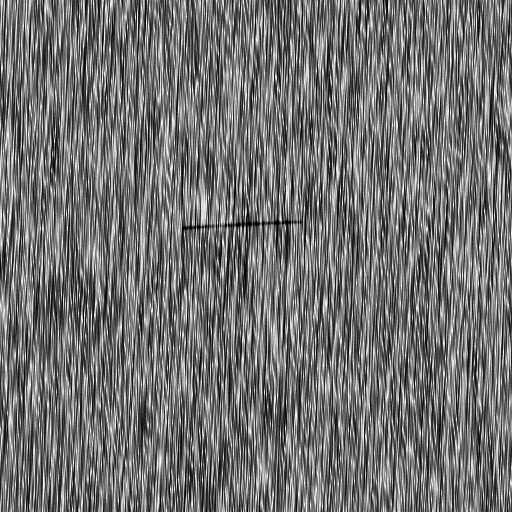}
\end{minipage}
\begin{minipage}[t]{0.23\textwidth}
\centering
\includegraphics[width=2.0cm]{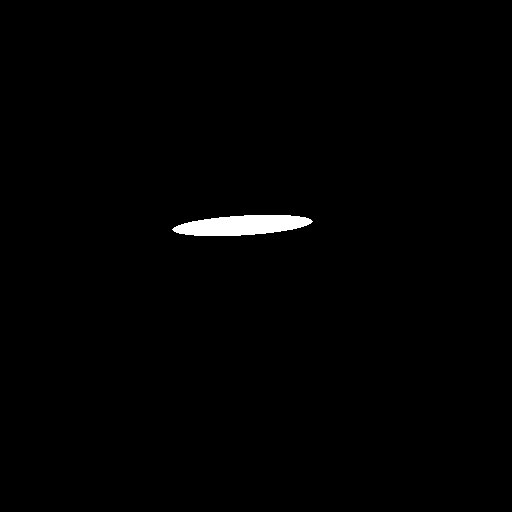}
\end{minipage}
\begin{minipage}[t]{0.23\textwidth}
\centering
\includegraphics[width=2.0cm]{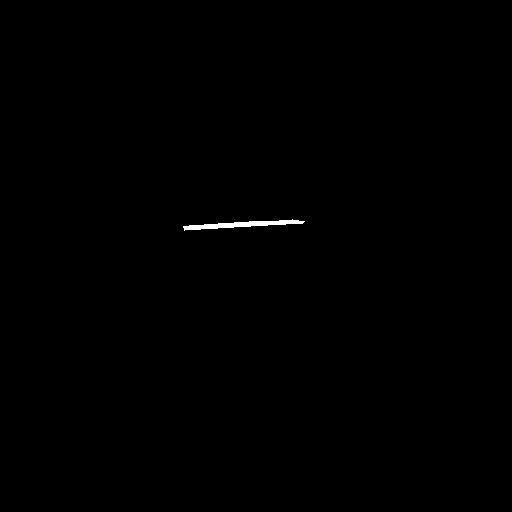}
\end{minipage}
\begin{minipage}[t]{0.23\textwidth}
\centering
\includegraphics[width=2.0cm]{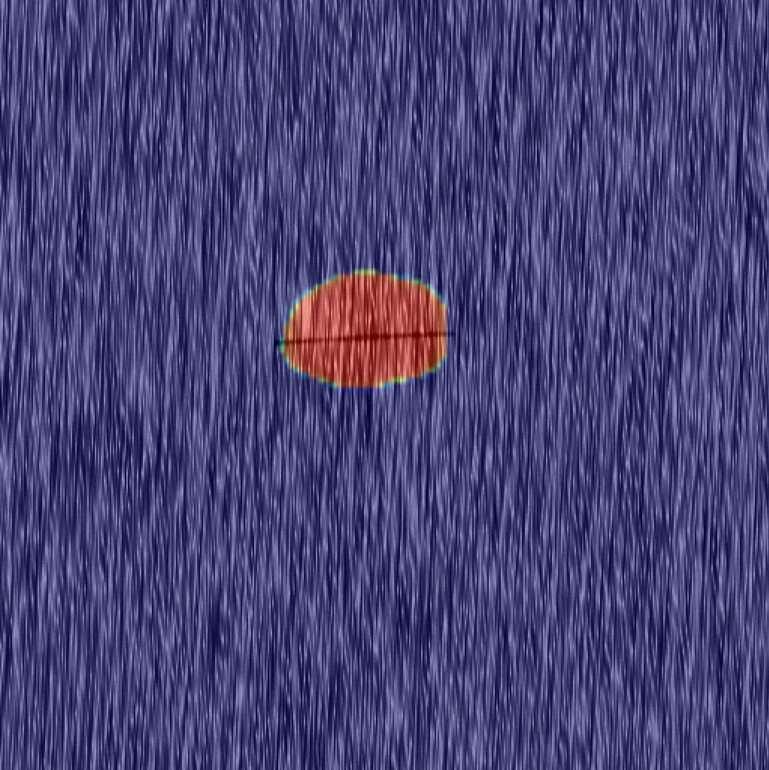}
\end{minipage}
\centering
\begin{minipage}[t]{0.23\textwidth}
\centering
\includegraphics[width=2.0cm]{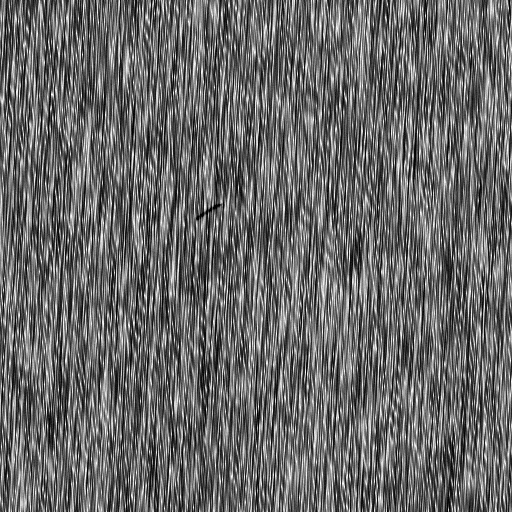}
\end{minipage}
\begin{minipage}[t]{0.23\textwidth}
\centering
\includegraphics[width=2.0cm]{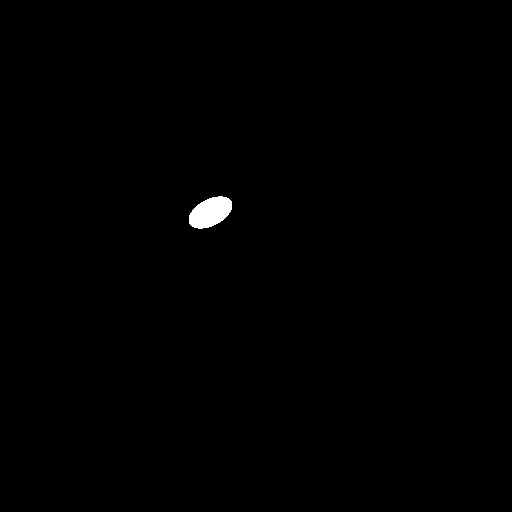}
\end{minipage}
\begin{minipage}[t]{0.23\textwidth}
\centering
\includegraphics[width=2.0cm]{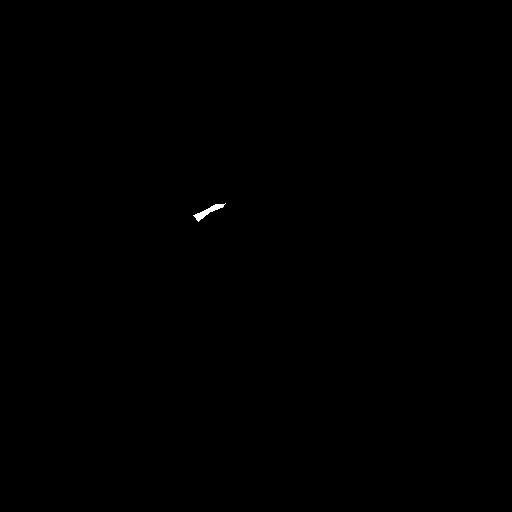}
\end{minipage}
\begin{minipage}[t]{0.23\textwidth}
\centering
\includegraphics[width=2.0cm]{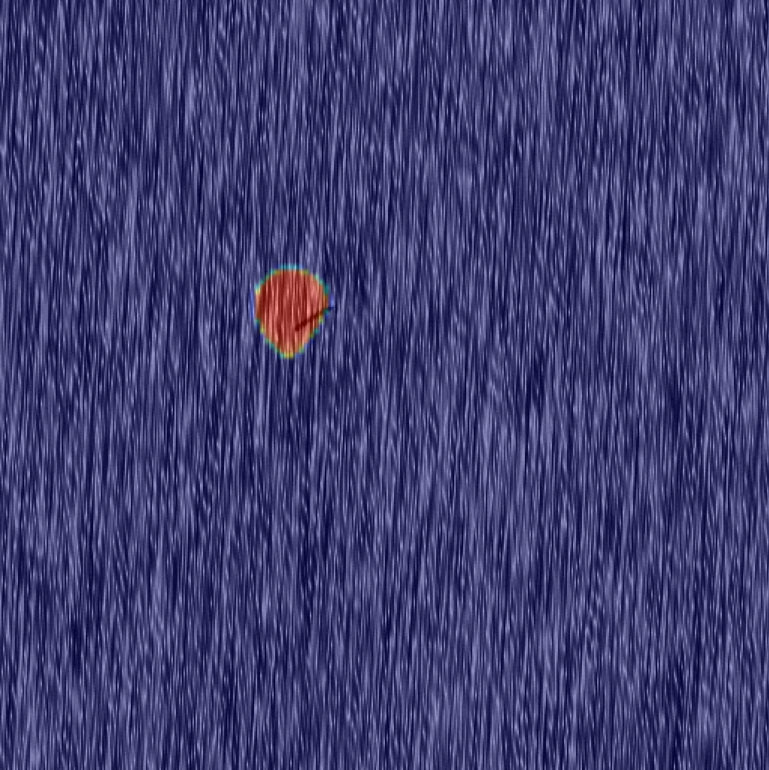}
\end{minipage}
\begin{minipage}[t]{0.23\textwidth}
\centering
Image
\end{minipage}
\begin{minipage}[t]{0.23\textwidth}
\centering
Label (old)
\end{minipage}
\begin{minipage}[t]{0.23\textwidth}
\centering
Label (new)
\end{minipage}
\begin{minipage}[t]{0.23\textwidth}
\centering
Heat Map
\end{minipage}
\centering
\captionsetup{justification=centering}
\subcaption[]{Visualization of DAGM Class 2}
\end{subfigure}
\end{figure*}
\begin{figure*}[p]\ContinuedFloat
\begin{subfigure}[t]{0.49\linewidth}
\centering
\begin{minipage}[t]{0.23\textwidth}
\centering
\includegraphics[width=2.0cm]{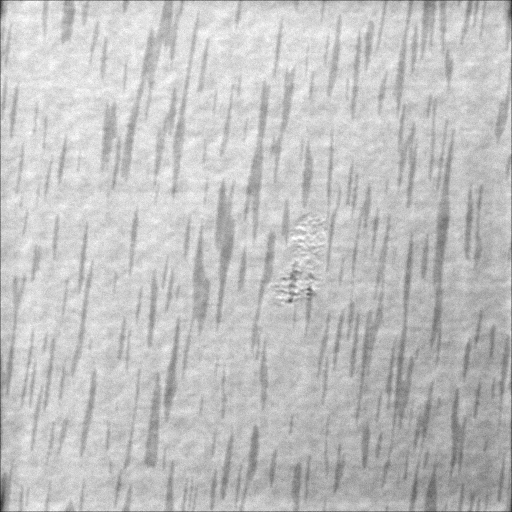}
\end{minipage}
\begin{minipage}[t]{0.23\textwidth}
\centering
\includegraphics[width=2.0cm]{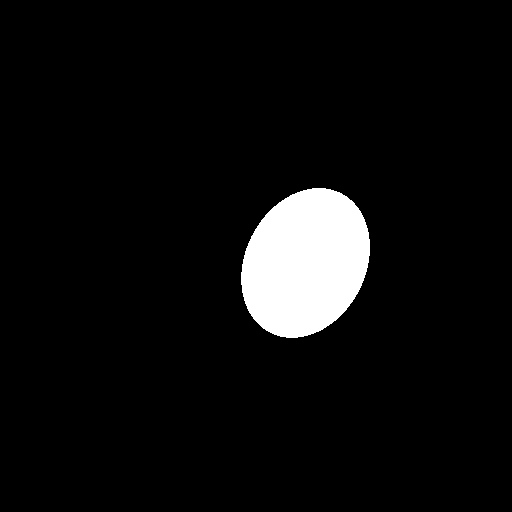}
\end{minipage}
\begin{minipage}[t]{0.23\textwidth}
\centering
\includegraphics[width=2.0cm]{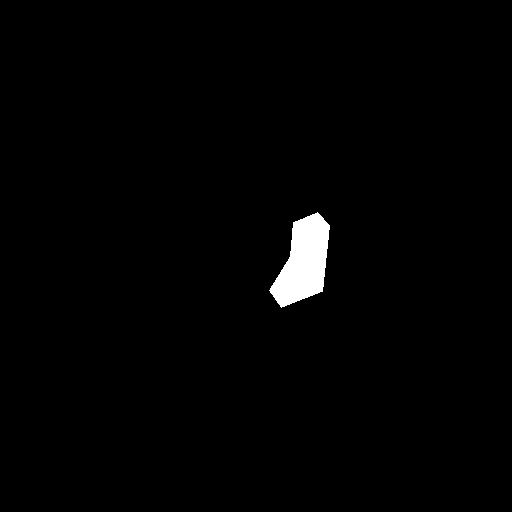}
\end{minipage}
\begin{minipage}[t]{0.23\textwidth}
\centering
\includegraphics[width=2.0cm]{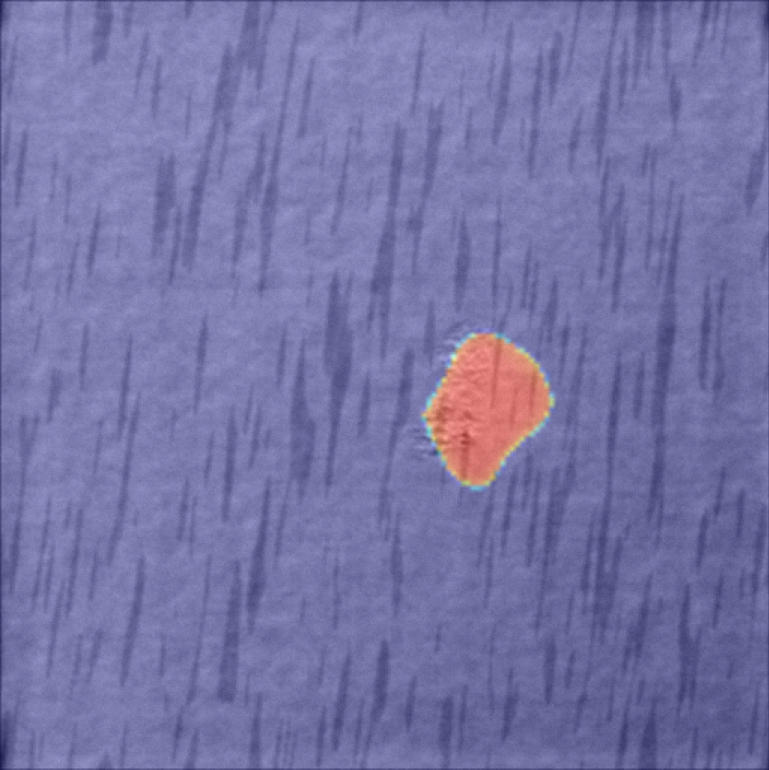}
\end{minipage}
\centering
\begin{minipage}[t]{0.23\textwidth}
\centering
\includegraphics[width=2.0cm]{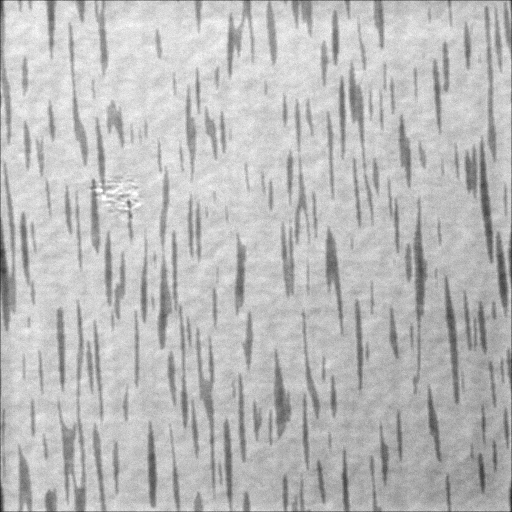}
\end{minipage}
\begin{minipage}[t]{0.23\textwidth}
\centering
\includegraphics[width=2.0cm]{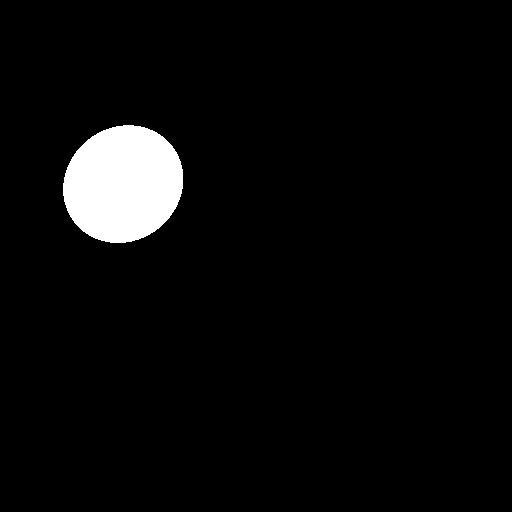}
\end{minipage}
\begin{minipage}[t]{0.23\textwidth}
\centering
\includegraphics[width=2.0cm]{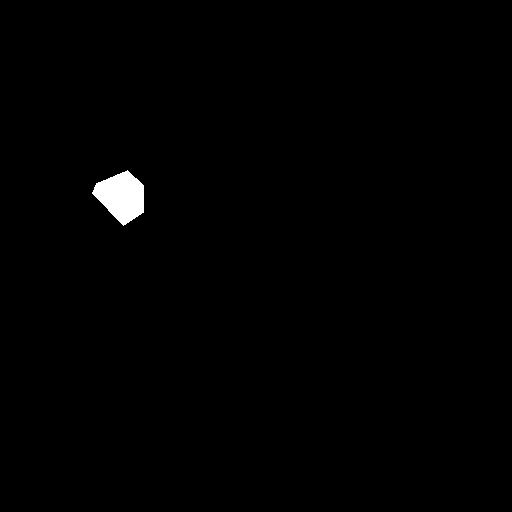}
\end{minipage}
\begin{minipage}[t]{0.23\textwidth}
\centering
\includegraphics[width=2.0cm]{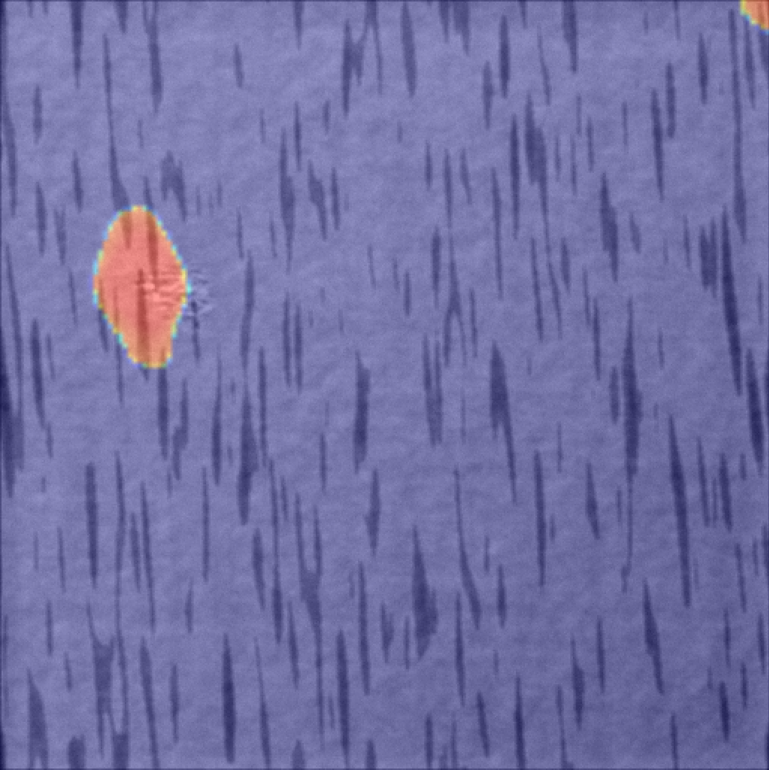}
\end{minipage}
\centering
\begin{minipage}[t]{0.23\textwidth}
\centering
\includegraphics[width=2.0cm]{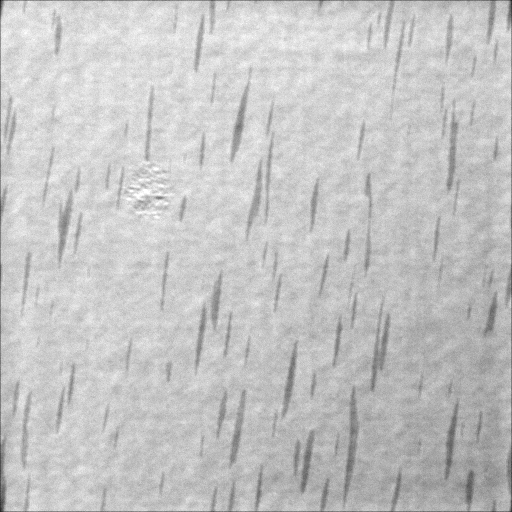}
\end{minipage}
\begin{minipage}[t]{0.23\textwidth}
\centering
\includegraphics[width=2.0cm]{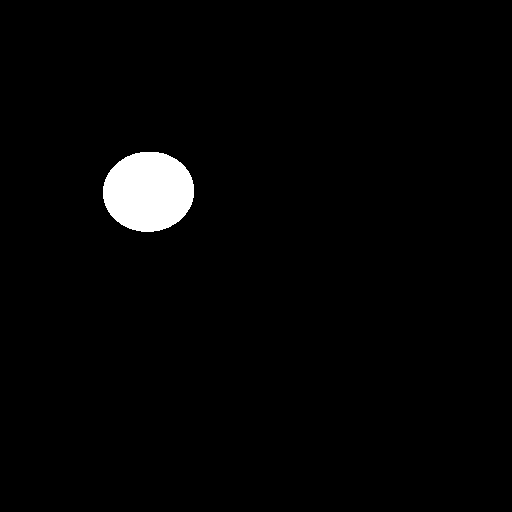}
\end{minipage}
\begin{minipage}[t]{0.23\textwidth}
\centering
\includegraphics[width=2.0cm]{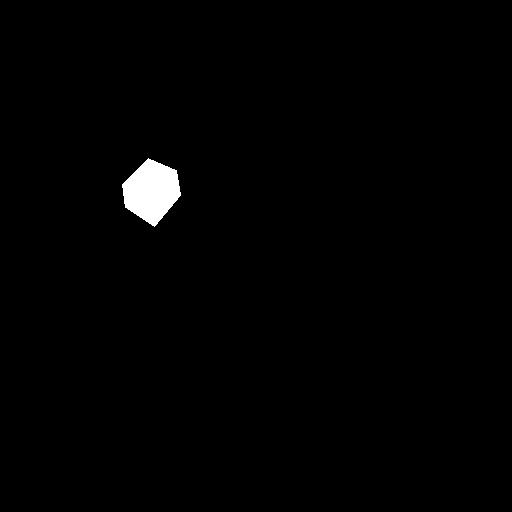}
\end{minipage}
\begin{minipage}[t]{0.23\textwidth}
\centering
\includegraphics[width=2.0cm]{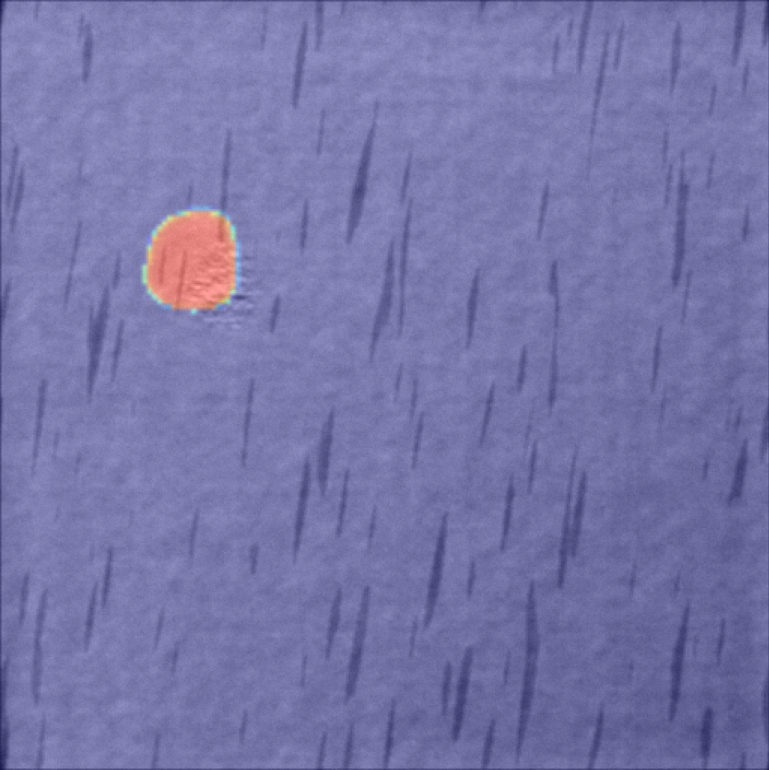}
\end{minipage}
\centering
\begin{minipage}[t]{0.23\textwidth}
\centering
\includegraphics[width=2.0cm]{exp_figure/DAGM2007_patchcore_anomaly_map/class7_blur/0038.PNG}
\end{minipage}
\begin{minipage}[t]{0.23\textwidth}
\centering
\includegraphics[width=2.0cm]{exp_figure/DAGM2007_patchcore_anomaly_map/class7_blur/0038_label.PNG}
\end{minipage}
\begin{minipage}[t]{0.23\textwidth}
\centering
\includegraphics[width=2.0cm]{exp_figure/DAGM2007_patchcore_anomaly_map/class7_blur/0038.jpg}
\end{minipage}
\begin{minipage}[t]{0.23\textwidth}
\centering
\includegraphics[width=2.0cm]{exp_figure/DAGM2007_patchcore_anomaly_map/class7_blur/0038_anomal_map.png}
\end{minipage}
\begin{minipage}[t]{0.23\textwidth}
\centering
Image
\end{minipage}
\begin{minipage}[t]{0.23\textwidth}
\centering
Label (old)
\end{minipage}
\begin{minipage}[t]{0.23\textwidth}
\centering
Label (new)
\end{minipage}
\begin{minipage}[t]{0.23\textwidth}
\centering
Heat Map
\end{minipage}
\centering
\captionsetup{justification=centering}
\subcaption[]{Visualization of DAGM Class 7}

\end{subfigure}
\begin{subfigure}[t]{0.49\linewidth}
\centering
\begin{minipage}[t]{0.23\textwidth}
\centering
\includegraphics[width=2.0cm]{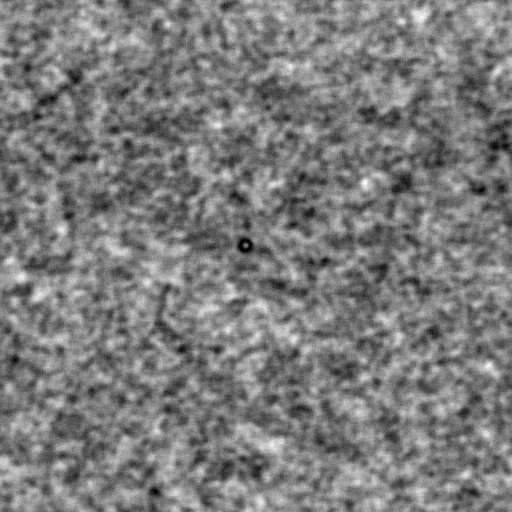}
\end{minipage}
\begin{minipage}[t]{0.23\textwidth}
\centering
\includegraphics[width=2.0cm]{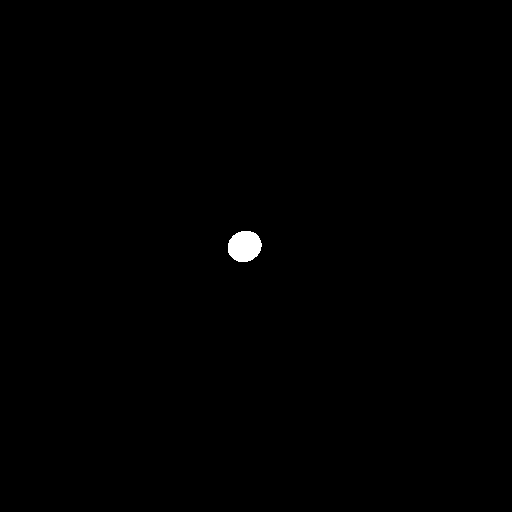}
\end{minipage}
\begin{minipage}[t]{0.23\textwidth}
\centering
\includegraphics[width=2.0cm]{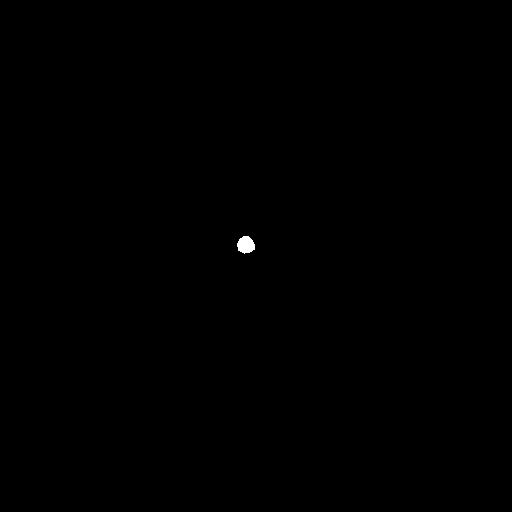}
\end{minipage}
\begin{minipage}[t]{0.23\textwidth}
\centering
\includegraphics[width=2.0cm]{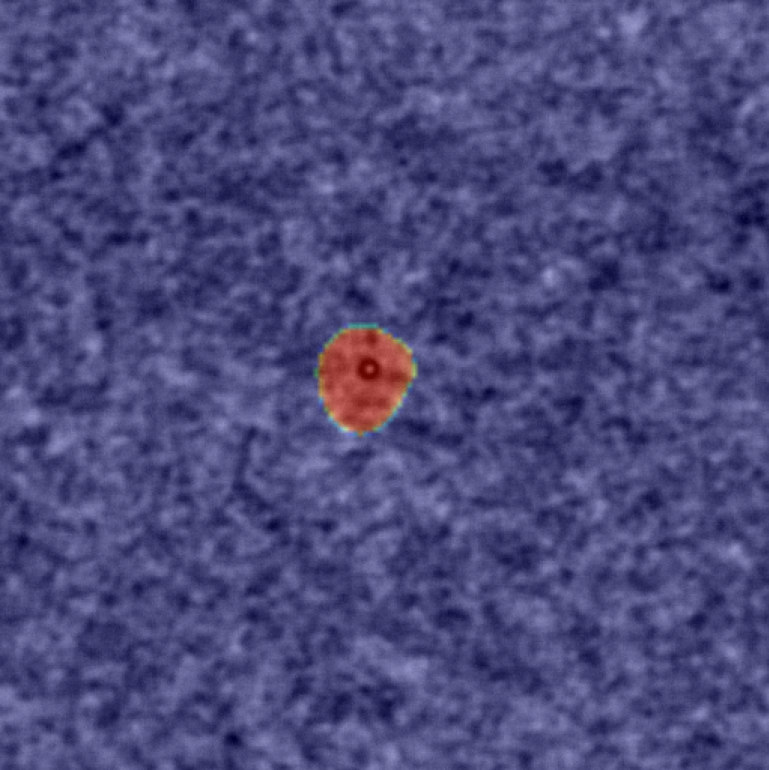}
\end{minipage}
\centering
\begin{minipage}[t]{0.23\textwidth}
\centering
\includegraphics[width=2.0cm]{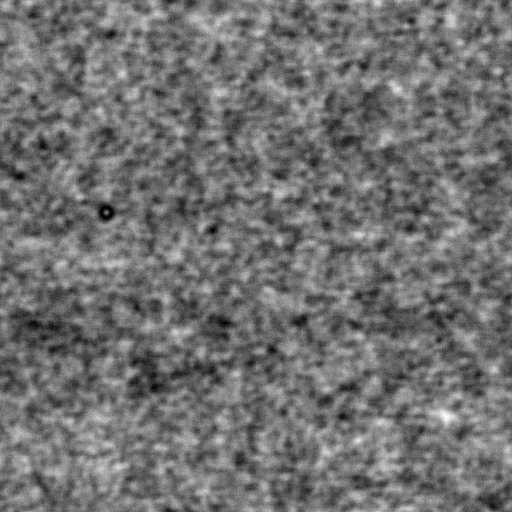}
\end{minipage}
\begin{minipage}[t]{0.23\textwidth}
\centering
\includegraphics[width=2.0cm]{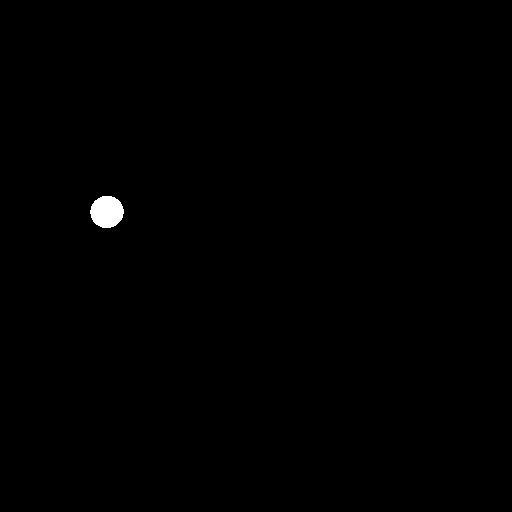}
\end{minipage}
\begin{minipage}[t]{0.23\textwidth}
\centering
\includegraphics[width=2.0cm]{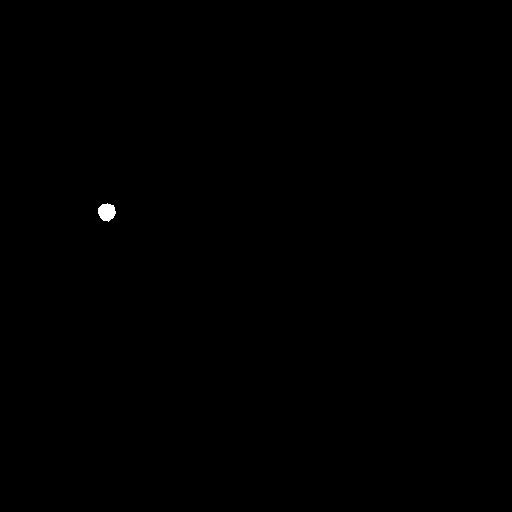}
\end{minipage}
\begin{minipage}[t]{0.23\textwidth}
\centering
\includegraphics[width=2.0cm]{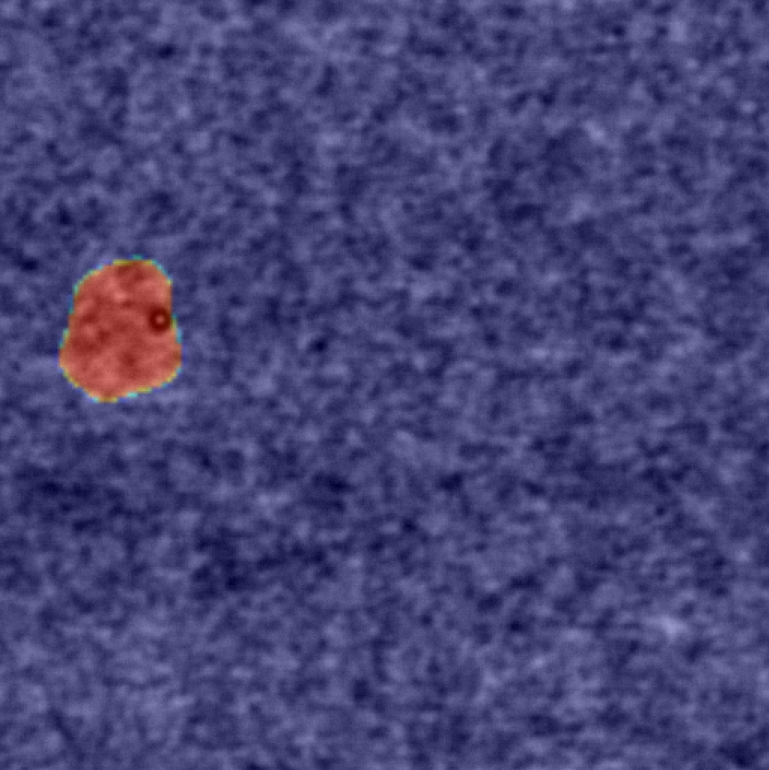}
\end{minipage}
\centering
\begin{minipage}[t]{0.23\textwidth}
\centering
\includegraphics[width=2.0cm]{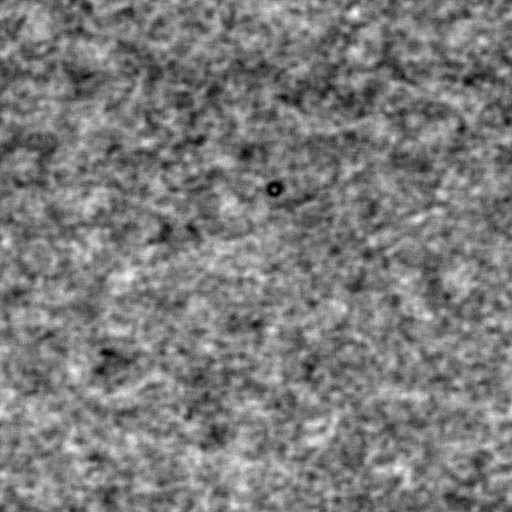}
\end{minipage}
\begin{minipage}[t]{0.23\textwidth}
\centering
\includegraphics[width=2.0cm]{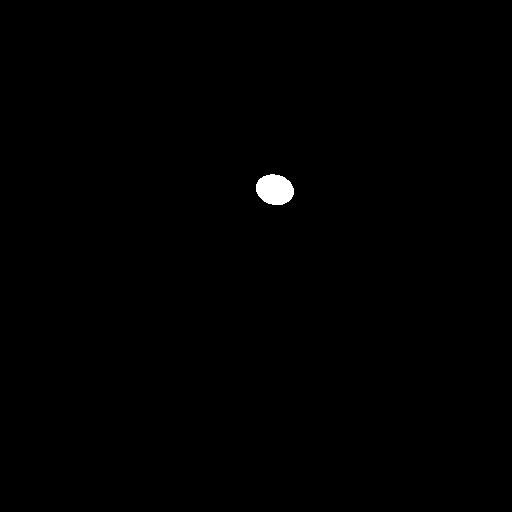}
\end{minipage}
\begin{minipage}[t]{0.23\textwidth}
\centering
\includegraphics[width=2.0cm]{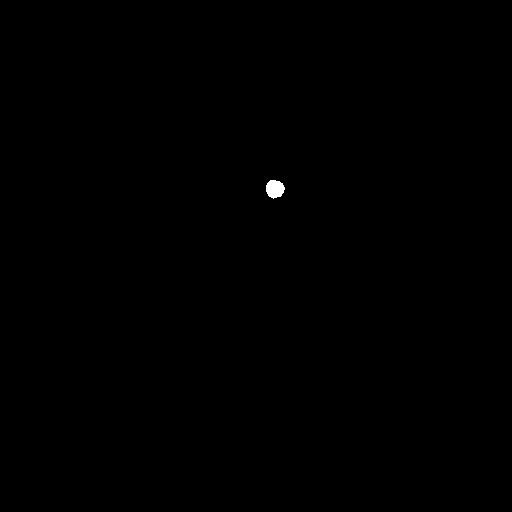}
\end{minipage}
\begin{minipage}[t]{0.23\textwidth}
\centering
\includegraphics[width=2.0cm]{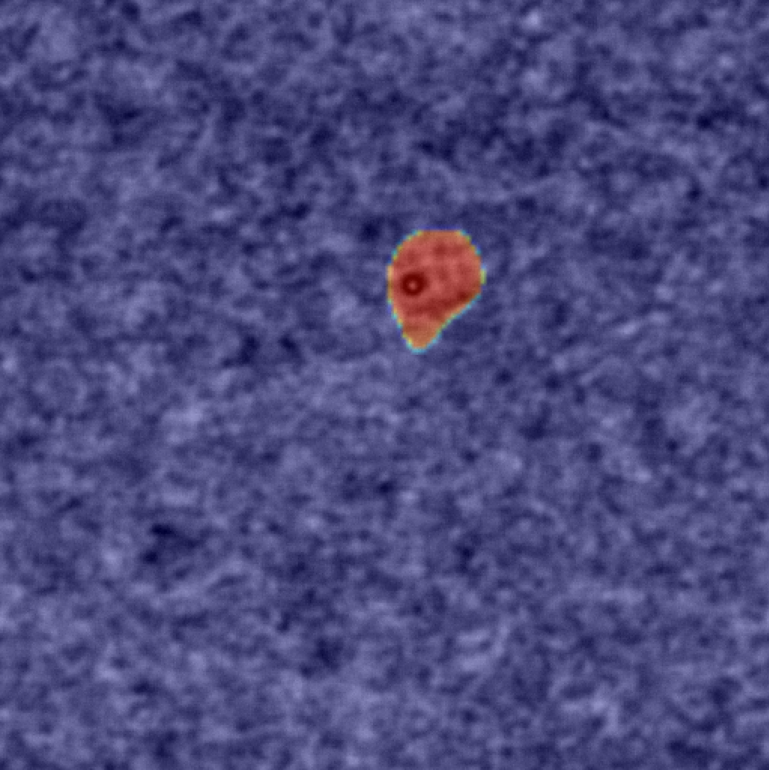}
\end{minipage}
\centering
\begin{minipage}[t]{0.23\textwidth}
\centering
\includegraphics[width=2.0cm]{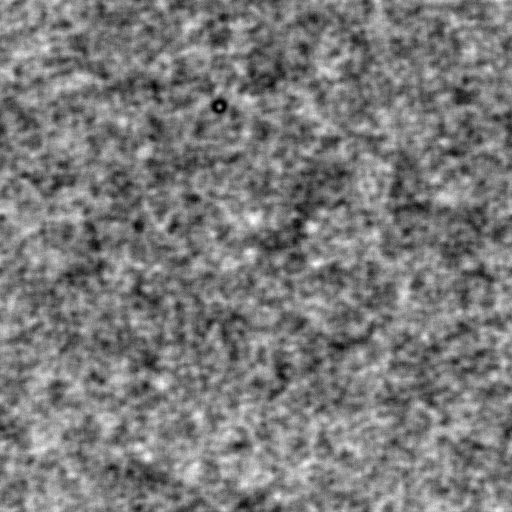}
\end{minipage}
\begin{minipage}[t]{0.23\textwidth}
\centering
\includegraphics[width=2.0cm]{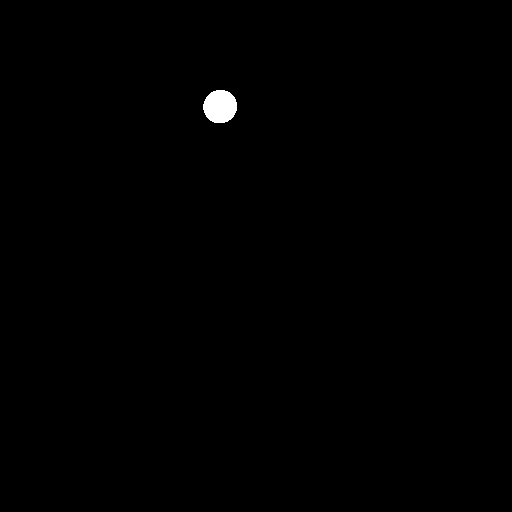}
\end{minipage}
\begin{minipage}[t]{0.23\textwidth}
\centering
\includegraphics[width=2.0cm]{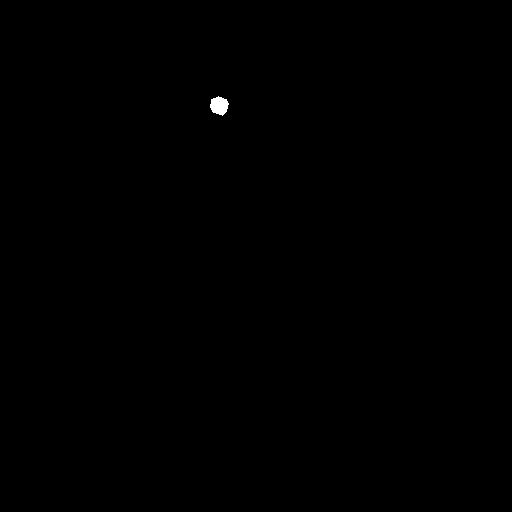}
\end{minipage}
\begin{minipage}[t]{0.23\textwidth}
\centering
\includegraphics[width=2.0cm]{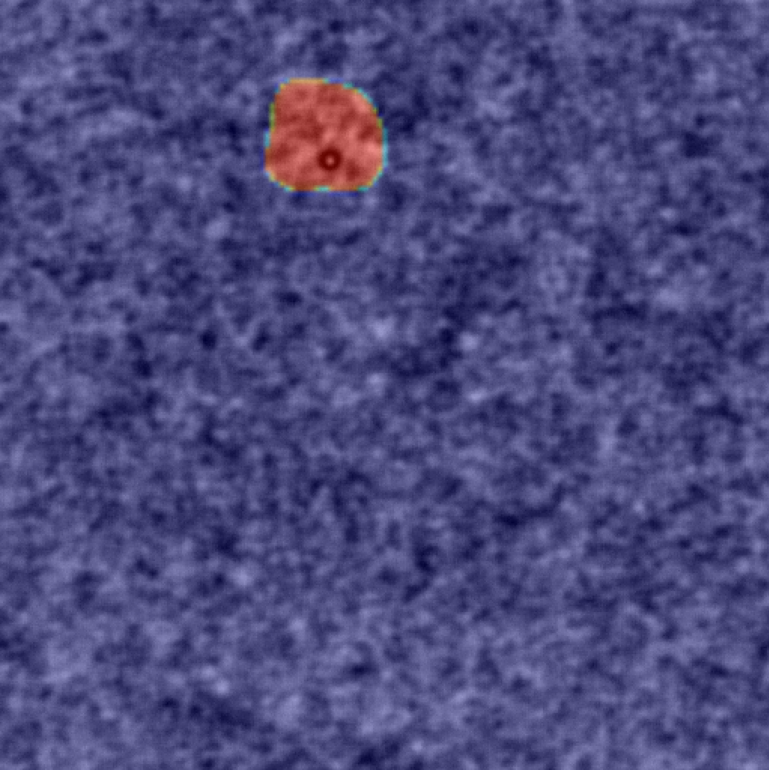}
\end{minipage}
\begin{minipage}[t]{0.23\textwidth}
\centering
Image
\end{minipage}
\begin{minipage}[t]{0.23\textwidth}
\centering
Label (old)
\end{minipage}
\begin{minipage}[t]{0.23\textwidth}
\centering
Label (new)
\end{minipage}
\begin{minipage}[t]{0.23\textwidth}
\centering
Heat Map
\end{minipage}
\centering
\captionsetup{justification=centering}
\subcaption[]{Visualization of DAGM Class 2}

\end{subfigure}
\vspace{-2mm}
\caption{Qualitative example visualization on DAGM}
\label{Visualization DAGM 1}
\end{figure*}
\begin{figure*}[p]
\begin{subfigure}[t]{0.49\linewidth}
\centering
\begin{minipage}[t]{0.23\textwidth}
\centering
\includegraphics[width=2.0cm]{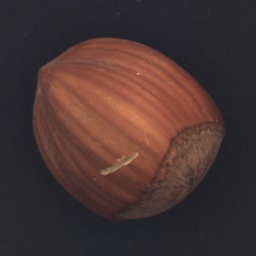}
\end{minipage}
\begin{minipage}[t]{0.23\textwidth}
\centering
\includegraphics[width=2.0cm]{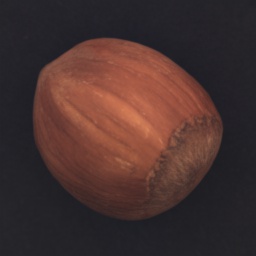}
\end{minipage}
\begin{minipage}[t]{0.23\textwidth}
\centering
\includegraphics[width=2.0cm, height=2.0cm]{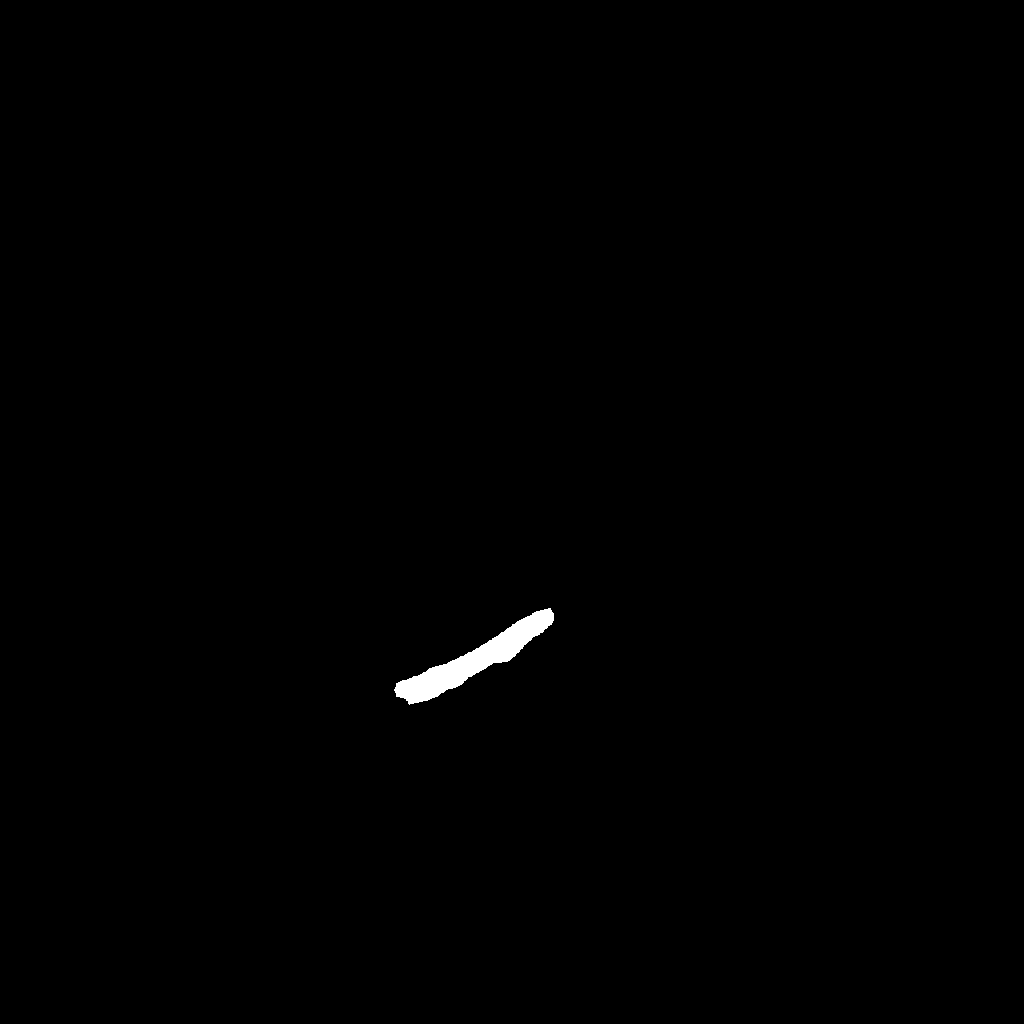}
\end{minipage}
\begin{minipage}[t]{0.23\textwidth}
\centering
\includegraphics[width=2.0cm]{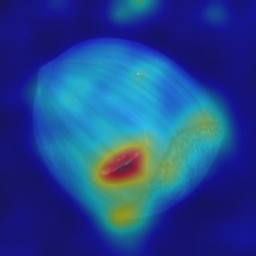}
\end{minipage}
\centering
\begin{minipage}[t]{0.23\textwidth}
\centering
\includegraphics[width=2.0cm]{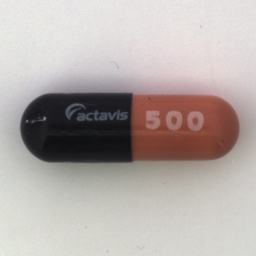}
\end{minipage}
\begin{minipage}[t]{0.23\textwidth}
\centering
\includegraphics[width=2.0cm]{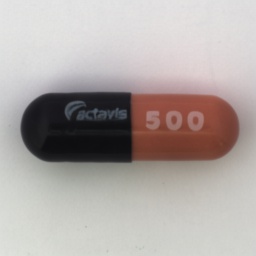}
\end{minipage}
\begin{minipage}[t]{0.23\textwidth}
\centering
\includegraphics[width=2.0cm, height=2.0cm]{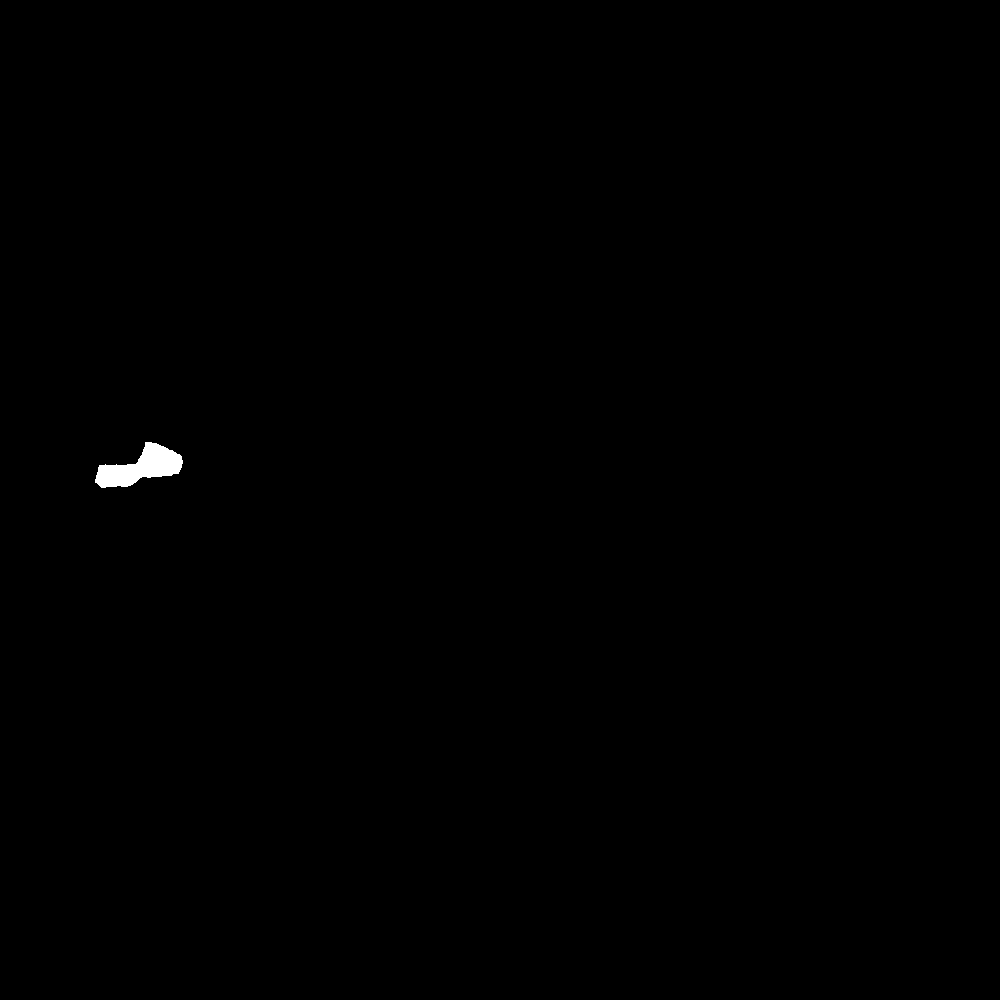}
\end{minipage}
\begin{minipage}[t]{0.23\textwidth}
\centering
\includegraphics[width=2.0cm]{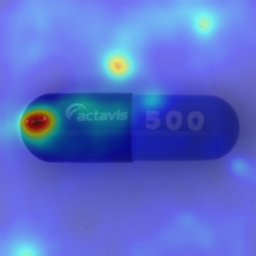}
\end{minipage}
\centering
\begin{minipage}[t]{0.23\textwidth}
\centering
\includegraphics[width=2.0cm]{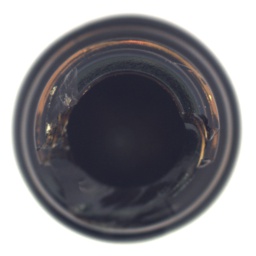}
\end{minipage}
\begin{minipage}[t]{0.23\textwidth}
\centering
\includegraphics[width=2.0cm]{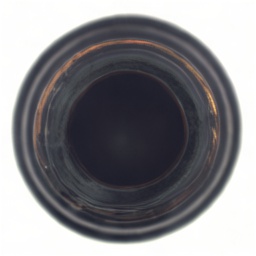}
\end{minipage}
\begin{minipage}[t]{0.23\textwidth}
\centering
\includegraphics[width=2.0cm, height=2.0cm]{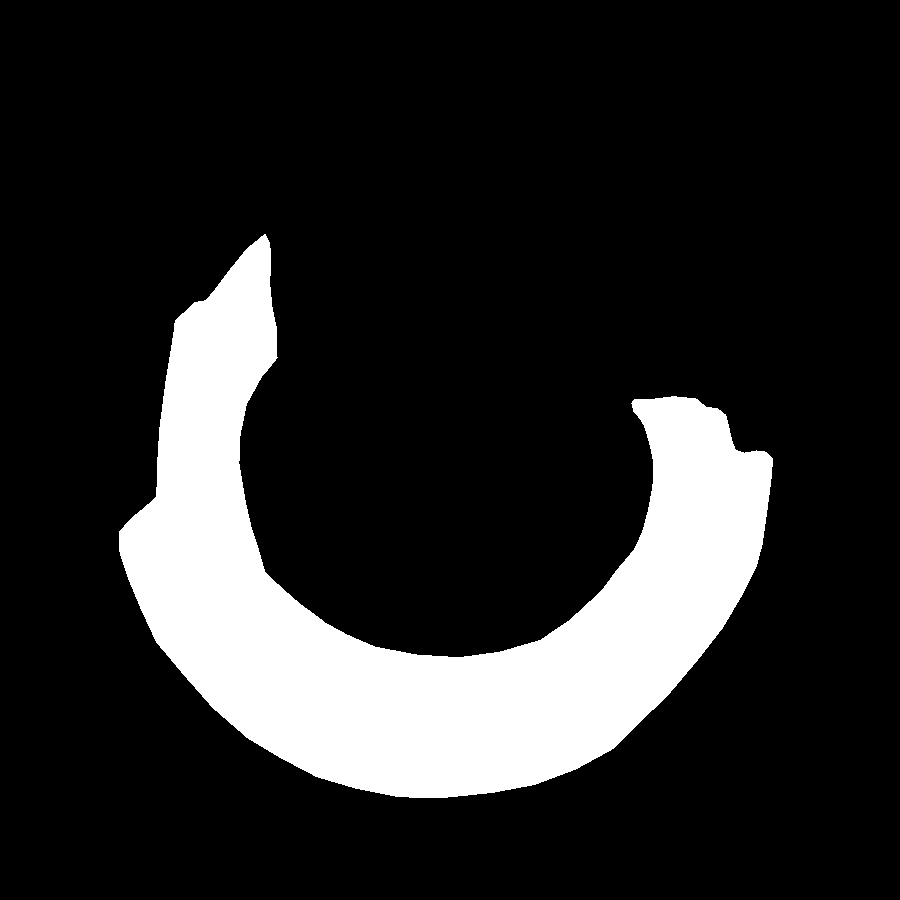}
\end{minipage}
\begin{minipage}[t]{0.23\textwidth}
\centering
\includegraphics[width=2.0cm]{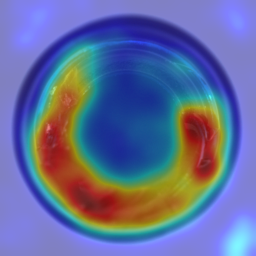}
\end{minipage}
\centering
\begin{minipage}[t]{0.23\textwidth}
\centering
\includegraphics[width=2.0cm]{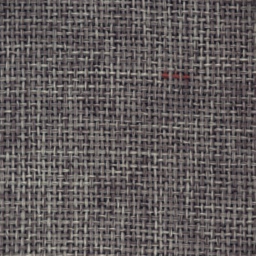}
\end{minipage}
\begin{minipage}[t]{0.23\textwidth}
\centering
\includegraphics[width=2.0cm]{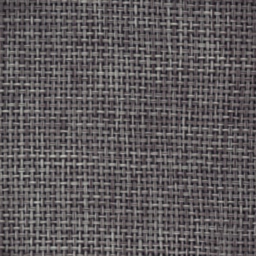}
\end{minipage}
\begin{minipage}[t]{0.23\textwidth}
\centering
\includegraphics[width=2.0cm, height=2.0cm]{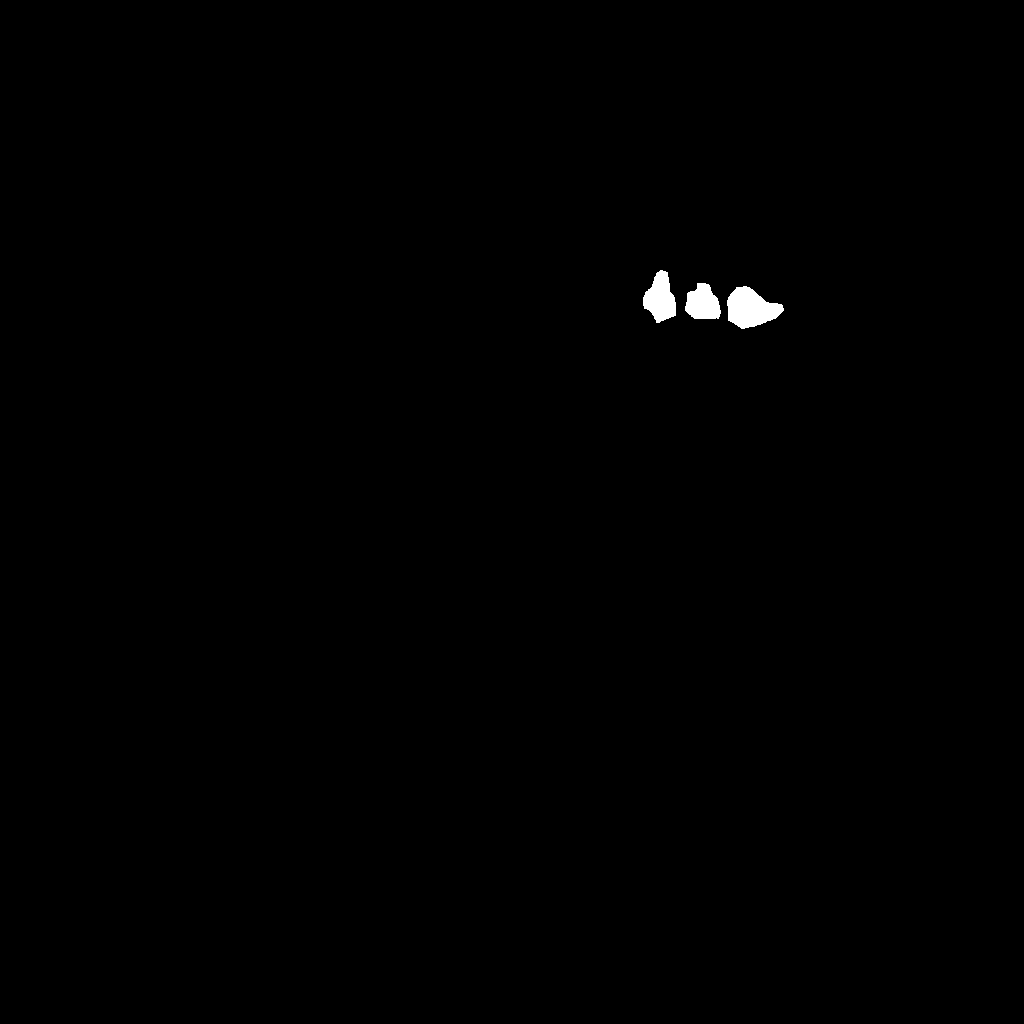}
\end{minipage}
\begin{minipage}[t]{0.23\textwidth}
\centering
\includegraphics[width=2.0cm]{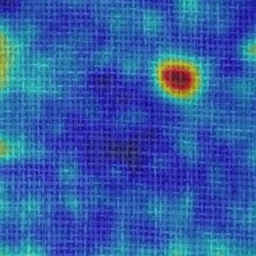}
\end{minipage}
\begin{minipage}[t]{0.23\textwidth}
\centering
Image
\end{minipage}
\begin{minipage}[t]{0.23\textwidth}
\centering
Image (Rec)
\end{minipage}
\begin{minipage}[t]{0.23\textwidth}
\centering
Label
\end{minipage}
\begin{minipage}[t]{0.23\textwidth}
\centering
Heat Map
\end{minipage}
\centering
\captionsetup{justification=centering}
\subcaption[]{Visualization of MVTec-AD \cite{mvtec}}

\end{subfigure}
\begin{subfigure}[t]{0.49\linewidth}
\centering
\begin{minipage}[t]{0.23\textwidth}
\centering
\includegraphics[width=2.0cm]{visualize/ViSA/candle/006-input.jpg}
\end{minipage}
\begin{minipage}[t]{0.23\textwidth}
\centering
\includegraphics[width=2.0cm]{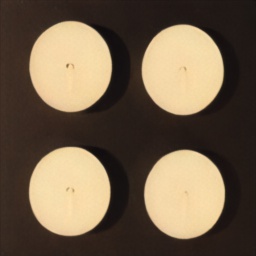}
\end{minipage}
\begin{minipage}[t]{0.23\textwidth}
\centering
\includegraphics[width=2.0cm, height=2.0cm]{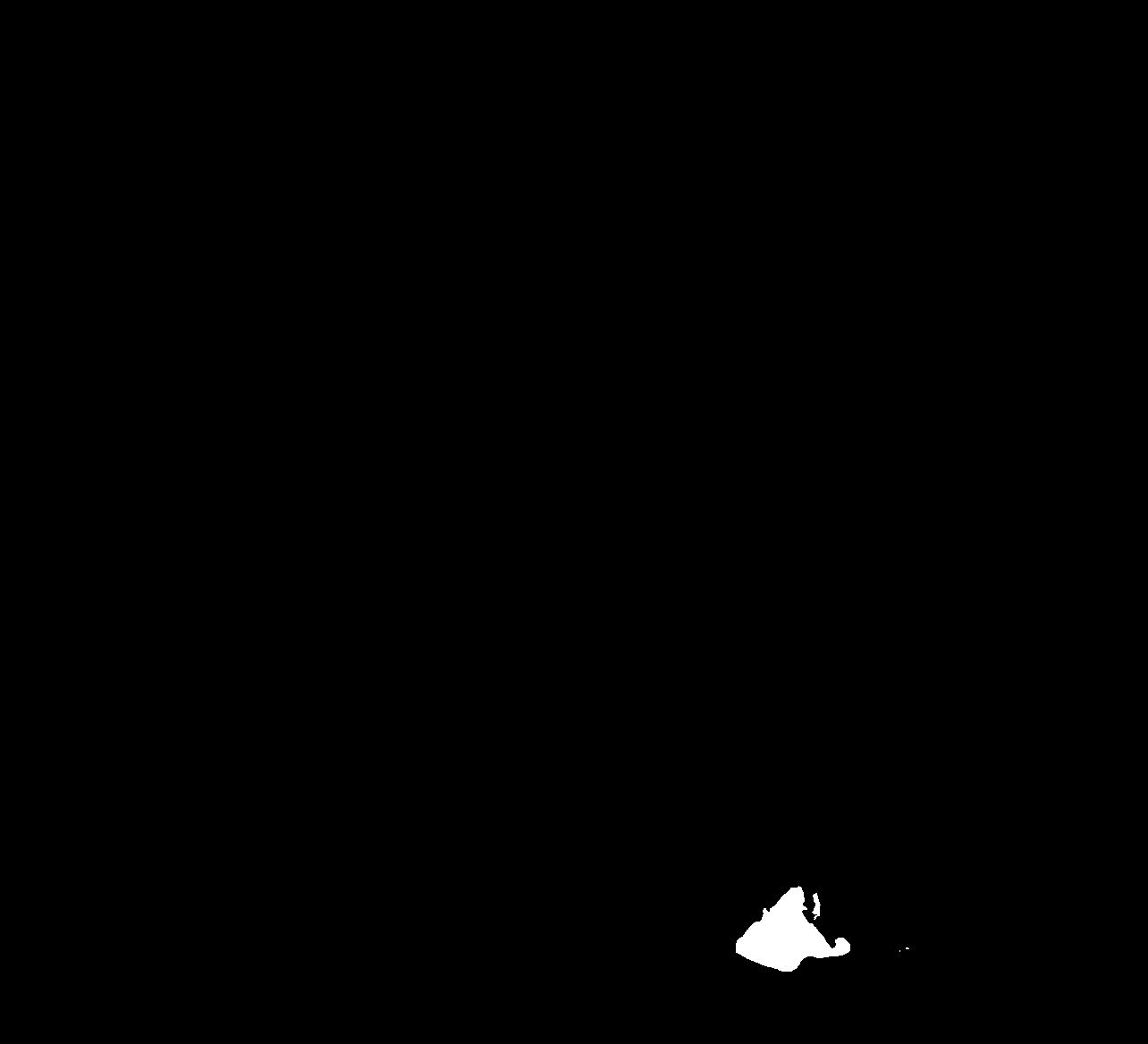}
\end{minipage}
\begin{minipage}[t]{0.23\textwidth}
\centering
\includegraphics[width=2.0cm]{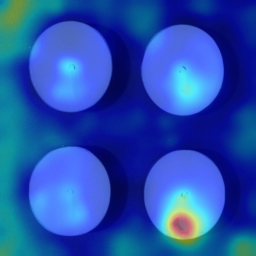}
\end{minipage}
\centering
\begin{minipage}[t]{0.23\textwidth}
\centering
\includegraphics[width=2.0cm]{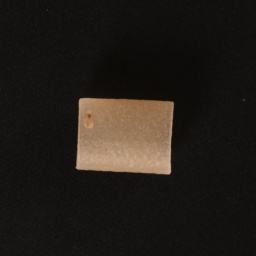}
\end{minipage}
\begin{minipage}[t]{0.23\textwidth}
\centering
\includegraphics[width=2.0cm]{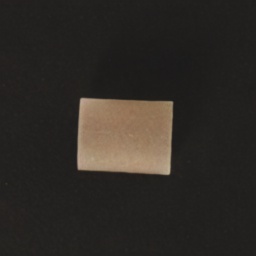}
\end{minipage}
\begin{minipage}[t]{0.23\textwidth}
\centering
\includegraphics[width=2.0cm, height=2.0cm]{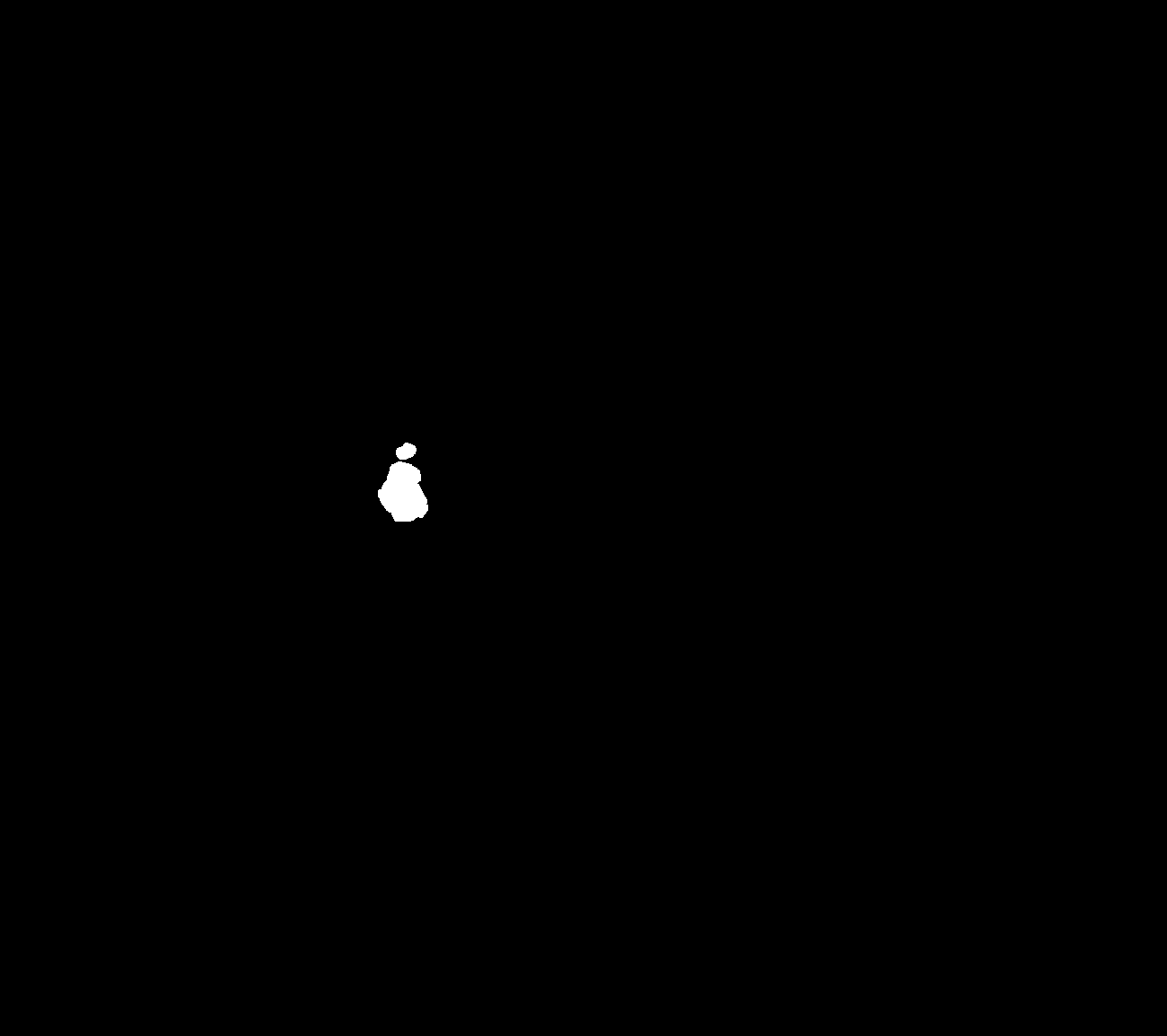}
\end{minipage}
\begin{minipage}[t]{0.23\textwidth}
\centering
\includegraphics[width=2.0cm]{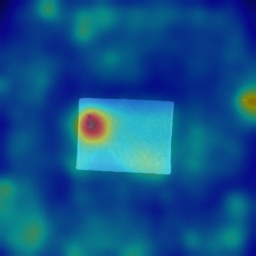}
\end{minipage}
\centering
\begin{minipage}[t]{0.23\textwidth}
\centering
\includegraphics[width=2.0cm]{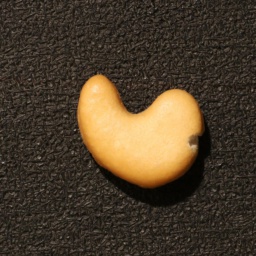}
\end{minipage}
\begin{minipage}[t]{0.23\textwidth}
\centering
\includegraphics[width=2.0cm]{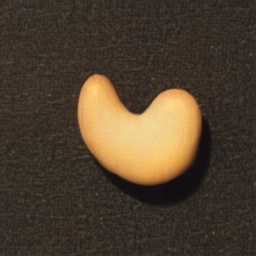}
\end{minipage}
\begin{minipage}[t]{0.23\textwidth}
\centering
\includegraphics[width=2.0cm, height=2.0cm]{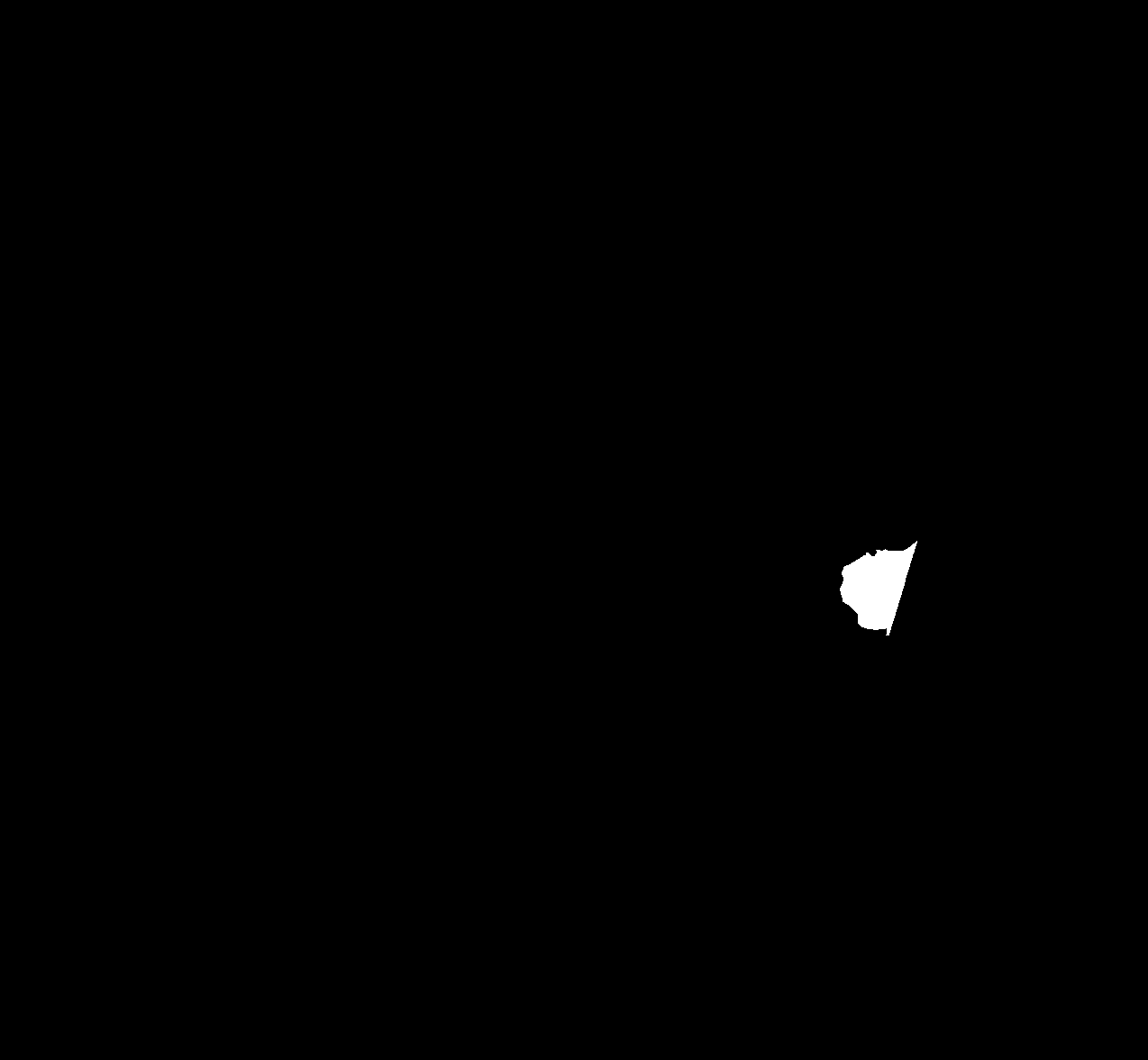}
\end{minipage}
\begin{minipage}[t]{0.23\textwidth}
\centering
\includegraphics[width=2.0cm]{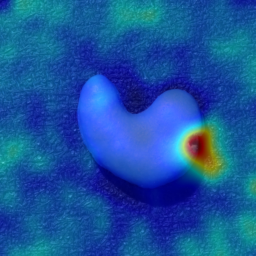}
\end{minipage}
\centering
\begin{minipage}[t]{0.23\textwidth}
\centering
\includegraphics[width=2.0cm]{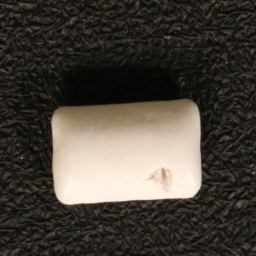}
\end{minipage}
\begin{minipage}[t]{0.23\textwidth}
\centering
\includegraphics[width=2.0cm]{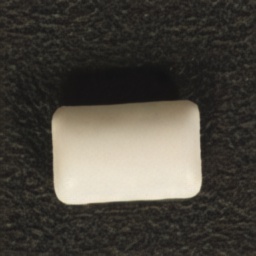}
\end{minipage}
\begin{minipage}[t]{0.23\textwidth}
\centering
\includegraphics[width=2.0cm, height=2.0cm]{visualize/ViSA/cashew/012.png}
\end{minipage}
\begin{minipage}[t]{0.23\textwidth}
\centering
\includegraphics[width=2.0cm]{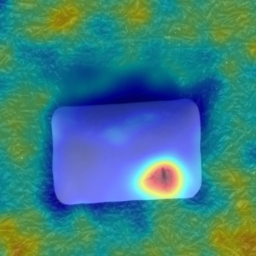}
\end{minipage}
\begin{minipage}[t]{0.23\textwidth}
\centering
Image
\end{minipage}
\begin{minipage}[t]{0.23\textwidth}
\centering
Image (Rec)
\end{minipage}
\begin{minipage}[t]{0.23\textwidth}
\centering
Label
\end{minipage}
\begin{minipage}[t]{0.23\textwidth}
\centering
Heat Map
\end{minipage}
\centering
\captionsetup{justification=centering}
\subcaption[]{Visualization of VisA \cite{VisA}}
\end{subfigure}
\end{figure*}
\begin{figure*}[p]\ContinuedFloat
\begin{subfigure}[p]{0.49\linewidth}
\begin{minipage}[p]{0.23\textwidth}
\centering
\includegraphics[width=2.0cm]{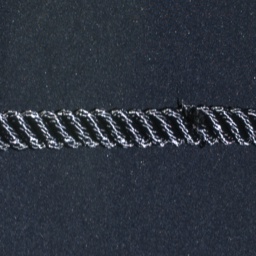}
\end{minipage}
\begin{minipage}[p]{0.23\textwidth}
\centering
\includegraphics[width=2.0cm]{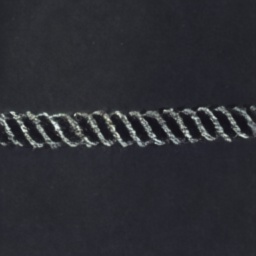}
\end{minipage}
\begin{minipage}[p]{0.23\textwidth}
\centering
\includegraphics[width=2.0cm, height=2.0cm]{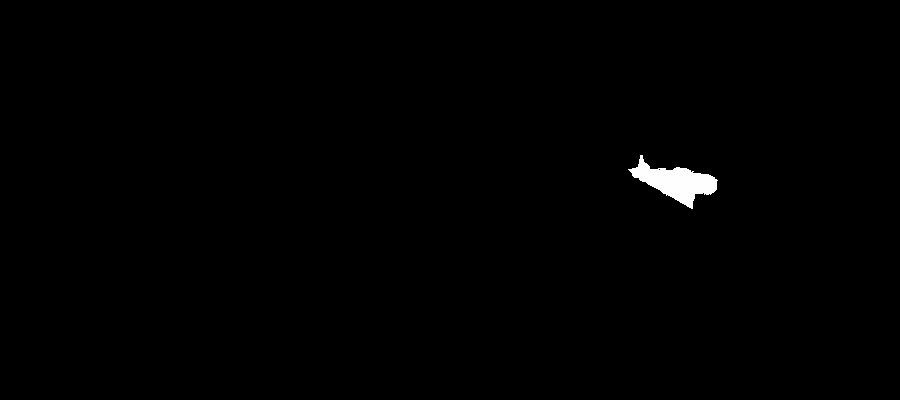}
\end{minipage}
\begin{minipage}[p]{0.23\textwidth}
\centering
\includegraphics[width=2.0cm]{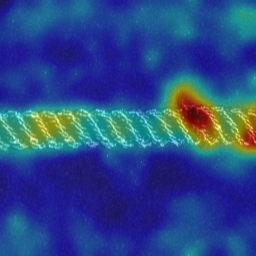}
\end{minipage}
\centering
\begin{minipage}[t]{0.23\textwidth}
\centering
\includegraphics[width=2.0cm]{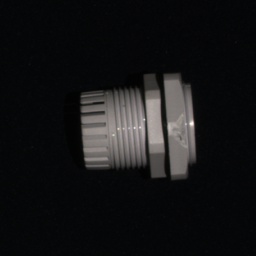}
\end{minipage}
\begin{minipage}[t]{0.23\textwidth}
\centering
\includegraphics[width=2.0cm]{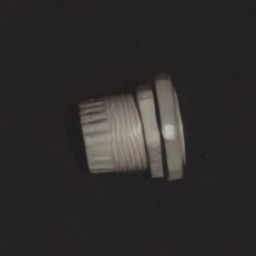}
\end{minipage}
\begin{minipage}[t]{0.23\textwidth}
\centering
\includegraphics[width=2.0cm]{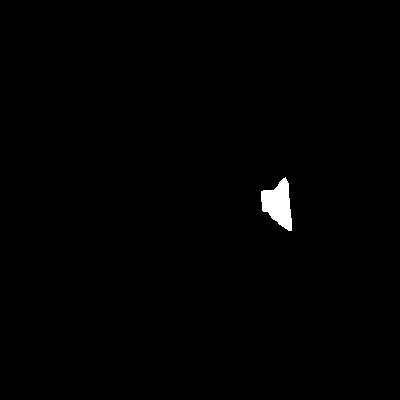}
\end{minipage}
\begin{minipage}[t]{0.23\textwidth}
\centering
\includegraphics[width=2.0cm]{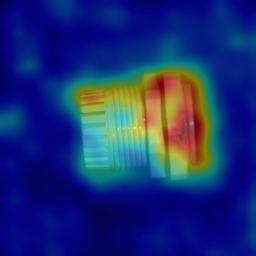}
\end{minipage}
\centering
\begin{minipage}[t]{0.23\textwidth}
\centering
\includegraphics[width=2.0cm]{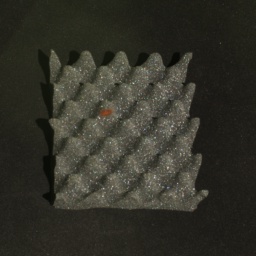}
\end{minipage}
\begin{minipage}[t]{0.23\textwidth}
\centering
\includegraphics[width=2.0cm]{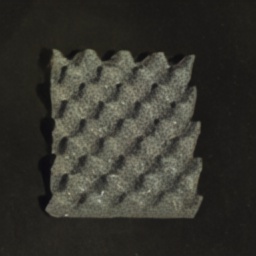}
\end{minipage}
\begin{minipage}[t]{0.23\textwidth}
\centering
\includegraphics[width=2.0cm]{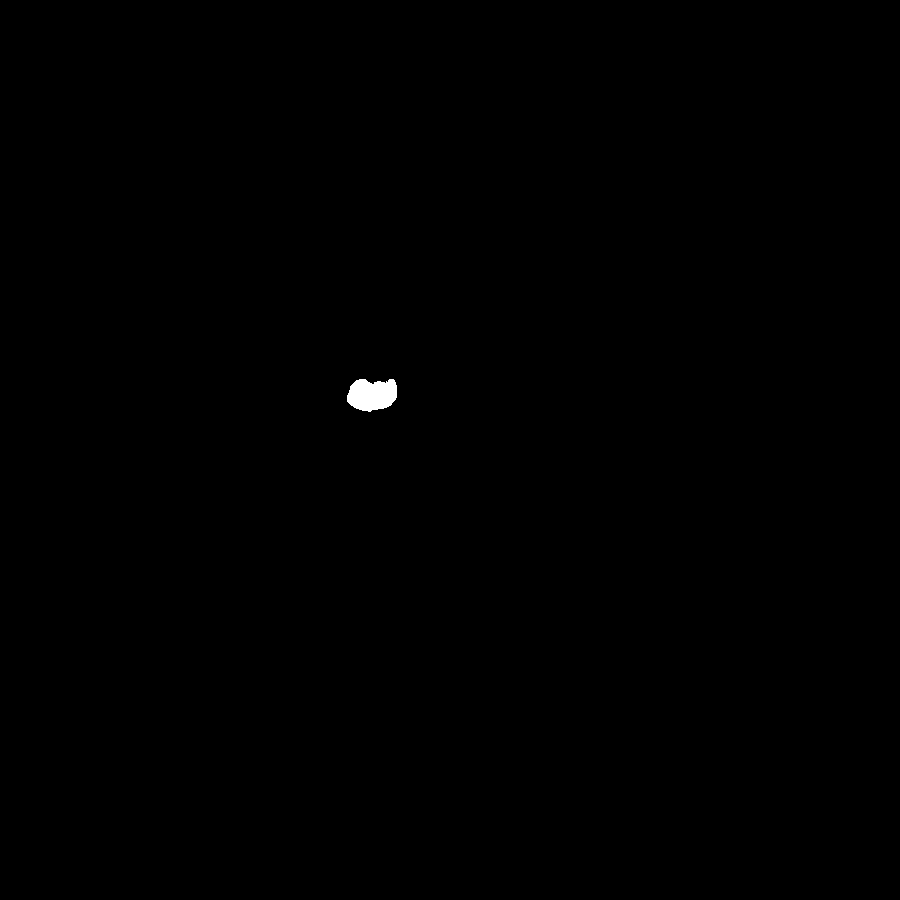}
\end{minipage}
\begin{minipage}[t]{0.23\textwidth}
\centering
\includegraphics[width=2.0cm]{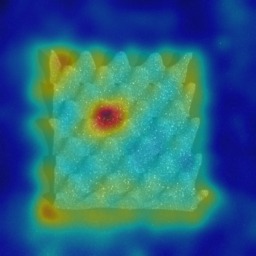}
\end{minipage}
\centering
\begin{minipage}[t]{0.23\textwidth}
\centering
\includegraphics[width=2.0cm]{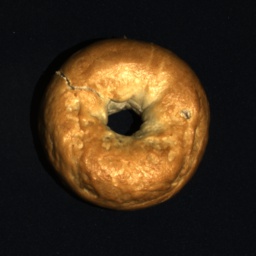}
\end{minipage}
\begin{minipage}[t]{0.23\textwidth}
\centering
\includegraphics[width=2.0cm]{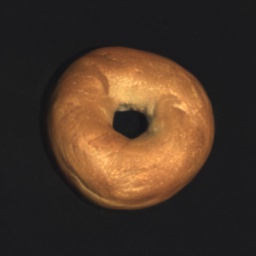}
\end{minipage}
\begin{minipage}[t]{0.23\textwidth}
\centering
\includegraphics[width=2.0cm]{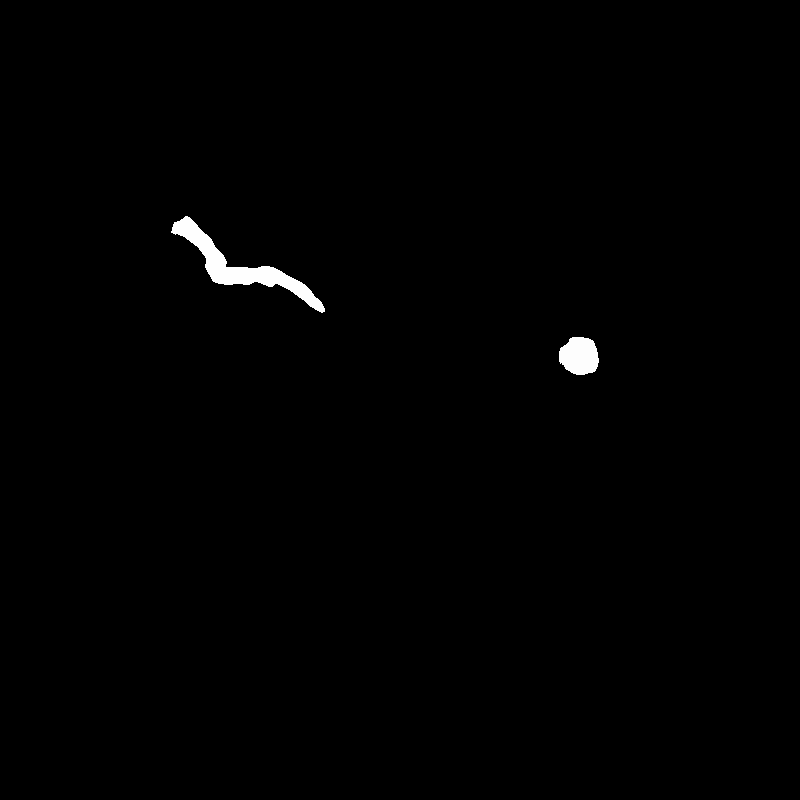}
\end{minipage}
\begin{minipage}[t]{0.23\textwidth}
\centering
\includegraphics[width=2.0cm]{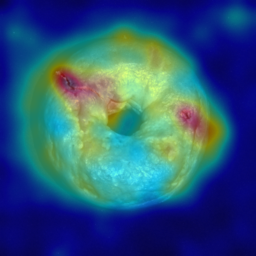}
\end{minipage}
\begin{minipage}[t]{0.23\textwidth}
\centering
Image
\end{minipage}
\begin{minipage}[t]{0.23\textwidth}
\centering
Image (Rec)
\end{minipage}
\begin{minipage}[t]{0.23\textwidth}
\centering
Label
\end{minipage}
\begin{minipage}[t]{0.23\textwidth}
\centering
Heat Map
\end{minipage}
\centering
\captionsetup{justification=centering}
\subcaption[]{Visualization of MVTec-3d \cite{MVTec-3d}}

\end{subfigure}
\begin{subfigure}[p]{0.49\linewidth}
\centering
\begin{minipage}[t]{0.23\textwidth}
\centering
\includegraphics[width=2.0cm]{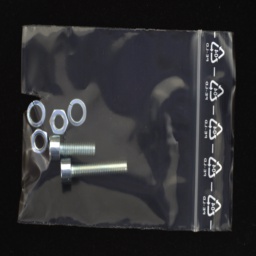}
\end{minipage}
\begin{minipage}[t]{0.23\textwidth}
\centering
\includegraphics[width=2.0cm]{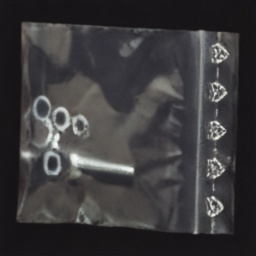}
\end{minipage}
\begin{minipage}[t]{0.23\textwidth}
\centering
\includegraphics[width=2.0cm, height=2.0cm]{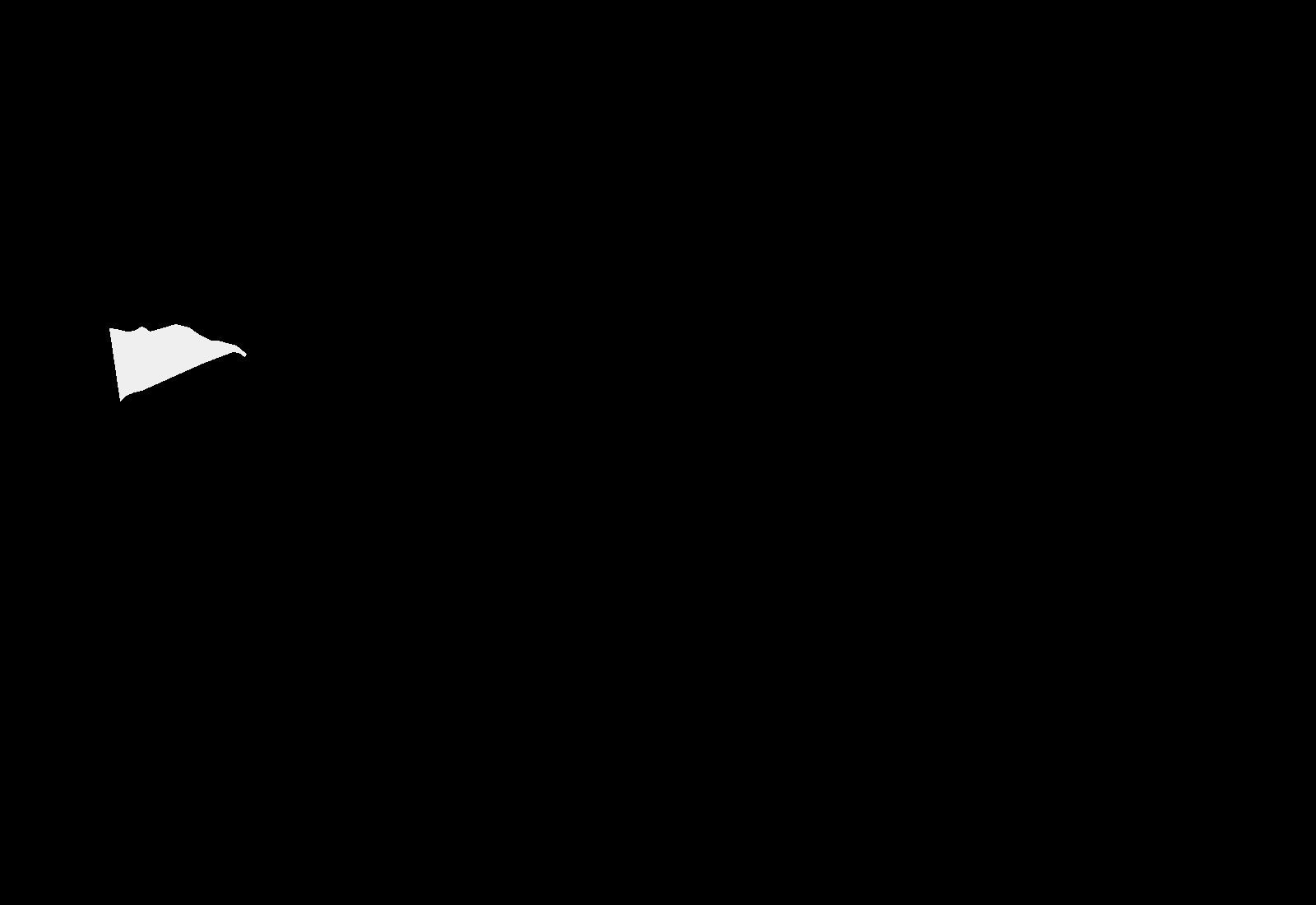}
\end{minipage}
\begin{minipage}[t]{0.23\textwidth}
\centering
\includegraphics[width=2.0cm]{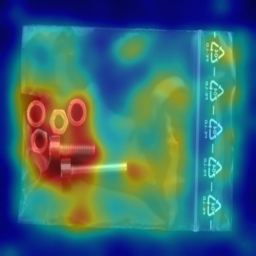}
\end{minipage}
\centering
\begin{minipage}[t]{0.23\textwidth}
\centering
\includegraphics[width=2.0cm]{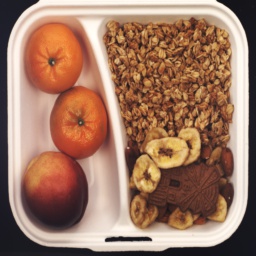}
\end{minipage}
\begin{minipage}[t]{0.23\textwidth}
\centering
\includegraphics[width=2.0cm]{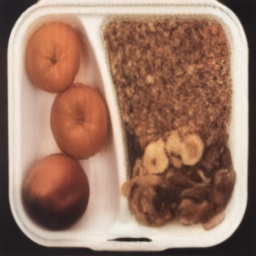}
\end{minipage}
\begin{minipage}[t]{0.23\textwidth}
\centering
\includegraphics[width=2.0cm, height=2.0cm]{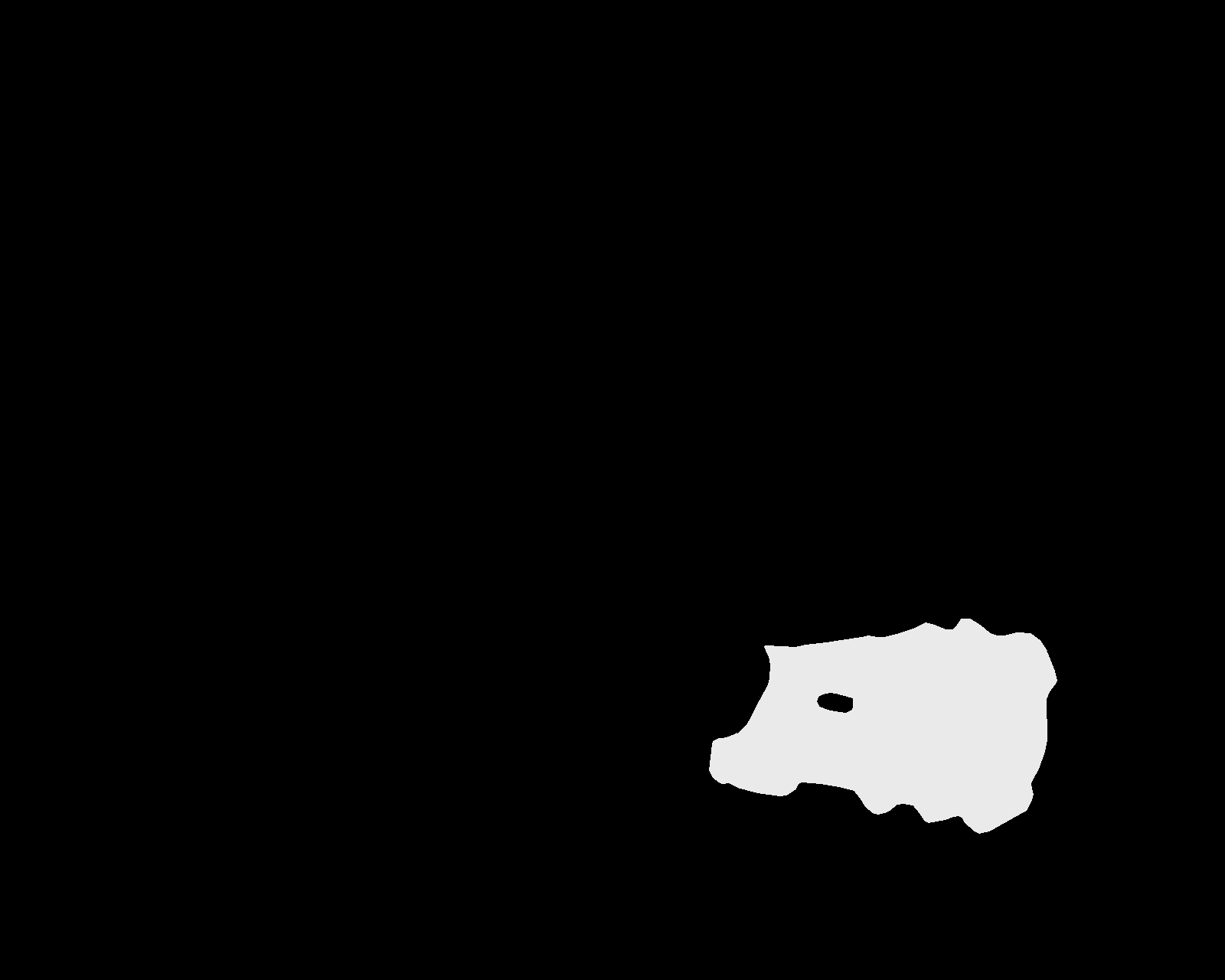}
\end{minipage}
\begin{minipage}[t]{0.23\textwidth}
\centering
\includegraphics[width=2.0cm]{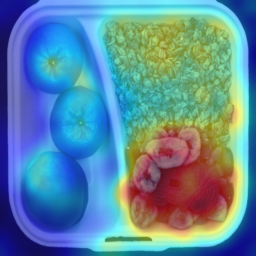}
\end{minipage}
\centering
\begin{minipage}[t]{0.23\textwidth}
\centering
\includegraphics[width=2.0cm]{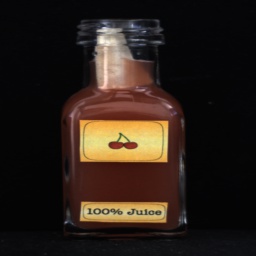}
\end{minipage}
\begin{minipage}[t]{0.23\textwidth}
\centering
\includegraphics[width=2.0cm]{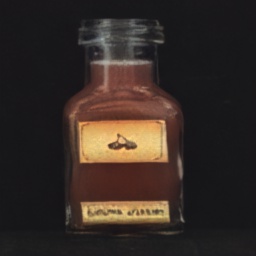}
\end{minipage}
\begin{minipage}[t]{0.23\textwidth}
\centering
\includegraphics[width=2.0cm, height=2.0cm]{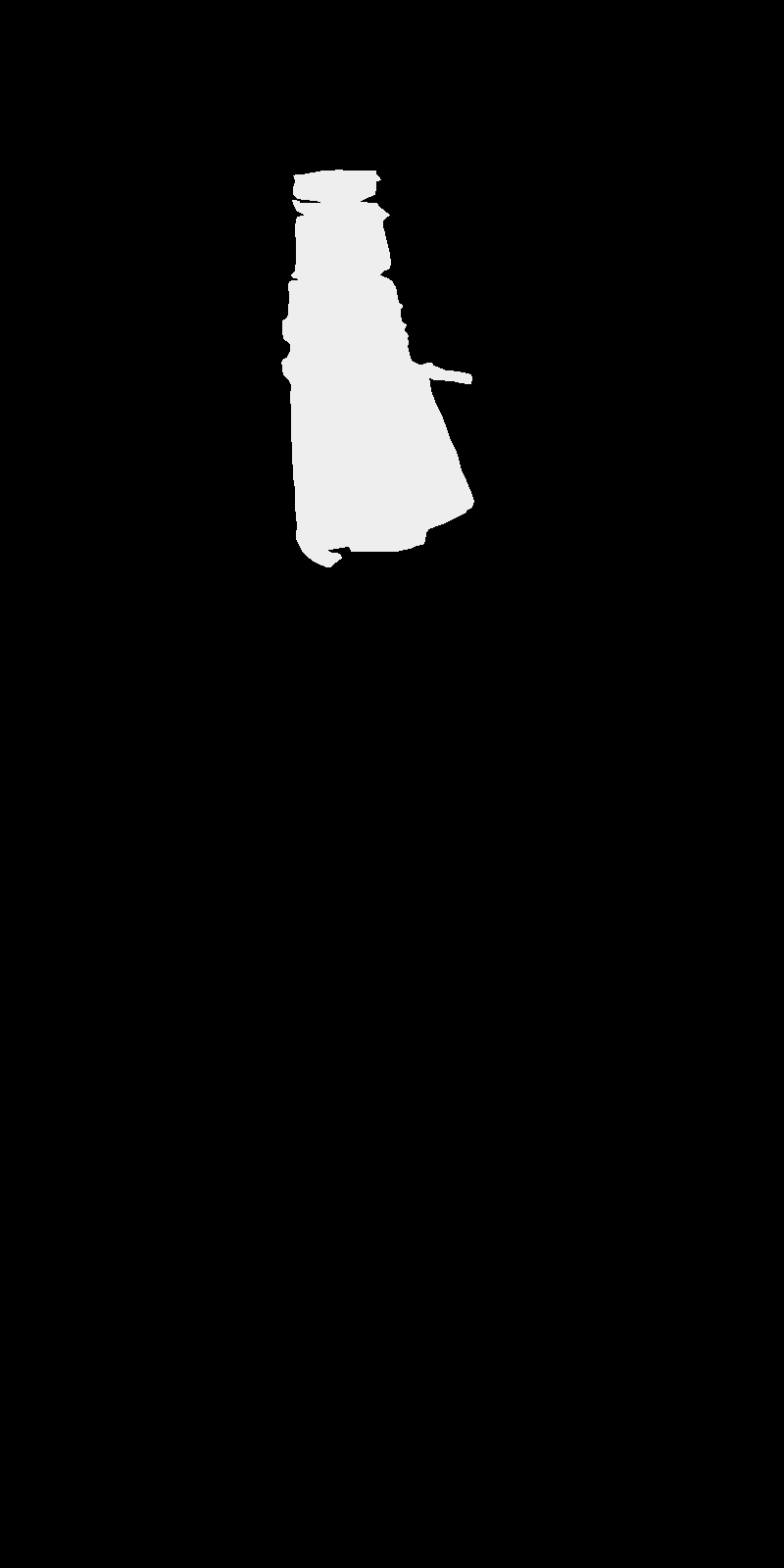}
\end{minipage}
\begin{minipage}[t]{0.23\textwidth}
\centering
\includegraphics[width=2.0cm]{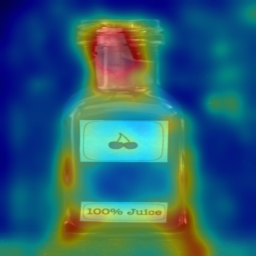}
\end{minipage}
\centering
\begin{minipage}[t]{0.23\textwidth}
\centering
\includegraphics[width=2.0cm]{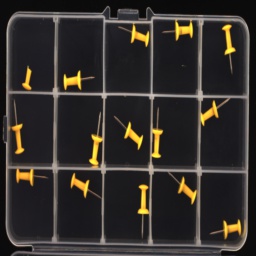}
\end{minipage}
\begin{minipage}[t]{0.23\textwidth}
\centering
\includegraphics[width=2.0cm]{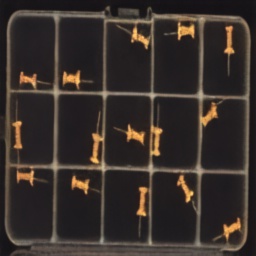}
\end{minipage}
\begin{minipage}[t]{0.23\textwidth}
\centering
\includegraphics[width=2.0cm, height=2.0cm]{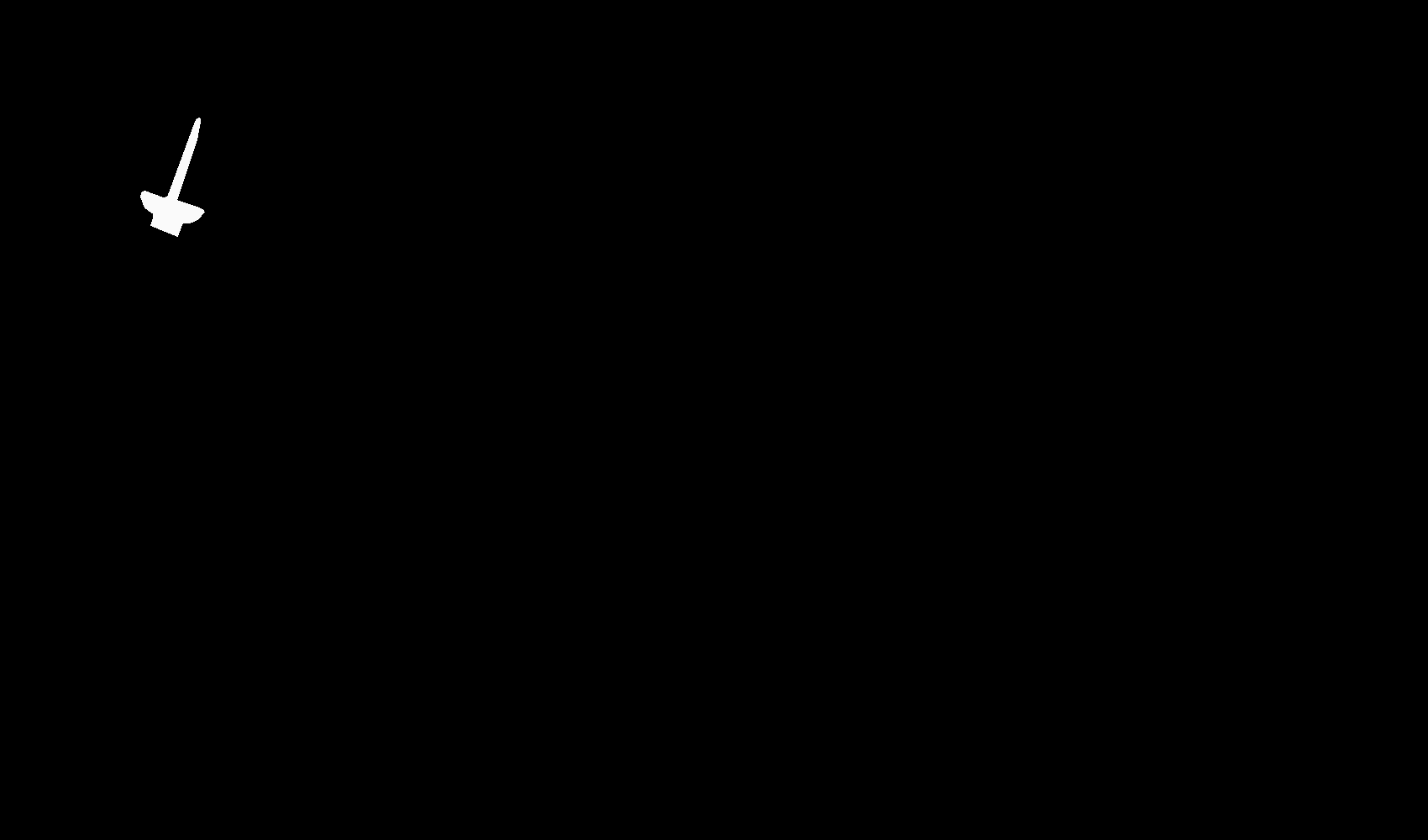}
\end{minipage}
\begin{minipage}[t]{0.23\textwidth}
\centering
\includegraphics[width=2.0cm]{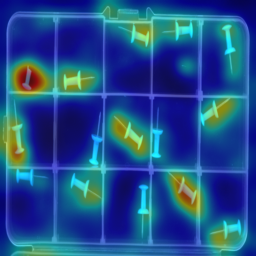}
\end{minipage}
\begin{minipage}[t]{0.23\textwidth}
\centering
Image
\end{minipage}
\begin{minipage}[t]{0.23\textwidth}
\centering
Image (Rec)
\end{minipage}
\begin{minipage}[t]{0.23\textwidth}
\centering
Label
\end{minipage}
\begin{minipage}[t]{0.23\textwidth}
\centering
Heat Map
\end{minipage}
\centering
\captionsetup{justification=centering}
\subcaption[]{Visualization of MVTec-Loco \cite{MVTec-loco}}

\end{subfigure}
\vspace{-2mm}
\caption{Qualitative example visualization on Datasets}
\label{Visualization Datasets}
\end{figure*}

\section{Acknowledge}
This work was supported by VisionX LLC. Models are trained on
8 NVIDIA A800 80GB PCIes and 8 NVIDIA GeForce RTX 3090s
\end{document}